\documentclass{article}

     \PassOptionsToPackage{numbers, compress}{natbib}
\usepackage[dblblindworkshop, final]{neurips_2025}

\usepackage[utf8]{inputenc} % allow utf-8 input
\usepackage[T1]{fontenc}    % use 8-bit T1 fonts
\usepackage{hyperref}       % hyperlinks
\usepackage{url}            % simple URL typesetting
\usepackage{booktabs}       % professional-quality tables
\usepackage{amsfonts}       % blackboard math symbols
\usepackage{nicefrac}       % compact symbols for 1/2, etc.
\usepackage{microtype}      % microtypography
\usepackage[dvipsnames]{xcolor}  % colors

\hypersetup{
    colorlinks=true, 
    linkcolor=Blue,
    filecolor=Blue,      
    urlcolor=Blue,
    citecolor=Blue % For biblatex.
    }

\usepackage{enumitem}
\setlist[itemize]{noitemsep, topsep=0pt}
\setlist[enumerate]{noitemsep, topsep=0pt}

\usepackage{adjustbox}
\usepackage{tabularray}
\UseTblrLibrary{booktabs}
\usepackage{array}
\newcolumntype{/}{!{\vrule width 1pt}}

\usepackage{amsmath}
\usepackage{amssymb}
\usepackage{mathtools}
\usepackage{amsthm}
\usepackage{dsfont} % mathbb for numerals
\usepackage{bbm}    % mathbb for numerals

\theoremstyle{plain}
\newtheorem{theorem}{Theorem}[section]

\theoremstyle{definition}
\newtheorem{definition}[theorem]{Definition}

\theoremstyle{remark}

\theoremstyle{example}
\newtheorem{example}[theorem]{Example}

\usepackage{tikz}
\usepackage{pgfplots,pgfmath,pgffor}
\usepackage{tikz-3dplot} % glow on edges
\usetikzlibrary {arrows.meta,arrows,patterns,decorations.pathreplacing,chains,positioning,shapes.geometric,shapes.callouts,calc,backgrounds,trees,pgfplots.fillbetween,quotes,calligraphy,automata,fit}

\usepackage{ragged2e}

\usepackage{wrapfig}

\usepackage[most]{tcolorbox}
\AtBeginEnvironment{tcolorbox}{\footnotesize}
\newenvironment{prompt}
    {\begin{tcolorbox}[enhanced,%breakable,
    attach boxed title to top center={yshift=-3mm,yshifttext=-1mm},
    colback=white,colframe=CornflowerBlue,colbacktitle=white,boxrule=0.5pt,coltext=black,%colframe=black!50
    title=Prompt,fonttitle=\color{CornflowerBlue}\itshape,fontupper=\ttfamily,size=small,
    boxed title style={size=small,colframe=CornflowerBlue,boxrule=0.75pt} ]
    \vspace{2mm}
    }
    { 
    \end{tcolorbox}
    }
\newenvironment{response}
    {\begin{tcolorbox}[ enhanced,%breakable,
    attach boxed title to top center={yshift=-3mm,yshifttext=-1mm},
    colback=white,colframe=WildStrawberry,colbacktitle=white,boxrule=0.5pt,coltext=black,%colframe=black!50
    title=Response,fonttitle=\color{WildStrawberry}\itshape,fontupper=\ttfamily,size=small,
    boxed title style={size=small,colframe=WildStrawberry,boxrule=0.75pt} ]
    \vspace{2mm}
    }
    { 
    \end{tcolorbox}
    }

\newcommand*\circled[1]{\tikz[baseline=(char.base)]{
            \node[shape=circle,draw,inner sep=1pt,minimum width=4mm] (char) {#1};}}

\usepackage{pifont}
\newcommand{\cmark}{{\color{black}\ding{51}}} % checkmark
\usepackage[toc,page,header]{appendix}
\usepackage{minitoc}
\title{CausalARC: \\ Abstract Reasoning with Causal World Models} 

\workshoptitle{Bridging Language, Agent, and World Models (LAW). $^*$Work done while an intern at YRIKKA; \texttt{maasch@cs.cornell.edu}}

\author{
    Jacqueline Maasch$^*$ \\
    Cornell Tech, New York, NY
    \And
    John Kalantari \\
    YRIKKA, New York, NY
    \And
    Kia Khezeli \\
    YRIKKA, New York, NY
} % Author order tentative, feel free to update.

\begin{document}

% FOR TOC IN APPENDIX
\doparttoc % Tell to minitoc to generate a toc for the parts
\faketableofcontents % Run a fake tableofcontents command for the partocs
%\part{} % Start the document part
%\parttoc % Insert the document TOC

\maketitle

\begin{abstract}
   On-the-fly reasoning often requires adaptation to novel problems under limited data and distribution shift. This work introduces CausalARC: an experimental testbed for AI reasoning in low-data and out-of-distribution regimes, modeled after the Abstraction and Reasoning Corpus (ARC). Each CausalARC reasoning task is sampled from a fully specified \textit{causal world model}, formally expressed as a structural causal model. Principled data augmentations provide observational, interventional,  and counterfactual feedback about the world model in the form of few-shot, in-context learning demonstrations. As a proof-of-concept, we illustrate the use of CausalARC for four language model evaluation settings: (1) abstract reasoning with test-time training, (2) counterfactual reasoning with in-context learning, (3) program synthesis, and (4) causal discovery with logical reasoning. Within- and between-model performance varied heavily across tasks, indicating room for significant improvement in language model reasoning.
   %\href{https://jmaasch.github.io/carc/}{https://jmaasch.github.io/carc/}
   %\includegraphics[height=1.5\fontcharht\font`\B]{figures/hf.png} [{\color{Blue}\textit{Project page and Hugging Face link released at camera-ready.}}] 
\end{abstract}

\section{Introduction}

\begin{wrapfigure}[16]{r}{0.35\linewidth}
    \centering
    \vspace{-5mm}
    \includegraphics[width=0.95\linewidth]{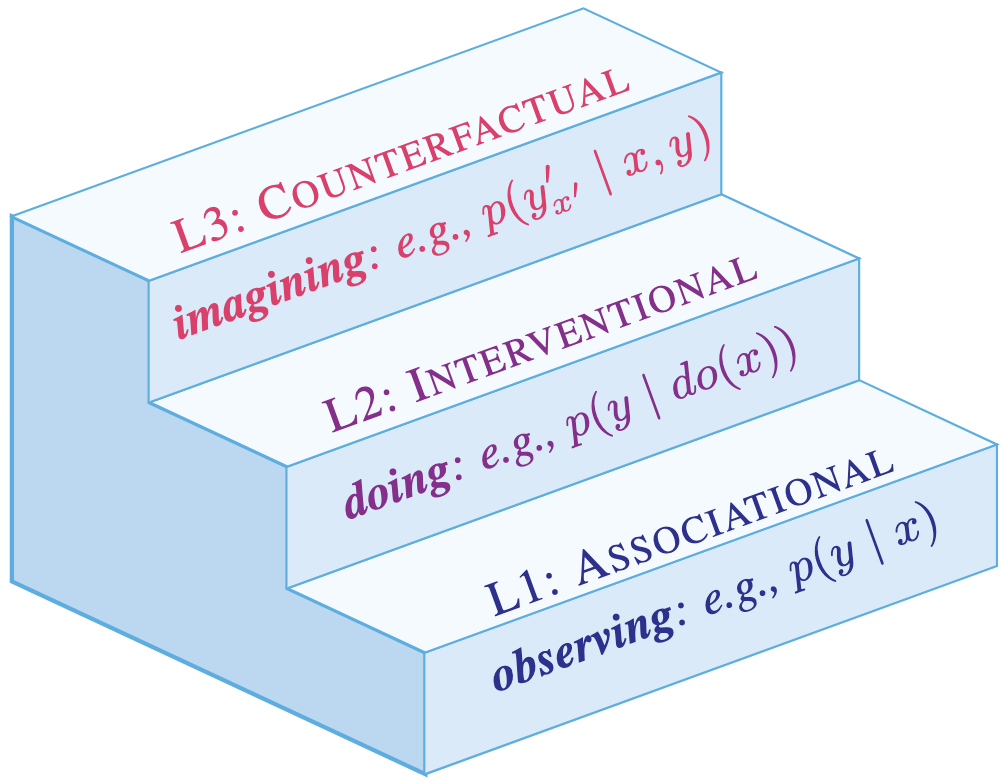}
    \caption{The PCH: %, also referred to as the \textit{ladder of causation}: 
     observing factual realities (L1), exerting actions to induce interventional realities (L2), and imagining alternate counterfactual realities (L3) \citep{bareinboim2022pch}.}
    \label{fig:pch}
\end{wrapfigure}

Humans are exceptional few- and zero-shot learners that use internal representations of the world to navigate novel scenarios \citep{ha2018world,lake2019human}.  These internal \textit{world models} can enable on-the-fly reasoning that is both \textit{robust} under ``known unknowns'' and \textit{flexible} under ``unknown unknowns'' \citep{chollet2019measure}. World models can encode beliefs about cause-effect relationships, supporting reasoning at all three levels of the Pearl Causal Hierarchy (PCH; Figure \ref{fig:pch}) \citep{bareinboim2022pch}. Though causal reasoning is a hallmark of human cognition \citep{goddu2024development} and a desideratum for human-like AI \citep{pearl2000models,lake2023human}, state-of-the-art generative models do not yet display robust and flexible reasoning at all three levels of the PCH 
\citep{jin2023cladder,gonzalez2024does,zevcevic2024parrots,shrestha2025causal,huyuk2024,maasch2025compositional}.

%The emergence of few-shot \citep{brown2020language} and in-context learning \citep{wei2023larger} in language models (LMs) raises new questions about the utility of world models in AI. 

Despite the emergence of few-shot \citep{brown2020language} and in-context learning \citep{wei2023larger} in language models (LMs), robust and flexible reasoning still poses a major distribution shift challenge. One promising direction for out-of-distribution problem-solving is \textit{test-time training} (TTT) \citep{sun2020test}, where model parameters are temporarily updated on each test instance (or batch) for on-the-fly adaptation to problems outside the pretraining distribution \citep{niu2022forgetting,liang2025comprehensive}. Testament to its utility for reasoning out-of-distribution, TTT was the dominant strategy in winning submissions for the first Abstraction and Reasoning Corpus (ARC) \citep{chollet2019measure,chollet2024arc,li2025combining,akyurek2024surprising}: the premier benchmark for general intelligence in AI to-date. ARC is a grid world of diverse abstract reasoning tasks that are generally solvable by humans. %, yet are challenging for narrow AI. 
Each task is governed by a unique rule that deterministically transforms input arrays to output arrays, as demonstrated by few-shot learning examples. Ablation studies show that fine-tuning plus TTT with few-shot, in-context learning can significantly improve LM performance on ARC relative to base models \citep{akyurek2024surprising}.

\begin{figure}
    \centering
    \includegraphics[width=\linewidth]{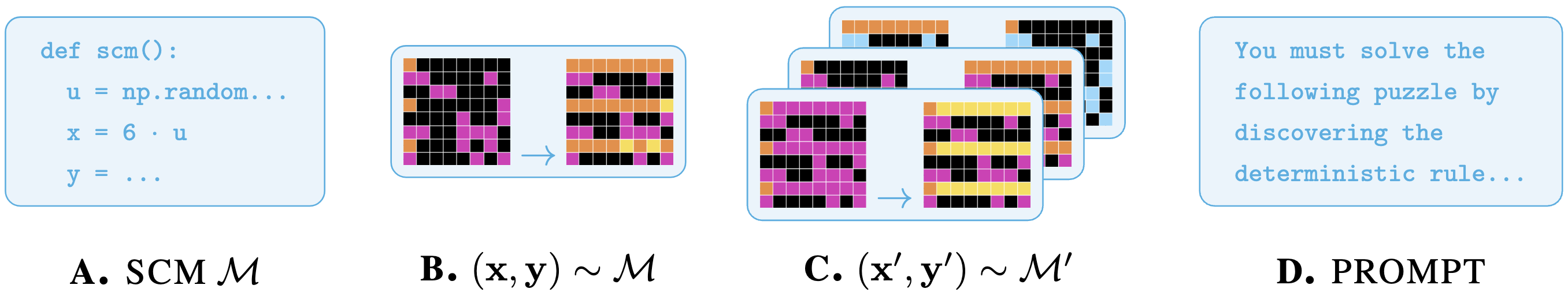}
    \caption{The CausalARC testbed. (\textbf{A}) First, SCM $\mathcal{M}$ is manually transcribed in Python code. (\textbf{B}) Input-output pairs are randomly sampled, providing observational (L1) learning signals about the world model. (\textbf{C}) Sampling from interventional submodels $\mathcal{M}'$ of $\mathcal{M}$ yields interventional (L2) samples $(\mathbf{x}',\mathbf{y}')$. Given pair $(\mathbf{x},\mathbf{y})$, performing multiple interventions while holding the exogenous context constant yields a set of counterfactual (L3) pairs. (\textbf{D}) Using L1 and L3 pairs as in-context demonstrations, we can automatically generate natural language prompts for diverse reasoning tasks.}
    \label{fig:demo}
\end{figure}

This work extends and reconceptualizes the ARC setup to support causal reasoning evaluation under limited data and distribution shift, while remaining amenable to TTT frameworks. We introduce CausalARC: an experimental testbed of ARC-like reasoning tasks sampled from fully specified \textit{causal world models} (Figure \ref{fig:demo}). World models are expressed as probabilistic \textit{structural causal models} (SCMs), a rich formalism for representing data generating processes and simulating hypothetical outcomes \citep{pearl2000models,pearl2013structural}. Given a fully specified SCM, all three levels of the PCH are well-defined: any observational (L1), interventional (L2), or counterfactual (L3) query can be answered about the environment under study \citep{bareinboim2022pch}. This formulation makes CausalARC an open-ended playground for testing reasoning hypotheses at all three levels of the PCH, with an emphasis on abstract, logical, and counterfactual reasoning.

\textbf{Design} \; CausalARC is designed to address multiple core problems in AI reasoning evaluation: 
        \begin{itemize}[leftmargin=*]
        \item \textit{Reasoning tasks as generative models.} 
        CausalARC frames each task as a generative model (SCM)  from which task instances can be randomly sampled.  Reasoning robustness can be benchmarked with respect to a distribution of task instances from each world model, enabling uncertainty estimation. Generative benchmarks that randomly sample from world models can also buffer against static benchmark data leakage, a major challenge in AI evaluation  \citep{mirzadeh2025gsm,xu2025reimagine,maasch2025compositional,gong2025phantomwiki}. 
        \item \textit{Disentangling signal from noise.} Differentiating true reasoning from statistical recall is a nontrivial objective in AI evaluation \citep{huyuk2024,xu2025reimagine,mirzadeh2025gsm,shojaee2025illusion}. ARC-like settings are conducive to this objective, as abstract reasoning relies more on innate cognitive priors than factual knowledge \citep{chollet2019measure}. Random task sampling also supports this objective by ensuring that exact test cases were never previously seen.
        %\item \textit{Robustness checks and uncertainty estimation.} Random sampling from the SCM enables distributions...
        \item \textit{Random sampling at all levels of the PCH}. Unlike prior ARC extensions, the user can sample L1, L2, and L3 distributions of any sample size from a given world model. This feature also facilitates causal discovery and causal inference queries that require a sample distribution.
        \item \textit{Fine-grained error analyses.} CausalARC is designed for hypothesis-driven model evaluations and detailed error analyses. As in ConceptARC \citep{moskvichev2023conceptarc}, tasks are labeled by reasoning theme (logical, counting, extension, and ordering). As in \citet{shojaee2025illusion}, CausalARC supports evaluation with respect to scaling problem complexity by configuring most tasks with tunable array sizes.
    \end{itemize}

\textbf{Intended Use Cases} \; CausalARC is a richly annotated experimental testbed for exploring diverse AI reasoning hypotheses.  Unlike ARC-Heavy \citep{li2025combining}, ARC-Potpourri \citep{li2025combining}, and RE-ARC \citep{hodel2024rearc}, CausalARC is not intended to be a large-scale fine-tuning dataset. We avoid the automated data augmentations used in  \citet{li2025combining}, \citet{hodel2024rearc}, and \citet{akyurek2024surprising} to ensure that data meet the stringent assumptions of the causal framework that we describe. Applications for CausalARC could include: benchmarking abstract reasoning in LMs, as in \citet{chollet2024arc}; benchmarking causal inference and causal discovery abilities, as in \citet{jin2023cladder} and \citet{jin2023can}; probing impacts of prompt formulation on in-context learning and TTT; benchmarking reasoning with respect to parametric assumptions on the SCM, topological properties of the causal graph, or scaling task complexity; etc. %CausalARC provides automated prompt generation for multiple settings, %including (1) control over the number of L1 and L3 in-context demonstrations and (2) system prompts tailored for 
%counterfactual reasoning, abstract reasoning, causal discovery, and program synthesis queries. 

%\input{figures/demo/figure_demo}

\textbf{Contributions}
\begin{enumerate}[leftmargin=*]
    \item[\S\ref{sec:causal_arc}] \textit{The CausalARC testbed}. We introduce an open-ended experimental testbed for AI reasoning  at all three levels of the PCH. %ARC-like reasoning tasks are accompanied by fully specified SCMs, expressed in mathematical notation and as Python programs. %CausalARC also contains noisy, manually-designed approximations of counterfactuals for 40 ARC-AGI tasks: 10 each of easy, medium, hard, and expert-level tasks. 
    We provide a static dataset and a public codebase for task generation.\footnote{See the official project page for Hugging Face and GitHub links: \href{https://jmaasch.github.io/carc/}{https://jmaasch.github.io/carc/}}
    \item[\S\ref{sec:results}] \textit{Empirical demonstrations.} As a proof-of-concept, we illustrate the use of CausalARC for four LM evaluation settings: (1) abstract reasoning with TTT, (2) counterfactual reasoning with in-context learning, (3) program synthesis, and (4) causal discovery with logical reasoning. Preliminary comparisons to ARC-AGI-1 suggest that CausalARC is of similar difficulty. %Results on setting (1) suggest that CausalARC is of comparable difficulty to ARC-AGI-1 for Llama 3 8B with fine-tuning and TTT.
\end{enumerate}

\section{Preliminaries}
\label{sec:prelim}

We briefly outline preliminaries on causal modeling and the ARC benchmark. Appendix \ref{appendix:prelims} contains extended preliminaries on world models, intelligence, generalization, and reasoning.

\subsection{Structural Causal Models}
\label{sec:prelim_causal}

The reasoning framework presented here draws from the theory of \textit{structural causal models} \citep{geffner2022probabilistic}. Moving forward, we denote random variables with uppercase letters, sets of random variables with bold letters, and  realizations in lowercase (e.g., $V = v$, $\mathbf{V} = \mathbf{v}$). %Calligraphic script is mainly used to denote models (e.g., SCMs, causal graphs, or neural networks). %and objects related to functions (e.g., sets of functions, domains, codomains, etc.).

\begin{definition}[Structural causal model (SCM), \citet{bareinboim2022pch}] \label{def:scm}
    An SCM is a tuple $\mathcal{M} \coloneqq \langle \mathbf{U}, p(\mathbf{u}), \mathbf{V}, \mathcal{F} \rangle$ where $\mathbf{U} = \{U_i\}_{i=1}^m$ is the set of exogenous variables explained by  mechanisms external to $\mathcal{M}$, $p(\mathbf{u})$ is the  distribution over $\mathbf{U}$, $\mathbf{V} = \{V_i\}_{i=1}^n$ is the set of  endogenous variables explained by variables in $\mathbf{U} \cup \mathbf{V}$, and $\mathcal{F} = \{f_i\}_{i=1}^n$ is the set of structural functions such that $v_i = f_i(\mathbf{pa}_{v_i}, u_i)$ for endogenous parent set $\mathbf{pa}_{v_i}$ and exogenous context $u_i$.
\end{definition}
%Any SCM is associated with a corresponding graphical model. In this work, we restrict our attention to SCMs that can be represented by directed acyclic graphs (DAGs). 
The joint factorization of $p(\mathbf{v})$ implied by $\mathcal{M}$ can be represented as a causal graph $\mathcal{G}$. This work assumes that $\mathcal{G}$ is a directed acyclic graph (DAG), justified by the fact that our testbed does not have a time series component (and thus no feedback loops). Additionally, we assume a positive SCM:  $p(v_i) > 0$ for every realization $V_i \in \mathbf{V} = v_i$ \citep{pearl_probabilities_1999}. We do not make assumptions about confounding. Future work could explore this setup under varying assumptions. 

%\jm{Actually, we probably want to relax Markovianity here.}

%\begin{definition}[Positive-Markovian SCM, \citet{pearl_probabilities_1999}] \label{def:pos_markovian_model}
%    An SCM is Markovian if its graphical representation is acyclic and exogenous variables $U_i$ are mutually independent. An SCM is positive-Markovian if it is Markovian and $p(v_i) > 0$ for every realization $V_i \in \mathbf{V} = v_i$.
%\end{definition}

Crucially, we can exert actions or \textit{interventions} on $\mathcal{M}$ to induce \textit{interventional distributions}. In the real world, intervention corresponds to controlled experimentation.

%In contrast, soft interventions do not fix $V_i$ to a constant value, but instead modify local conditional distributions.
%{\color{Purple}
%If you're going to mention that this work assumes SCM is acyclic, you should spell out all other  key identifiability assumptions upfront here as well (" satisfies the Causal Markov condition,  every intervention has nonzero support”,etc). That way, when you later invoke back-door arguments or discuss “valid” interventions, the reader already knows where you stand.
%} << JM: absolutely! Added above.

\begin{definition}[Hard intervention]
    A hard intervention $do(V_i = v_i)$ replaces the true causal function $f_i(\mathbf{pa}_{v_i}, u_i)$ with the constant function evaluating to $v_i$. 
\end{definition}

\begin{definition}[Soft intervention]
    A soft intervention modifies the local conditional distribution $p(v_i \mid \mathbf{pa}_{v_i})$ to some new distribution $q({v_i} \mid \mathbf{pa}'_{v_i})$, where  $\mathbf{pa}'_{v_i}$ may or may not differ from $\mathbf{pa}_{v_i}$.
\end{definition}

Next, we can define \textit{counterfactuals}. In the real world, counterfactuals are not measurable through controlled experimentation. Instead, they correspond to  alternate hypothetical realities.

\begin{definition}[Counterfactual, \citet{pearl2013structural}] \label{def:cf}
 Let $\mathcal{M}_x$ be the submodel of $\mathcal{M}$ induced by hard or soft intervention on $X \in \mathbf{V}$. Let $Y \in \mathbf{V}$ be a variable whose value we wish to query. The counterfactual $Y_x$ under model $\mathcal{M}$ is then expressed as $Y_x(\mathbf{u}) \coloneqq Y_{\mathcal{M}_x}(\mathbf{u})$.
\end{definition}

That is, the counterfactual under the original SCM is equal to the value taken by $Y$ under the interventional submodel $\mathcal{M}_x$, with the exogenous context $\mathbf{u}$ fixed. For hard interventions, $Y_x(\mathbf{u}) = y$ could be verbalized as ``$Y$ would have been $y$ had $X$ been $x$ in context $\mathbf{U} = \mathbf{u}$'' \citep{pearl2013structural}. Thus, counterfactuals are jointly distributed random variables in a shared probability space over $\mathbf{U}$ \citep{tian2000probabilities}.

With these definitions of interventions and counterfactuals in hand, we can define a \textit{hierarchy of information} encoded in the distributions induced by the submodels of $\mathcal{M}$ (Figure \ref{fig:pch}).

%Graphically, hard interventions correspond to eliminating all edges entering into $V_i$. Soft interventions do not alter the underlying graph. In genetics, a hard intervention would be a \textit{gene knockout} experiment, whereas a soft intervention would correspond to a \textit{gene knockdown} \citep{brouillard2020differentiable}. \jm{Add electrical engineering example. See \citep{richens2024robust} for abstract examples.} In this work, the underlying SCM defining each reasoning task is unknown. Thus, we make the simplifying assumption that all interventions performed are either hard or soft, though the classification of each individual intervention may be unknown. For ease of exposition, subsequent expressions in this section will generally assume that variables are binary. 

\begin{definition}[Pearl Causal Hierarchy (PCH), \citet{bareinboim2022pch}] \label{def:pch}
    Let $\mathcal{M}$ be a fully specified SCM, per Definition \ref{def:scm}. The PCH is the set of all observational (layer L1), interventional (layer L2), and counterfactual (layer L3) distributions induced by $\mathcal{M}$.
\end{definition}

Given access to the fully specified SCM $\mathcal{M}$, the PCH is well-defined: the information encoded in $\mathcal{M}$ enables valuation for any quantity at L1, L2, and L3 \citep{bareinboim2022pch}. Notably, lower layers of the PCH generally underdetermine higher layers: L$_j$ encodes information sufficient to answer queries at L$_{i \leq j}$, yet knowledge of L$_{i < j}$ is \textit{almost never} sufficient to answer queries at L$_j$. We refer the reader to the Causal Hierarchy Theorem (CHT) for formal proof, as presented in \citet{bareinboim2022pch}. %That is, L$_j$ rarely \textit{collapses} to L$_{i < j}$. We say that L$_j$ of the causal hierarchy \textit{collapses} to L$_{i < j}$ relative to $\mathcal{M}  \in \Omega$ if $\mathcal{M} \sim_i \mathcal{M}'$ implies that $\mathcal{M} \sim_j \mathcal{M}'$ for all $\mathcal{M}' \in \Omega$ \citep{bareinboim2022pch}. 

%\begin{theorem}[Causal Hierarchy Theorem (CHT), %\citet{bareinboim2022pch}] \label{theorem:cht}
%    For almost any SCM, the PCH does not collapse.
%\end{theorem}

%Following from the CHT, answering queries at L$_i$ of the PCH generally requires  knowledge of L$_{j \geq i}$.

%\begin{corollary}[Lower levels of the PCH underdetermine higher levels, \citet{bareinboim2022pch}]
%\label{corollary:underdetermined}
 %Following from the CHT, answering queries at L$_i$ of the PCH generally requires  knowledge of L$_{j \geq i}$.   
%\end{corollary}

%The CHT is highly consequential for causal inference and statistics: L2 collapsing to L1 would imply that causal queries can be answered via correlation, while L3 collapsing to L2 would imply that counterfactual queries can be answered via controlled experimentation \citep{bareinboim2022pch}. Per the CHT, the probability that these forms of collapse occur for a given SCM approaches zero.

%\textbf{From prior notes: }Note that $p(y_x)$ and $p(y_{x'})$ are interventional distributions, while $p(x,y)$, $p(x',y)$, $p(x,y')$, and $p(x',y')$ are observational distributions \citep{li2022probabilities}. The causal effects $p(y_x)$ and $p(y_{x'})$ can be obtained through randomized experimentation or from observational data under sufficient conditions (e.g., conditional exchangeability achieved through valid covariate adjustment) \citep{tian2000probabilities}.

\subsection{ARC-AGI: The Abstraction \& Reasoning Corpus}
\label{sec:prelim_arc}

\begin{figure}
    \centering
    \begin{tabular}{c/c/c/c/c/c}
    \toprule[1pt]
    \multicolumn{3}{c}{\footnotesize\textsc{ARC-AGI-1}} & \multicolumn{3}{c}{\footnotesize\textsc{ARC-AGI-2}} \\
    \cmidrule(lr){1-3} \cmidrule(lr){4-6} 
        \includegraphics[height=0.03\textheight]{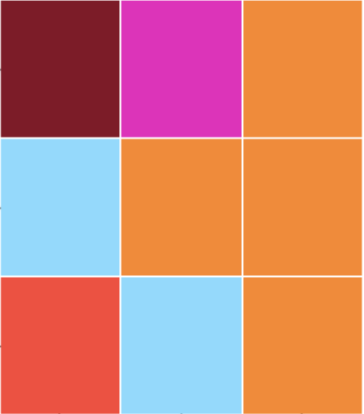} $\to$ \includegraphics[height=0.03\textheight]{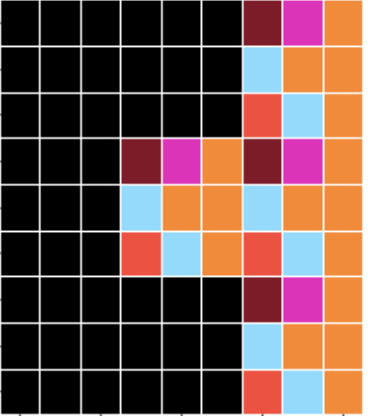} & 
        \includegraphics[height=0.03\textheight]{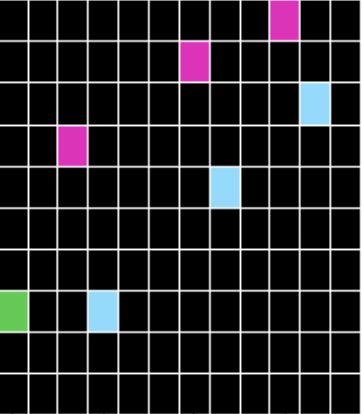} $\to$ \includegraphics[height=0.03\textheight]{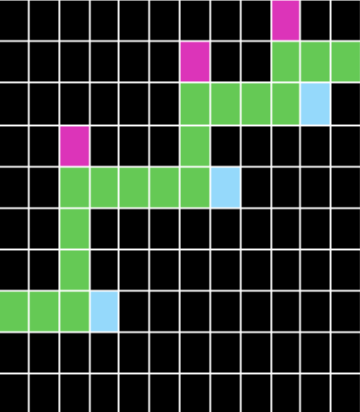} &
        \includegraphics[height=0.03\textheight]{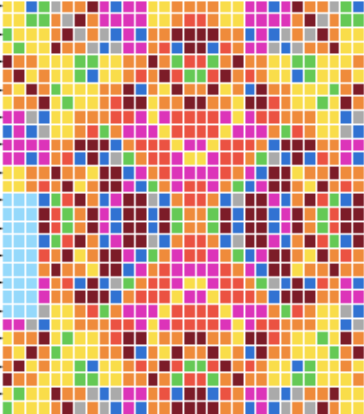} $\to$ \includegraphics[height=0.03\textheight]{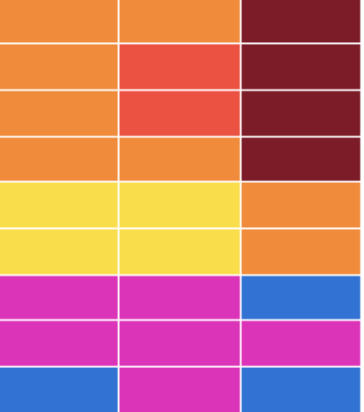} &
        \includegraphics[height=0.03\textheight]{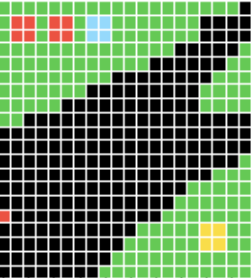} $\to$ \includegraphics[height=0.03\textheight]{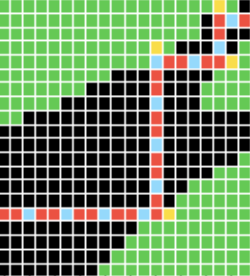} &
        \includegraphics[height=0.03\textheight]{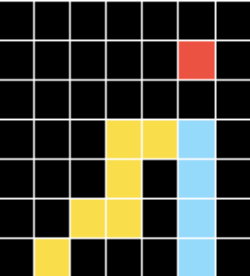} $\to$ \includegraphics[height=0.03\textheight]{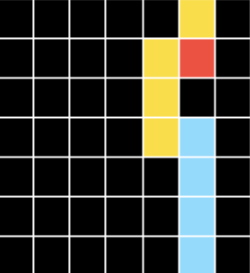} &
        \includegraphics[height=0.03\textheight]{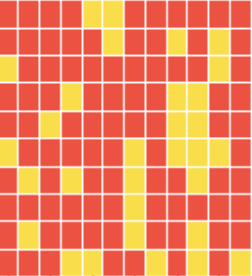} $\to$ \includegraphics[height=0.03\textheight]{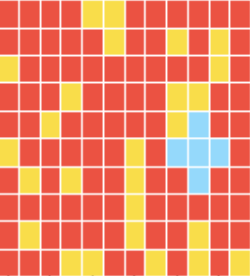} 
        \\
    \bottomrule[1pt]
    \end{tabular}
    \caption{Example input-output pairs from ARC-AGI-1 and ARC-AGI-2.}
    \label{fig:arc}
\end{figure}

\textbf{Background} \; The first Abstraction and Reasoning Corpus (ARC-AGI-1) was introduced in 2019 \citep{chollet2019measure}.\footnote{\href{https://arcprize.org/arc-agi/1/}{https://arcprize.org/arc-agi/}} At the time of writing, the second ARC challenge is  underway\footnote{\href{https://www.kaggle.com/competitions/arc-prize-2025}{https://www.kaggle.com/competitions/arc-prize-2025}} and the third is in development. ARC is a grid world, represented by two-dimensional arrays of pixels that can take on one of ten colors each. For ARC-AGI-1, data consist of a training set ($n = 400$ reasoning tasks), a public evaluation set ($n = 400$), a semi-private evaluation set ($n = 100$), and a fully private evaluation set ($n = 100$). Each grid is an array of any dimensionality from $1 \times 1$ to $30 \times 30$ (Figure \ref{fig:arc}). Arrays vary widely in appearance, at times appearing as randomly dispersed colors. Arrays can also feature distinct multi-pixel shapes, referred to as \textit{sprites} (as in computer graphics).%: two-dimensional bitmaps that are bit block transferred (\textit{blitted}) onto the background.

The test-taker must solve each task by discovering the deterministic rule or transformation that maps input arrays to output arrays. Each task provides approximately 2–5 input-output pairs as examples to demonstrate the rule, with no additional clues provided. An average human should ostensibly be able to solve most or all tasks from these demonstrations alone, with no specialized knowledge or training. Instead, problem-solving requires innate cognitive priors, such as elementary arithmetic, basic geometry, and intuitive physics \citep{chollet2019measure}.

%Instead, problem-solving requires a \jm{``minimal set of cognitive priors, such as object permanence, elementary number sense, and basic geometric and topological concepts.''}

\textbf{Notation} \; Let $\mathcal{D} \coloneqq \mathcal{D}_{train} \cup \mathcal{D}_{eval}$ denote the ARC dataset. As we are concerned with in-context learning and TTT but not pretraining or fine-tuning, we consider only $\mathcal{D}_{eval}$ moving forward. Each instance in $\mathcal{D}_{eval}$ is a reasoning task $\mathbf{T}_i$, such that $\mathcal{D}_{eval} \coloneqq \{\mathbf{T}_i\}_{i=1}^n$. Each $\mathbf{T}_i$ is an $m$-shot learning task associated with its own input space $\mathcal{X}$ and output space $\mathcal{Y}$. As such, $\mathbf{T}_i$ is comprised of $m+1$ tuples of paired input and output arrays, where $m$ tuples are \textit{demonstration pairs} $(\mathbf{x}_{train},\mathbf{y}_{train}) \in \mathcal{X}^{m} \times \mathcal{Y}^{m}$ and one is a \textit{test pair} $(\mathbf{x}_{test}, \mathbf{y}_{test}) \in \mathcal{X} \times \mathcal{Y}$. Thus, $\mathbf{T}_i \coloneqq \{ (\mathbf{x}_{train},\mathbf{y}_{train})\} \cup \{(\mathbf{x}_{test}, \mathbf{y}_{test})\}$. When sufficiently clear, we use $(\mathbf{x},\mathbf{y})$ to denote a single random input-output pair. Each $\mathbf{T}_i$ is governed by a deterministic \textit{rule} or \textit{transformation} $\delta_i : \mathcal{X} \to \mathcal{Y}$, which maps each input array to its respective output. Let $\mathcal{A}$ be our test-taker (e.g., an LM). To succeed on $\mathbf{T}_i$, $\mathcal{A}$ must learn $\delta_i$ sufficiently well to correctly predict $\mathbf{y}_{test}$ from $\mathbf{x}_{test}$ (with predictions denoted as $\widehat{\mathbf{y}}_{test}$). $\mathcal{D}_{eval}$ is presented to $\mathcal{A}$ as string representations of Python arrays. As in \citet{akyurek2024surprising}, we encode colors as  integers: $[0,{\color{NavyBlue}1},{\color{Red}2},{\color{Green}3},{\color{Goldenrod}4}, {\color{Gray}5}, {\color{Magenta}6}, {\color{Orange}7}, {\color{SkyBlue}8}, {\color{Maroon}9}]$. %When indexing in arrays, we use notation similar to Python (e.g., $\mathbf{y}[:,-1]$ denotes all rows of $\mathbf{y}$ for the final column). 

%\jk{You assume a deterministic rule $\delta_i$ for each task. Later, some CausalARC tasks incorporate stochastic $\mathbf{u}$ at the \emph{input} stage but still produce deterministic $\mathbf{y}$ given $\mathbf{x}$. If this is true you may want to consider adding something like: Assumption (Deterministic transduction): $\forall i,\ \mathbf{y}=\delta_i(\mathbf{x})$ so that $p(\mathbf{y}\mid \mathbf{x})$ is a point mass. This may be needed for Proposition~\ref{prop:rule_is_causal} and for treating $(\mathbf{x}_{cf},\mathbf{y}_{cf})$ as labeled supervision.}

%\subsection{Defining \& Measuring AI Reasoning} 

%We need to clearly define what kind of reasoning we are talking about, and how it is measured.

%\begin{definition}[Reasoning]
%    The process of drawing valid logical conclusions about novel phenomena using prior knowledge and new information.
%\end{definition}

%In general, we can measure reasoning by evaluating the correctness of our logical conclusions on novel tasks.

%\begin{definition}[Counterfactual reasoning]    
%\end{definition}

%In this work, measuring counterfactual reasoning: distance($\mathbf{y}_{cf},\widehat{\mathbf{y}}_{cf}$) for some distance metric that goes to zero when $\mathbf{y}_{cf} == \widehat{\mathbf{y}}_{cf}$. \jm{Need to think more about this.}
%loss(cf prediction) during TTT.

%\begin{definition}[Abstract reasoning]    
%\end{definition}

%Measuring abstract reasoning: distance($\mathbf{y}_{test},\widehat{\mathbf{y}}_{test}$) for some distance metric that goes to zero when $\mathbf{y}_{test} == \widehat{\mathbf{y}}_{test}$.

\section{CausalARC: Abstract Reasoning with Causal World Models}
\label{sec:causal_arc}

\subsection{ARC Through a Causal Lens}
\label{sec:arc_causal_lens}

To lay the groundwork for CausalARC, we first outline a causal interpretation of the original ARC dataset and introduce the assumptions underlying our design choices.

\textbf{SCMs: Causal World Models, Causal Programs} \; As in ARC, CausalARC features high task diversity: each task instance is sampled from a unique world model. In program synthesis approaches to ARC, world models are  specified as computer programs (e.g., a Python program) \citep{li2025combining}. The present work assumes a \textit{causal world model} describing the data generating process, which we define in the formal language of SCMs. As the PCH is well-defined given a fully specified SCM \citep{bareinboim2022pch}, this offers an information-rich world model upon which we can perform symbolic operations. 

Classically, we can express SCMs in mathematical notation (e.g., Example \ref{example:scm_xor}). Alternatively, we can draw from the program synthesis view and frame SCMs as \textit{causal programs} that are implemented in a programming language. Under this framing, the act of sampling corresponds to executing the program with different random seeds. Intervening corresponds to principled alterations of the program's logic, while counterfactuals correspond to performing multiple interventions on the same random seed (i.e., the exogenous context is held constant, as required by Definition \ref{def:cf}). The choice to model reasoning tasks as generative causal world models borrows intuition from prior work in reinforcement learning (RL), where causal world models are expressed as Bayesian networks or SCMs \citep{richens2024robust}. Formally, \citet{richens2024robust} prove the conditions under which ``any agent capable of adapting to a sufficiently large set of distributional shifts must have learned a causal model of the data generating process.'' Though we do not explore RL, we follow the intuition that learning new causal world models on the fly supports  generalization under distribution shift. 

\textbf{Representing Random Variables} \; At the highest level of abstraction, it is clear that $\mathbf{x}$ causes $\mathbf{y}$. We can denote this graphically as \circled{\small$\mathbf{x}$} $\to$ \circled{\footnotesize$\mathbf{y}$}, where \circled{\small$\mathbf{x}$} and \circled{\footnotesize$\mathbf{y}$} are multivariate ``supernodes'' in the causal DAG. At finer granularity, we could choose to model each exogenous variable as Bernoulli (Example \ref{example:scm_xor}) or as a categorical random variable that can take on values 0 through 9 (Example \ref{example:scm_colors}). Arbitrary causal functions could result in endogenous random variables with complex distributions. Additionally, an array can be described by grid-level features over all or some elements (e.g., the total number of pixels taking a specific value, the total number of specific sprites, etc.). These can also be modeled as random variables, as in Example \ref{example:scm_colors}. Note that all stochasticity in our model arises from the random sampling of our exogenous variables, and we restrict our attention to deterministic causal functions for endogenous variables. Any $\mathbf{x}$ maps to only one $\mathbf{y}$, just as in the original ARC.

\begin{figure}[!t]
    \centering
        \begin{tabular}{c c c c c}
            \toprule[1pt]
            \multicolumn{2}{c}{\textbf{A.} \textit{Original demonstrations}} & \multicolumn{1}{c}{\textbf{B.} \textit{$\mathbf{x},\mathbf{y} \sim \mathcal{M}_{\texttt{31d5ba1a}}$}} & \textbf{C.} $\mathcal{G}_{\texttt{31d5ba1a}}$ \\
            \cmidrule(lr){1-2} \cmidrule(lr){3-3} \cmidrule(lr){4-4} 
             \includegraphics[width=0.07\textwidth]{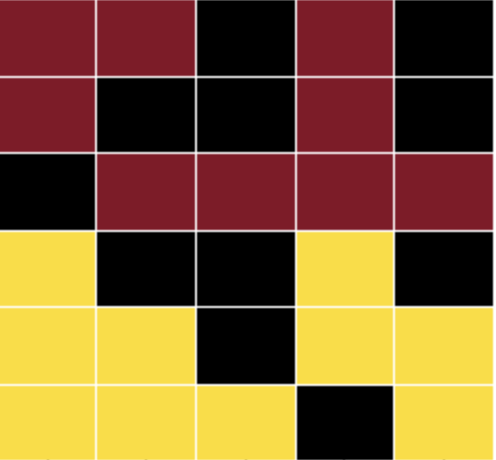} $\to$
            \includegraphics[width=0.07\textwidth]{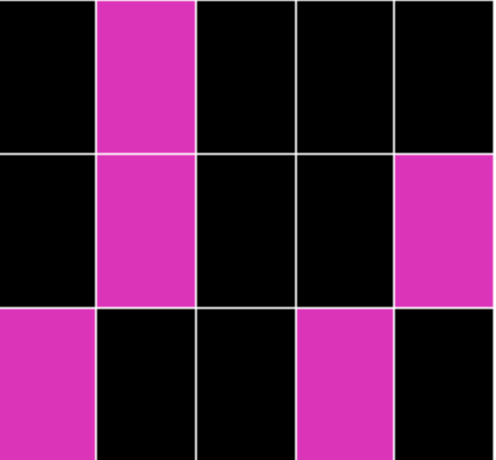} & 
            \includegraphics[width=0.07\textwidth]{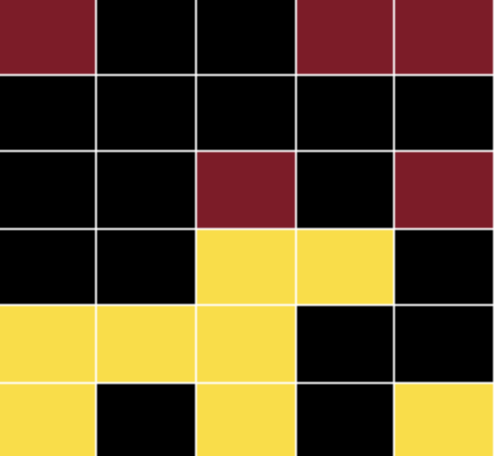} $\to$
            \includegraphics[width=0.07\textwidth]{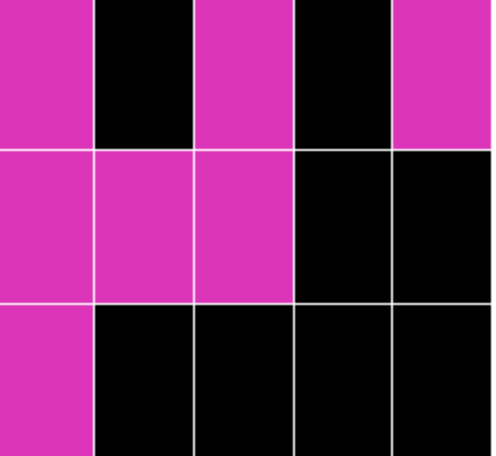} & 
            \includegraphics[width=0.07\textwidth]{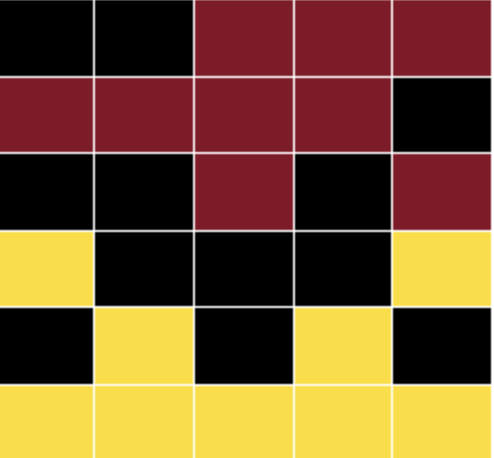} $\to$
            \includegraphics[width=0.07\textwidth]{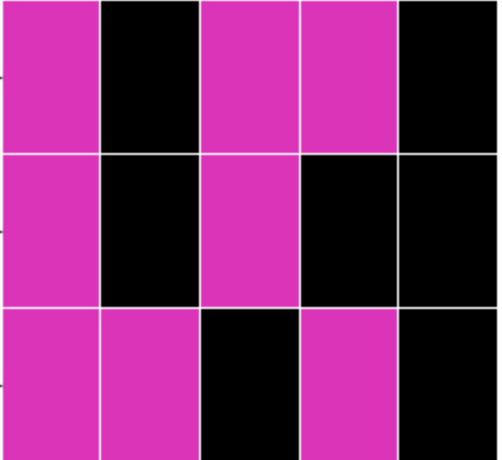} & 
            \begin{tikzpicture}
            \tikzstyle{mystyle}=[
                    circle,
                    minimum size=14mm,
                    draw=black,
                    thick,
                    inner sep=0pt,
                    scale=0.55
                    ]
            % Nodes.
            \node[mystyle, draw=Maroon, fill=Maroon!30] (xij) at (0,0) {$\mathbf{x}[i,j]$} ;
            \node[mystyle, draw=Goldenrod, fill=Goldenrod!30] (xi3j) at (2.5,0) {$\mathbf{x}[i+3,j]$} ;
            \node[mystyle, draw=Magenta, fill=Magenta!30] (ykl) at (1.25,0) {$\mathbf{y}[i,j]$} ;
            % Edges.
            \draw[->,thick] (xij) -- (ykl) ;
            \draw[->,thick] (xi3j) -- (ykl) ;
            \end{tikzpicture}
            \\
            %%%%%%%%%%%%%%%%%%%%%%%%%%
            %%%%%%%%%%%%%%%%%%%%%%%%%%
            \midrule[1pt]
             \multicolumn{2}{c}{\textbf{D.} \textit{Hard interventions on $\mathcal{M}_{\texttt{31d5ba1a}}$}} &  \multicolumn{2}{c}{\textbf{E.} \textit{Soft interventions on $\mathcal{M}_{\texttt{31d5ba1a}}$}}  \\
             \cmidrule(lr){1-2} \cmidrule(lr){3-4}
             $\mathbf{x}[i,j] = 0$ for $i <3$ & $\mathbf{x}[i,j] = {\color{Maroon}9}$ for $i <3$ & ${\color{Maroon}9} \cdot \mathbf{u}[i,j] \; \text{to} \; {\color{Maroon}9} \cdot \neg \mathbf{u}[i,j]$ & 
             ${\color{Maroon}9} \cdot \mathbf{u}[i,j] \; \text{to} \; {\color{NavyBlue}1}\cdot \mathbf{u}[i,j]$ \\
            \includegraphics[width=0.07\textwidth]{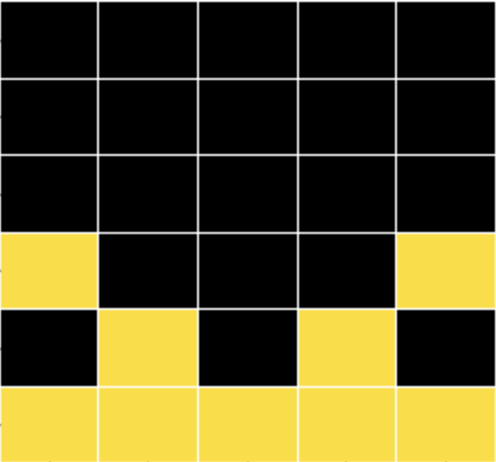} $\to$
            \includegraphics[width=0.07\textwidth]{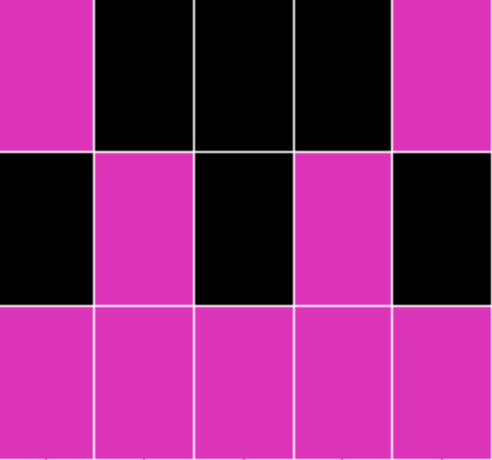} & 
            \includegraphics[width=0.07\textwidth]{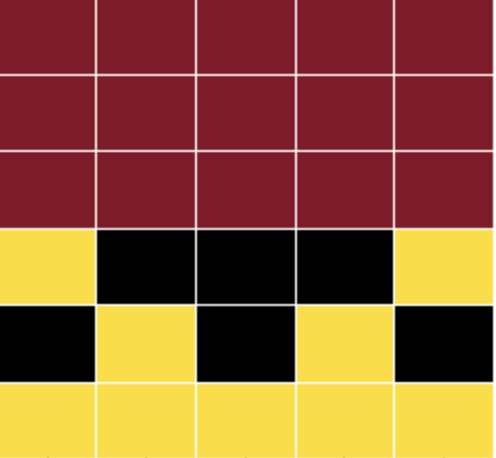} $\to$
            \includegraphics[width=0.07\textwidth]{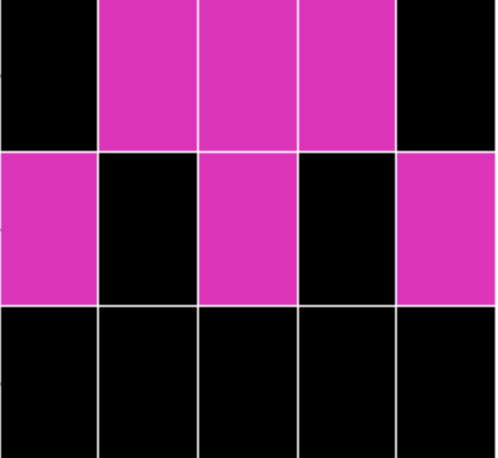} & \includegraphics[width=0.07\textwidth]{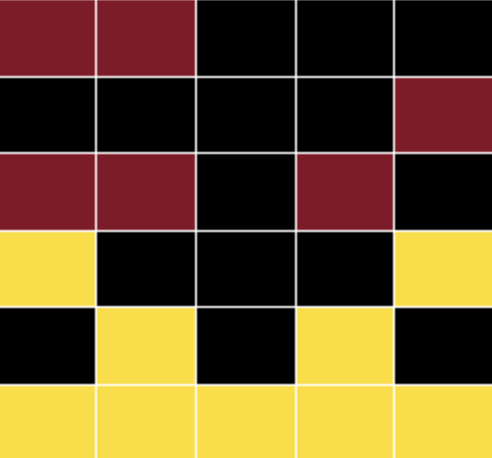} $\to$
            \includegraphics[width=0.07\textwidth]{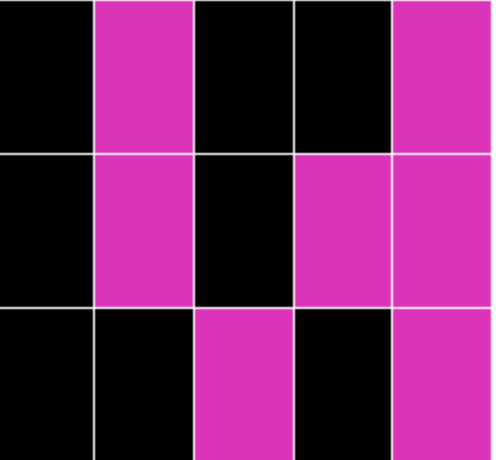} & 
            \includegraphics[width=0.07\textwidth]{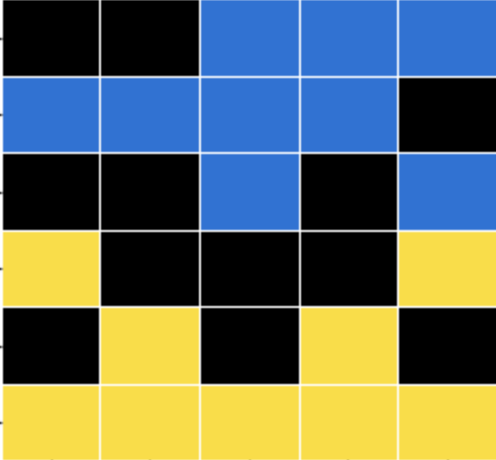} $\to$
            \includegraphics[width=0.07\textwidth]{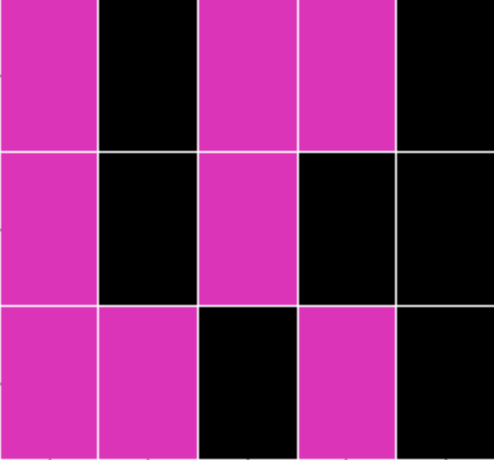} 
           \\
            \bottomrule[1pt]
        \end{tabular}
    %%%%%%%%%%%%%%
    %%%%%%%%%%%%%%
    \caption{Input-output arrays for ARC-AGI-1 task \texttt{31d5ba1a} \citep{chollet2024arc}. (\textbf{A}) A subset of the official demonstration pairs $(\mathbf{x}_{train}, \mathbf{y}_{train})$. (\textbf{B}) A random sample from SCM $\mathcal{M}_{\texttt{31d5ba1a}}$ defined in Example \ref{example:scm_xor}. (\textbf{C}) Causal DAG $\mathcal{G}_{\texttt{31d5ba1a}}$ representing $\mathcal{M}_{\texttt{31d5ba1a}}$, where $\mathbf{y}[i,j] = {\color{Magenta}6} \cdot \mathrm{xor} \left( \mathbf{x}[i,j], \; \mathbf{x}[i+3,j] \right)$ for $i \in [0,2], j \in [0,4]$. (\textbf{D}-\textbf{E}) Samples from interventional submodels of $\mathcal{M}_{\texttt{31d5ba1a}}$, where exogenous variables are held constant and the causal effects of interventions propagate to $\mathbf{y}$. }
    \label{fig:scm_xor}
\end{figure}

\textbf{Grids as Samples from Latent SCMs} \; We assume that an underlying SCM $\mathcal{M}_i$ exists for each $\mathbf{T}_i$. Thus, we assume that every array in $\mathbf{T}_i$ is sampled from the SCM (denoted $\mathbf{T}_i \sim \mathcal{M}_i$). In some original ARC tasks, a unique $\mathcal{M}_i$ can be easily derived from the limited examples provided by $(\mathbf{x}_{train},\mathbf{y}_{train})$. In other cases, multiple %(\textit{causally consistent}  \citep{rubenstein2017causal,scholkopf2021toward}?) 
SCMs might be compatible with the limited information provided by $(\mathbf{x}_{train},\mathbf{y}_{train})$, yielding some form of equivalence class. This latter case is analogous to the causal discovery setting where the full unique DAG is not identifiable, though its Markov equivalence class is \citep{spirtes2000causation}. 
To support the assumption that $\mathbf{T}_i \sim \mathcal{M}_i$, we consider cases where an SCM is easily recoverable from observing $(\mathbf{x}_{train},\mathbf{y}_{train})$ (Examples \ref{example:scm_xor}, \ref{example:scm_colors}).  Samples from the recovered SCMs are indistinguishable from those provided in the official task (Figures \ref{fig:scm_xor}, \ref{fig:scm_colors}).

\begin{example}[A fully recovered SCM] \label{example:scm_xor}
    Consider ARC-AGI-1 task \texttt{31d5ba1a} (expert level \citep{legris2024,legris2025}; Figure \ref{fig:scm_xor}). We can define an SCM $\mathcal{M}_{\texttt{31d5ba1a}} = \langle \mathbf{U}, p(\mathbf{u}), \mathbf{V}, \mathcal{F} \rangle$ where $\mathbf{V}$ are the elements of our arrays, $f \in \mathcal{F}$ are logical $\mathrm{xor}$\footnote{Our notation follows the \texttt{numpy} implementation of logical operators, where 0 = \textsc{false} and any nonzero value is treated as the Boolean 1 = \textsc{true}. All logical operators return integer values in $[0,1]$.} and/or scalar multiplication, and $p(\mathbf{u})$ is Bernoulli.
    \begin{align}
         \mathbf{u}[i,j] &\sim \mathrm{Ber}(0.5) & \text{for} \; i \in [0,5], j \in [0,4] \\
         \mathbf{x}[i,j] &= \begin{cases} {\color{Maroon}9} \cdot \mathbf{u}[i,j] \quad \text{if} \; i < 3 \\ {\color{Goldenrod}4} \cdot \mathbf{u}[i,j] \quad \text{else} \end{cases} & \text{for} \; i \in [0,5], j \in [0,4] \\
         %\mathrm{id}_{\mathbf{u}} \\
        %%
        \mathbf{y}[i,j] &= {\color{Magenta}6} \cdot \mathrm{xor} \left( \mathbf{x}[i,j], \; \mathbf{x}[i+3,j] \right) & \text{for} \; i \in [0,2], j \in [0,4].
    \end{align}
\end{example}

\begin{figure}
    \centering
    \includegraphics[width=0.8\linewidth]{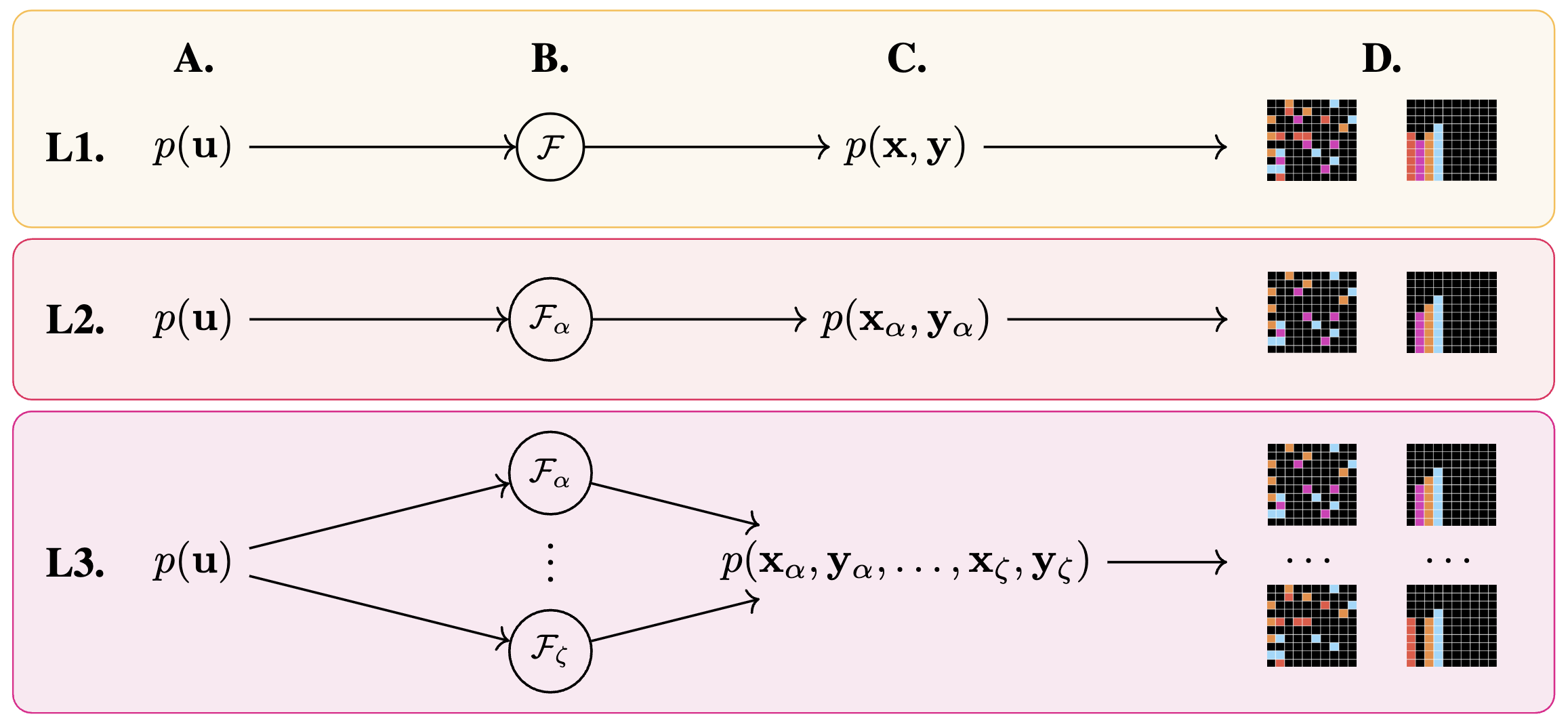}
    \caption{Jointly observed counterfactuals in CausalARC. L1, L2, and L3 denote the rungs of the PCH (Figure \ref{fig:pch}). (\textbf{A}) The distribution over the exogenous context (i.e., the external state). (\textbf{B}) Transformations applied to the exogenous context (e.g., functions $\mathcal{F}$ in the observational world; updated functions $\mathcal{F}_\alpha$ under intervention $\alpha$). (\textbf{C}) Induced distributions, following from the applied transformation. (\textbf{D}) CausalARC samples from each rung of the PCH. Adapted from  \citep{bareinboim2022pch} (Figure 27.2).}
    \label{fig:joint_cfs}
\end{figure}

With our formally defined SCM, we can perform hard or soft interventions and observe the impacts on output $\mathbf{y}$. When we hold $\mathbf{U} = \mathbf{u}$ constant, we claim the resulting arrays are \textit{counterfactuals} (Definition \ref{def:cf}; Figure \ref{fig:joint_cfs}). As shown in Figure \ref{fig:scm_xor}D, hard interventions on $\mathbf{x}$ impact the color distribution over $\mathbf{y}$. In Figure \ref{fig:scm_xor}E, we observe that soft interventions on the causal function for $\mathbf{x}[i,j]$ may or may not impact the output $\mathbf{y}$. As the color of $\mathbf{y}[i,j]$ is impacted by whether $\mathbf{x}[i,j]$ is zero or nonzero but not by the specific nonzero value taken, interventions performed in Figure \ref{fig:scm_xor}E (right) do not influence $\mathbf{y}$.

%Thus, working with the distilled $\delta_i$ is simpler and more intuitive.

\textbf{Rules as Observable Proxies for SCMs} \; In practice, the SCM describing a random ARC task will be too complex to easily derive by hand from limited examples. Thus, we proceed under the assumption that the true SCM $\mathcal{M}_i$ explaining $\mathbf{T}_i$ is unknown. We can choose to model rule $\delta_i$ as an observable and human-interpretable proxy for the latent $\mathcal{M}_i$. In practice,  however, $\delta_i$ is likely to be a lossy distillation of $\mathcal{M}_i$: often, $\mathcal{A}$ (human or machine) only learns as much about $\mathcal{M}_i$ as is needed to output $\widehat{\mathbf{y}}_{test}$ (for example, parameter $p$ to the Bernoulli distribution over the exogenous variables is not generally needed to correctly estimate $\widehat{\mathbf{y}}_{test}$). However, without access to the data generating process provided by $\mathcal{M}_i$, we cannot randomly sample from the true distribution nor guarantee that data augmentations yield true counterfactuals. We introduce CausalARC to address these limitations.

\subsection{Constructing CausalARC}

\textbf{Task Generation} \; All tasks were constructed in Python using built-in functions and \texttt{numpy}. The task construction procedure was as follows. First, an SCM was defined in mathematical notation. This SCM was then manually translated to Python, with functionality for performing hard and/or soft interventions and obtaining sample distributions. Task instances sampled from SCMs are returned as dictionaries in a format consistent with the official ARC dataset\footnote{\href{https://www.kaggle.com/competitions/arc-prize-2024}{https://www.kaggle.com/competitions/arc-prize-2024}} to ensure compatibility with existing ARC pipelines. To date, the static CausalARC dataset is composed of 50 task instances, each with five demonstration pairs, one test pair, and 5–10 counterfactual pairs per demonstration pair.\footnote{See the official project page for Hugging Face and GitHub links: \href{https://jmaasch.github.io/carc/}{https://jmaasch.github.io/carc/}}

\textbf{Causal Annotations} \; In addition to counterfactual examples, each task dictionary is annotated by a human expert with at least one representation of the causal world model: (1) in all cases, a string representation of the Python program is included; (2) in a subset of tasks, the mathematical notation for the SCM is provided; and (3) in another (non-disjoint) subset, the adjacency matrix of the causal graph is provided. Representation (1) is potentially useful for program synthesis settings (e.g., for supervising induction models, as in \citep{li2025combining}), while (3) is especially useful for causal discovery use cases.

\textbf{Task Themes} \; To date, tasks in CausalARC fall into the following human-labeled categories:

\begin{enumerate}[leftmargin=*]
    \item \textit{Counting.} Tasks require elements to be counted (e.g., total pixels of a certain color).
    \item \textit{Extension.} Paths or sprites must be extended according to some rule.
    \item \textit{Logical.} Tasks require reasoning over causal functions that are logical operators.
    \item \textit{Ordering.} Elements must be ordered according to some rule, colored according to some order, etc.
    %\item \textit{Sprite objects.} Tasks require reasoning over the properties of sprites (e.g., location, orientation).
\end{enumerate}

In the static dataset, each theme currently features 10 task instances (with the exception of logical tasks, for which there are 20). As with the concept labels in ConceptARC \citep{moskvichev2023conceptarc} and the difficulty labels provided by \citet{legris2024,legris2025}, task labels allow the user to do fine-grained error analyses. For example, one could compare the performance of test-taker $\mathcal{A}$ on logical tasks when causal functions are \textit{xor} versus \textit{and}, or tasks requiring counting versus ordering, etc.

\textbf{Jointly Observed Counterfactuals} \; The fundamental problem of causal inference is that counterfactuals cannot be jointly observed in the physical universe. In our synthetic grid world, we can exploit the fact that we have full control over the data generating process and exogenous factors can be held constant across multiple interventions. Thus, this synthetic environment allows for the juxtaposition of `treated' samples under multiple interventions alongside their `untreated' counterparts, without violating the assumption that $\mathbf{U} = \mathbf{u}$ in all instances. In this way, we justify our choice to model the data augmentations on $\mathbf{T}_i$ as a set of jointly observed counterfactuals (Figure \ref{fig:joint_cfs}). 

\textbf{Prompt Sampling} \; CausalARC provides functionality for randomly sampling prompts from the underlying SCM according to user specifications. Currently, the following aspects can vary: total number of in-context demonstrations, type of in-context demonstrations (all L1 versus alternating L1/L3), and query theme (counterfactual reasoning, abstract reasoning, program synthesis, and causal discovery). For counterfactual reasoning, the LM is prompted to predict the output under intervention for a previously seen L1 example (e.g., Figures \ref{fig:l1_prompt_cf_reasoning}, \ref{fig:l3_prompt_cf_reasoning}). For abstract reasoning, the LM is prompted to predict the output array for an L1 input, as in conventional ARC setups \citep{akyurek2024surprising}. For program synthesis, the LM is prompted to generate a Python program that expresses the SCM (e.g., Figures \ref{fig:l1_prompt_prog_synth}, \ref{fig:l3_prompt_prog_synth}). For causal discovery, the LM is prompted to predict a property of the underlying SCM, such as the causal parents of an array element or the form of a causal function (e.g., Figures \ref{fig:l1_prompt_discovery}, \ref{fig:l3_prompt_discovery}). 

\begin{figure}[!t]
    \centering
    \begin{tabular}{c/c/c/c}
    \toprule[1pt]
    \textsc{\textbf{a.} counting} & \textsc{\textbf{b.} extension} & \textsc{\textbf{c.} logical} & \textsc{\textbf{d.} ordering} \\
        \includegraphics[height=0.05\textheight]{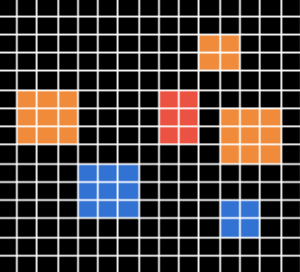} $\to$ \includegraphics[height=0.05\textheight]{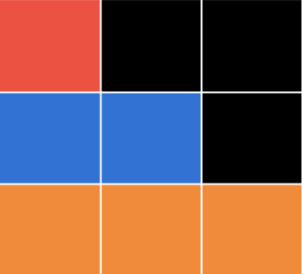} & 
        \includegraphics[height=0.05\textheight]{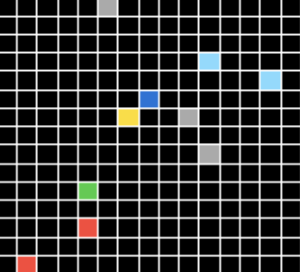} $\to$ \includegraphics[height=0.05\textheight]{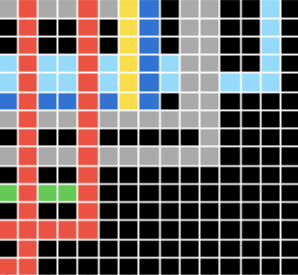} &
        \includegraphics[height=0.05\textheight]{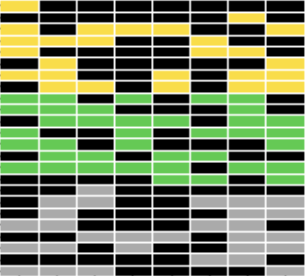} $\to$ \includegraphics[height=0.05\textheight]{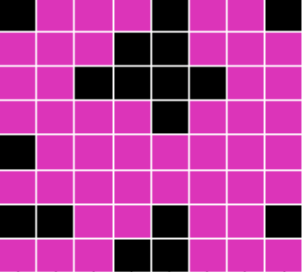} & 
        \includegraphics[height=0.05\textheight]{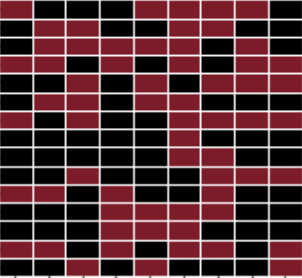} $\to$ \includegraphics[height=0.05\textheight]{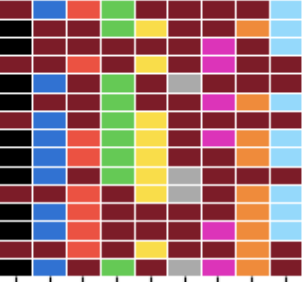} \\
    \bottomrule[1pt]
    \end{tabular}
    \caption{Example demonstration pairs for CausalARC themes.}
    \label{fig:themes}
\end{figure}

\section{Empirical Demonstrations}
\label{sec:results}

As a proof-of-concept, we illustrate four potential use cases for CausalARC: (1) abstract reasoning with TTT, (2) counterfactual reasoning with in-context learning, (3) program synthesis, and (4) causal discovery with logical reasoning. Full experimental details and results are in Appendix \ref{appendix:experimental_design}, including a full description of the tasks used for each experiment. %Results are preliminary and warrant further investigation. %Experiments were designed to showcase a range of behavior (from poor to excellent) to demonstrate the validity of task design.

\textbf{Models} \; Models were selected for diversity. We compare designated reasoning models (e.g., o4-mini) versus vanilla LMs (e.g., GPT-4o mini); older, less powerful models (e.g., Claude Haiku 3.5) versus newer, more powerful models (e.g., Claude Sonnet 4); and  open-source models (e.g., LLama 3 8B, Llama 4 Scout 17B) versus closed-source models (all others). All models were run with default reasoning effort (e.g., default for Grok 4 was high effort).\footnote{See \href{https://jmaasch.github.io/carc/static/pdfs/causal\_arc\_amazon.pdf}{Maasch et al. (2026)} for further explorations of the impacts of reasoning effort on CausalARC performance.} Full model details are in Table \ref{tab:models}.

\textbf{Prompt Formulation} \; We explored impacts of prompt formulation with respect to PCH level and total number of in-context examples. To isolate the effects of PCH level on LM performance, we compared prompts with L1 versus alternating L1/L3 in-context demonstrations while holding the total number of  demonstrations constant (e.g., Figure \ref{fig:l1_prompt_prog_synth} versus \ref{fig:l3_prompt_prog_synth}). 

\textbf{Metrics} \; As in \citet{akyurek2024surprising}, LM performance was measured by output accuracy (i.e., whether the output array was exactly correct) and relative Hamming distance (HD; i.e., the number of positions where the true array differs from the predicted array, normalized by the number of elements in the true array). For HD, a score of 0 indicates correctness while a score of 1 indicates complete failure (including cases where the LM returns a malformed array or no response at all).

%\textbf{Performance Metrics} Percent correct and Hamming distance (define in appendix).

%\textbf{CausalARC Tasks} Each single-operator task category (\textit{and}, \textit{or}, \textit{xor}) featured  $n = 3$ tasks. Each multi-operator task category (alternator or composition) featured  $n = 2$ tasks. From each task, five prompts were sampled per level of the PCH (L1, L3). Each prompt contained a unique permutation of L1 and L3 demonstrations, combined with a counterfactual test case.

\subsection{Abstract Reasoning with Test-Time Training}

\textbf{Motivation} \; To gauge the difficulty of CausalARC, we benchmarked MARC with TTT \citep{akyurek2024surprising} on the full static dataset. %(10 tasks sampled for each theme besides logical, for which there were 20 samples). 
MARC was the second-place paper winner for ARC-AGI-1. It takes a neural transduction approach using a Llama 3 8B base model fine-tuned on large ARC-like datasets, with TTT plus in-context learning at inference. Automated data augmentations (e.g., geometric transformations and color permutations) increase the sample size of the TTT dataset for each task. We employ MARC as it is provided in the public code base, with no modifications.\footnote{\href{https://github.com/ekinakyurek/marc}{https://github.com/ekinakyurek/marc}} 

\begin{figure}[!t]
    \centering
    \includegraphics[height=0.115\textheight]{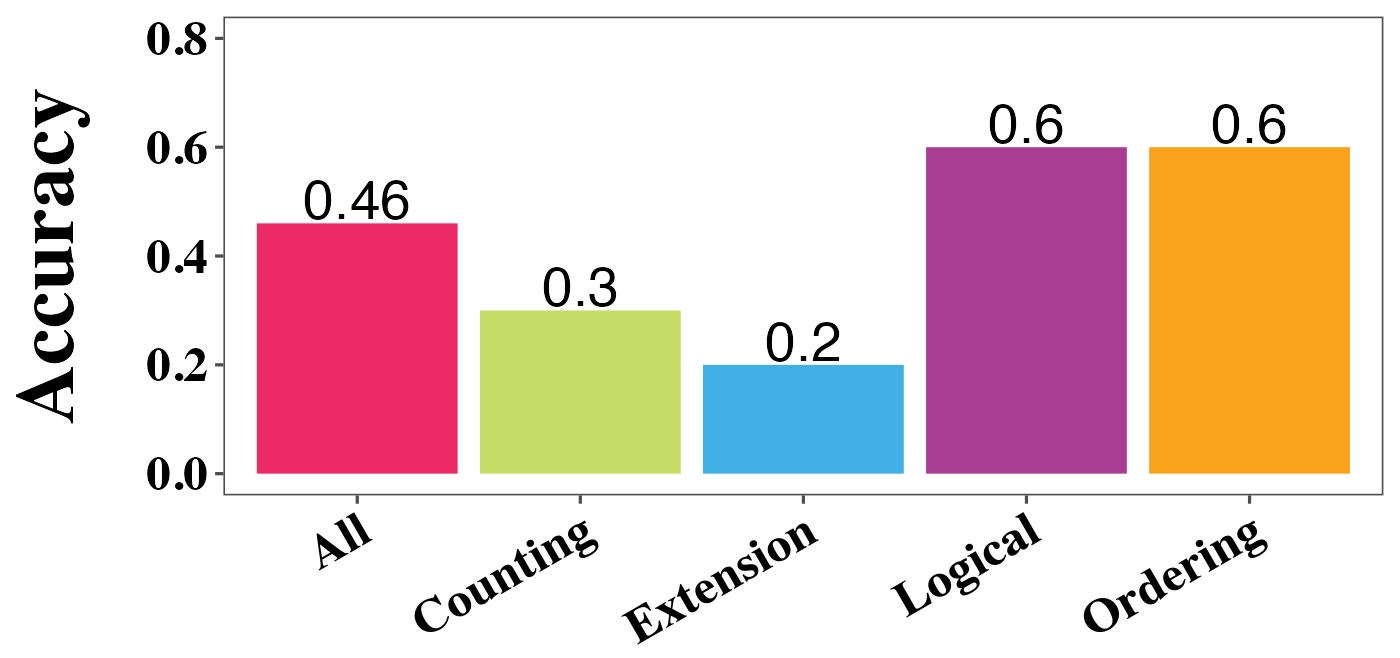} %\hspace{5mm}
    \includegraphics[height=0.115\textheight]{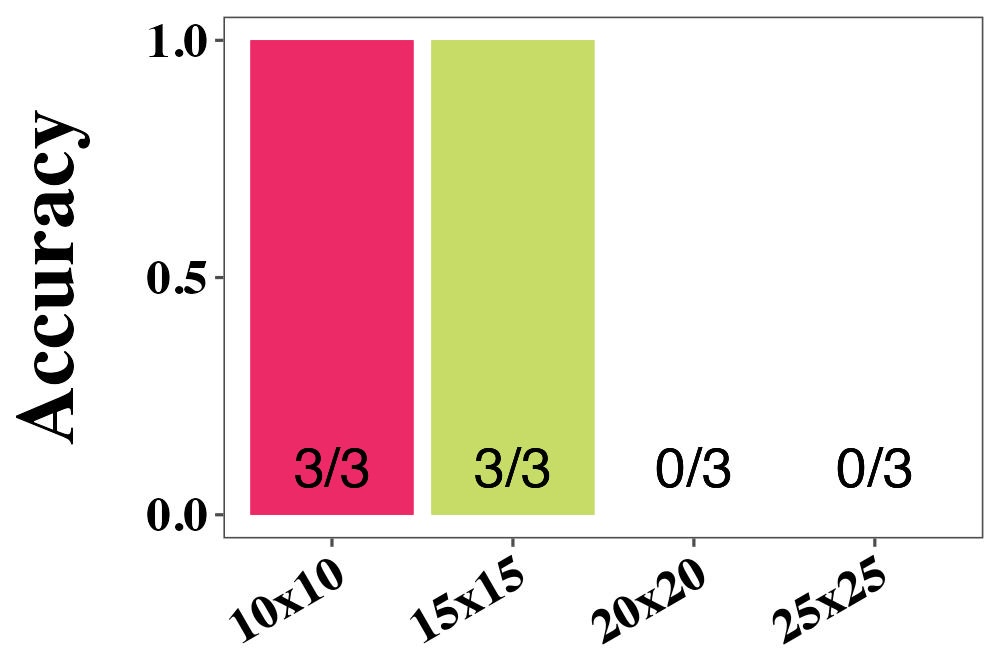}
    \includegraphics[height=0.115\textheight]{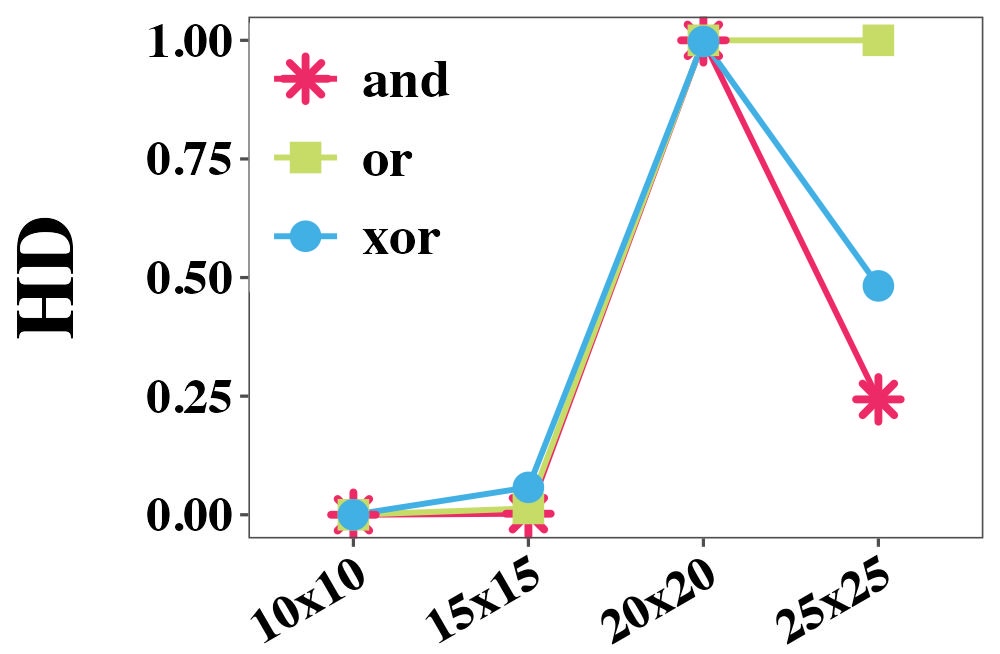}
    \caption{(Left) Accuracy by CausalARC theme for MARC with TTT (Llama 3 8B base) \citep{akyurek2024surprising}. (Center and right) Performance on \textit{and}, \textit{or}, and \textit{xor} tasks sampled from \texttt{SCMdky5} as array size increases.}
    \label{fig:marc_ttt}
\end{figure}

\textbf{Results \& Discussion} \; Overall accuracy on CausalARC was 46\%, with wide variation across themes (Figure \ref{fig:marc_ttt}, left). This is similar to MARC's pure transduction score of 47.1\% accuracy on ARC-AGI-1 \citep{akyurek2024surprising}, suggesting that CausalARC is of comparable difficulty. Extension and counting were significantly more challenging for MARC than logic and ordering, though we cannot establish that this is due to the underlying concepts versus other artifacts of these task distributions. MARC shows signs of struggling with large output arrays on both ARC-AGI-1 and CausalARC, as exemplified by steep dropoffs in performance for logical tasks as array size increases (Figure \ref{fig:marc_ttt}, center).

\subsection{Counterfactual Reasoning with In-Context Learning}

\begin{wrapfigure}[16]{r}{0.45\linewidth}
\vspace{-4mm}
    \centering
    \includegraphics[width=\linewidth]{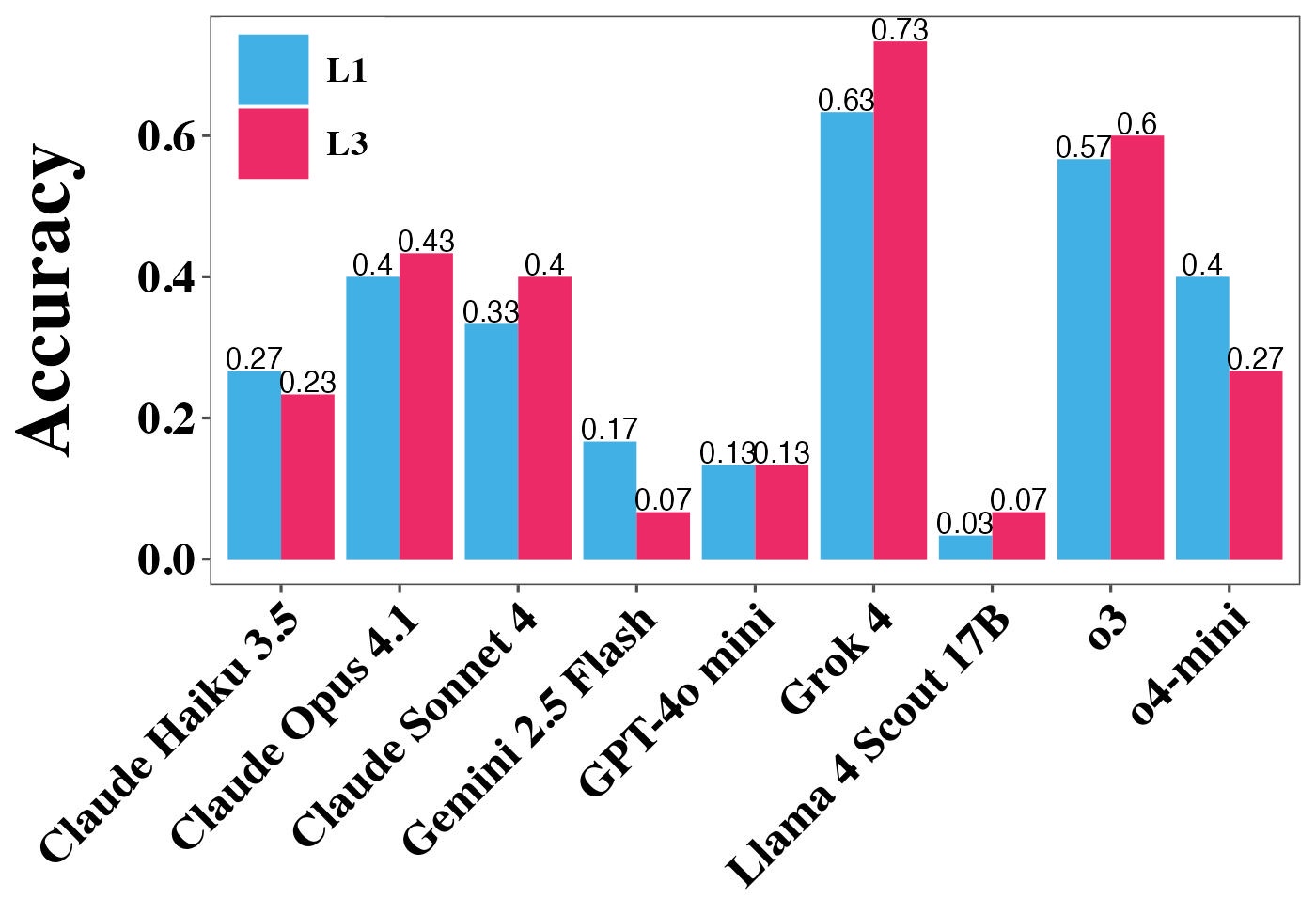} 
    \caption{Counterfactual reasoning on six counting, extension, and ordering tasks. Scores are over five random prompt samples.}
    \label{fig:cf_not_logical}
\end{wrapfigure}

\textbf{Motivation} \; This experiment demonstrates the use of CausalARC for counterfactual reasoning evaluation with few-shot, in-context learning demonstrations. For each task, three in-context demonstrations were presented before a final test case that prompted the LM to predict the counterfactual output for a previously seen L1 demonstration. In addition to measuring the ability of the LM to predict counterfactual samples from the underlying world model, this experiment compares performance when in-context demonstrations are sampled from L1 versus L3.

\textbf{Results \& Discussion} \; Performance varied widely across models (Figures \ref{fig:cf_not_logical}, \ref{fig:cf_not_logical_tasks},  \ref{fig:logical_results}, \ref{fig:logical_results_compose}). L3 demonstrations did not consistently confer benefits. This result could arise from lower demonstration diversity in L3 prompts (as L3 demonstrations are often similar to their L1 counterpart, due to the exogenous context being fixed). %As L3 demonstrations are impossible to obtain under normal (real-world) circumstances, the implications of this preliminary result are not unfortunate. %However, tests should be repeated on additional prompt formulations and greater task diversity.
Strong performance from recent closed-source models on logical tasks (Figure \ref{fig:logical_results}) could be an artifact of fine-tuning on ARC, as these tasks are similar to ARC-AGI-1 task \texttt{31d5ba1a} (which used \textit{xor} functions; Example \ref{example:scm_xor}). %(it cannot be ruled out that strong performance from Gemini 2.5 and Grok 4 on \textit{xor} tasks was due to pretraining on ARC, as \texttt{31d5ba1a} used \textit{xor} functions).

%Grok 4 is advertised as having ``advanced reasoning capabilities.''

\subsection{Program Synthesis}

\textbf{Motivation} \; Winning ARC strategies fall into two dominant camps \citep{chollet2024arc}: (1) neural transduction approaches, where $\widehat{\mathbf{y}}_{test}$ is directly predicted (as in MARC); and (2) neural induction with program synthesis, where the neural network outputs a program implementing $\delta_i$ that is executed to obtain $\widehat{\mathbf{y}}_{test}$. As the first-place ARC-AGI-1 paper winner, BARC \cite{li2025combining} took a meta-learning approach \citep{mishra2018simple} that combined neural transduction with Python program synthesis, demonstrating improved performance relative to transduction or induction alone. MARC saw significant performance gains when ensembling with BARC, further supporting the combined use of induction and transduction \citep{akyurek2024surprising}. Given the success and community interest surrounding program synthesis for reasoning, we also apply CausalARC for this use case. %LMs were prompted to return a Python program for generating output $\widehat{\mathbf{y}}$ from input $\mathbf{x}$, which captures the underlying causal relationships (Figures \ref{fig:l1_prompt_prog_synth}, \ref{fig:l3_prompt_prog_synth}). 
Accuracy was measured with respect to the output $\widehat{\mathbf{y}}_{test}$ generated by passing a test input $\mathbf{x}_{test}$ to the LM-generated program.

\textbf{Results \& Discussion} \; Total in-context examples had a notable impact on program correctness (Figures \ref{fig:prog_synth_discovery_in_text},  \ref{fig:prog_synth_bar}). Maximum performance was attained by o4-mini, which showed possible signs of fine-tuning for ARC-like program synthesis (Figure \ref{fig:o4_fine_tuning}). Accuracy peaked at six in-context examples for all models except Sonnet 4 (which saw monotonic increases) and Llama 4 Scout 17B (which failed to output any correct programs). L3 prompts rarely conferred benefits (Figure \ref{fig:prog_synth_bar}).

\subsection{Causal Discovery with Logical Reasoning}

\textbf{Motivation} \; Causal discovery is the body of theory and methods concerned with learning causal graphical structures from data \citep{glymour2019review}. %As causal effect estimation generally assumes at least partial prior knowledge of the underlying causal graph, structure learning is an important prerequisite for much of causal inference. 
Though there is increasing interest in using LMs to support causal discovery, recent evaluations show steep shortcomings \citep{tu2023causal,jin2023can,gao2023chatgpt}. As CausalARC tasks are annotated with SCMs and (in a subset) the corresponding adjacency matrix, CausalARC is conducive to benchmarking LMs on causal discovery tasks. As a proof-of-concept, we tested the performance of state-of-the-art LMs on causal discovery with respect to scaling problem complexity (i.e., array size). Rather than directly asking for adjacencies in the causal graph, LMs were asked to predict the logical operators acting as causal functions (which implies knowledge of the causal parents of each output cell, and requires logical reasoning). Example prompts are provided in Figures \ref{fig:l1_prompt_discovery} and \ref{fig:l3_prompt_discovery}. Future work could test the ability of models to explicitly list parent-child relationships, or to return adjacency matrices.

%\begin{wrapfigure}[35]{r}{0.45\linewidth}
\begin{figure}[!t]

    \centering
\includegraphics[width=0.45\linewidth]{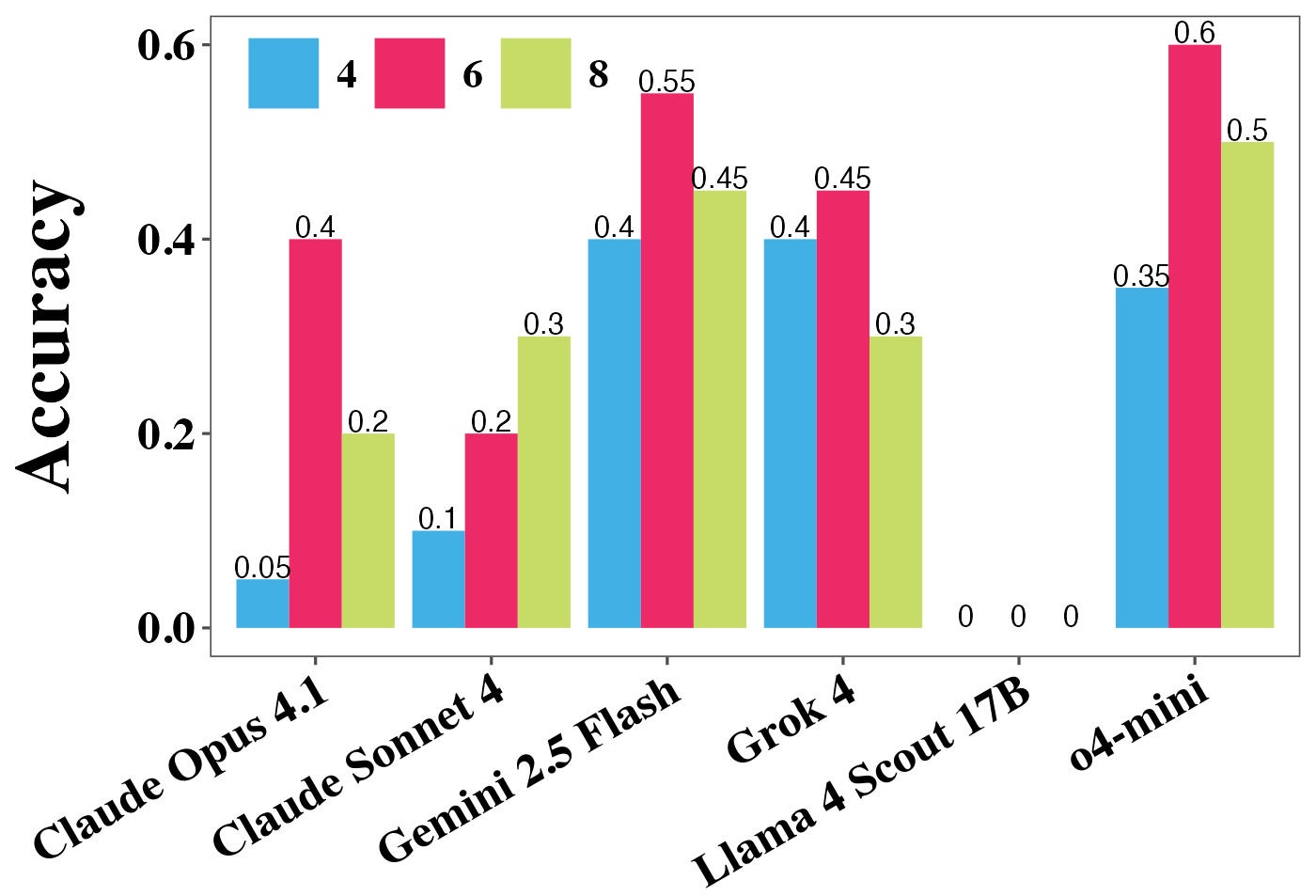} \hspace{5mm}
    \includegraphics[width=0.45\linewidth]{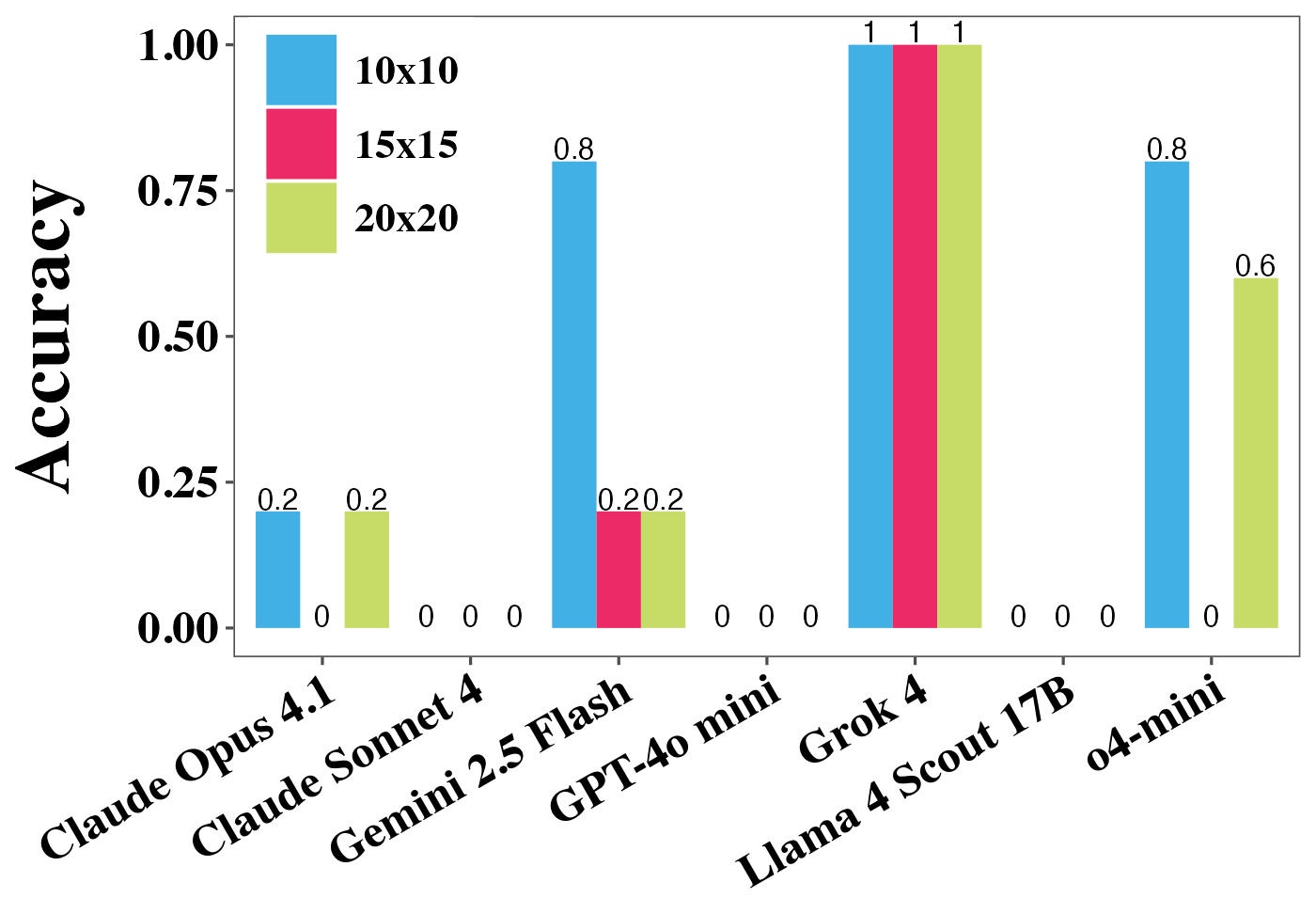} 
    \caption{(Left) Program synthesis on four counting and extension tasks as total in-context demonstrations increased. (Right) Causal discovery with logical reasoning as array size increased. Scores were over five random L1 prompt samples. Results for L3 prompts are in Appendix \ref{appendix:experimental_results}}
    \label{fig:prog_synth_discovery_in_text}
\end{figure}
%\end{wrapfigure}

\textbf{Results \& Discussion} \; Performance varied widely across models (Figures \ref{fig:prog_synth_discovery_in_text}, \ref{fig:discovery_logical}). Grok 4 consistently outperformed, likely related to the default reasoning effort set to high.\footnote{See \href{https://jmaasch.github.io/carc/static/pdfs/causal\_arc\_amazon.pdf}{Maasch et al. (2026)} for further explorations of the impacts of reasoning effort on CausalARC performance.} All other models displayed unstable performance, as monotonic declines in accuracy were not consistently observed. L3 prompts did not consistently confer benefits both within and across models, though benefits were greater than for program synthesis (Figure \ref{fig:discovery_logical}). Correct responses varied in the extent to which they explicitly enumerated parent-child relationships (Figures \ref{fig:gemini_response_discovery}, \ref{fig:grok_response_discovery}). Future work could query LMs directly for adjacencies, using causal discovery algorithms as baselines. As a proof-of-concept, we ran the popular constraint-based PC\footnote{We selected PC as it is one of the most widely cited baseline methods and has multiple open-source implementations. Additionally, the assumptions of this method are met by our data generating processes, unlike causal discovery methods that assume additive noise models (e.g., LiNGAM \citep{shimizu2006linear, shimizu2014lingam}, nonlinear ANMs \citep{hoyer2008nonlinear}).} algorithm \citep{spirtes2000causation} on L1 distributions sampled from \texttt{SCMdky5} (Table \ref{tab:pc_results}, Figure \ref{fig:pc_results}). Poor performance of PC on samples with \textit{xor} causal function is likely due to a violation of faithfulness: the common assumption that statistical independencies correspond to separations in the underlying graph. Logical \textit{xor} and deterministic functions both famously cause faithfulness violations, which can significantly undermine causal discovery \citep{marx2021weaker}. This may suggest that CausalARC could be extended for testing robustness to faithfulness violations in discrete data, a problem previously explored for continuous data \citep{olko2025faithfulness}.

\section{Conclusion}
\label{sec:conclusion}

This work introduces CausalARC as an open-ended AI reasoning testbed that accommodates diverse experimental setups, including TTT and in-context learning for counterfactual, abstract, and logical reasoning. Results presented in this proof-of-concept are preliminary and require further investigation.

\textbf{Benchmarking Takeaways} \; Within- and between-model performance varied heavily across tasks. Unsurprisingly, there were some signs that state-of-the-art proprietary models benefited from fine-tuning on ARC and its extensions. For example, o4-mini offered ``ARC-style input/output'' in some program synthesis responses, despite ARC never being mentioned in the prompt; Claude Sonnet 4 volunteered to perform program synthesis on counterfactual reasoning tasks without being prompted to do so; and Grok 4 displayed exceptional performance on \textit{xor} logical problems, which are known to exist in ARC-AGI-1. Additionally, model reasoning effort (i.e., inference-time scaling) had significant impacts on benchmark performance, as further explored in \href{https://jmaasch.github.io/carc/static/pdfs/causal\_arc\_amazon.pdf}{Maasch et al. (2026)}. Despite this, we do not observe consistent evidence of saturation on CausalARC in state-of-the-art models. Results suggest that there is significant room for LM reasoning to improve in the areas tested.

\textbf{Limitations \& Future Directions} \; While the manual design of this dataset allows us to maintain strict causal assumptions, this also inhibits scaling and prevents use for fine-tuning and large-scale benchmarking. Unlike ARC \citep{chollet2024arc} and ConceptARC \citep{moskvichev2023conceptarc}, CausalARC was never tested on humans. However, results from MARC with TTT \citep{akyurek2024surprising} suggest that CausalARC is of similar difficulty to ARC-AGI-1. To address these shortcomings, future work could expand CausalARC with additional tasks, task themes, and human performance data.

\section*{Acknowledgments}

Author J. Maasch acknowledges the US National Science
Foundation Graduate Research Fellowship under Grant No.
DGE–2139899.

\section*{Impact Statement}

The possibility of AI reasoning emergence has broad scientific, economic, and social implications, including matters of safety and fairness. Though this work aims to promote rigorous reasoning evaluation, strong performance on CausalARC is \textit{necessary but not sufficient} for demonstrating that LMs can reason. Results should be interpreted with great caution when deploying LMs for reasoning tasks that could have safety or fairness consequences.

%\section*{References}

\bibliography{references}

%%%%%%%%%%%%%%%%%%%%%%%%%%%%%%%%%%%%%%%%%%%%%%%%%%%%%%%%%%%%

\clearpage

\appendix
\setcounter{figure}{0} % this has to come before the next command renewals, otherwise appendix links will not work.
\setcounter{table}{0}
\renewcommand\thefigure{\thesection.\arabic{figure}}  
\renewcommand\thetable{\thesection.\arabic{table}}

\addcontentsline{toc}{section}{Appendix} % Add the appendix text to the document TOC
\part{Appendix} % Start the appendix part
\parttoc % Insert the appendix TOC

\section{Extended Preliminaries}
\label{appendix:prelims}

\subsection{Defining and Measuring Reasoning}
\label{appendix:reasoning}

% We would expect an intelligent AI reasoner to display flexible adaptation to novel problems from limited demonstration.

Any framework claiming to measure reasoning should clearly define its target of measure.  Reasoning is a famously difficult concept to concretely define and has taken on many meanings across time and research domains. To begin, we first declare our chosen definitions of \textit{generalization} and \textit{intelligence}. We take a definition of generalization that requires adaptability in novel settings. 

\begin{definition}[Generalization, \citet{chollet2019measure}]
\label{def:generalization}
    The ability to handle scenarios or tasks that differ from previously encountered situations, demonstrating both \textit{robustness} and \textit{flexibility}.
\end{definition}

\begin{definition}[Robustness, \citet{chollet2019measure}]
\label{def:robustness}
    Adaptation to \textit{known unknowns} within a single task or well-defined set of tasks.
\end{definition}
 
\begin{definition}[Flexibility, \citet{chollet2019measure}]
\label{def:flexibility}
    Adaptation to \textit{unknown unknowns} across a broad category of related tasks.
\end{definition}

In CausalARC, an example of robustness would be high accuracy on a large number of test cases sampled from one causal world model on which the test-taker has undergone TTT. An example of flexibility would be high accuracy on a large number of unique causal world models. 

We follow \citet{chollet2019measure} in  defining the intelligence of a system as a measure of \textit{skill-acquisition efficiency}, rather than skill itself. This is consistent with the conventional wisdom that human intelligence tests should measure cognitive capacities in a general sense, and not task-specific skills acquired through experience or practice \citep{chollet2019measure}.

\begin{definition}[Intelligence, \citet{chollet2019measure}]
    Skill-acquisition efficiency over a range of tasks, controlling for priors, experience, and generalization difficulty.
\end{definition}

Thus, we take a \textit{general intelligence} to be one with high skill acquisition efficiency and high generalization over a large range of tasks. To date, human-like general intelligence has yet to be achieved in AI. We emphasize the need to control for experience and priors, as ``unlimited priors or experience can produce systems with little-to-no generalization power (or intelligence) that exhibit high skill at any number of tasks'' \citep{chollet2019measure}. This latter case (where skill outpaces generalization) is reminiscent of current shortcomings in large LM reasoning \citep{mirzadeh2025gsm,shojaee2025illusion,xu2025reimagine}, where pretraining on web-scale corpora and extensive fine-tuning could be seen as nearly ``unlimited priors or experience.''

Finally, we define our notions of \textit{reasoning}: cognitive processes that are core attributes of intelligent thinking. We focus on abstract, logical, and counterfactual reasoning, as these are the forms tested in the current iteration of CausalARC. This definition of abstract reasoning draws from the ARC Prize.\footnote{\href{https://arcprize.org/arc-agi}{https://arcprize.org/arc-agi}}

\begin{definition}[Abstract reasoning]
\label{def:abstract_reasoning}
    The process of drawing valid conclusions about novel visual information by identifying patterns using innate cognitive priors, rather than the accumulation of concrete, domain-specific knowledge.
\end{definition}

Here, validity is measured by correctness (e.g., accuracy). Innate cognitive priors are primitives acquired at birth or early development under minimal supervision, broadly pertaining to concepts like objects, actions, number, and space (e.g., a grasp of intuitive physics, such as gravity; basic numeracy; basic geometric operations, such as rotation; a sense of relative size or magnitude, etc.) \citep{spelke2007core}.

\begin{definition}[Logical reasoning]
\label{def:logical_reasoning}
    The process of drawing valid conclusions by applying formal logical rules to novel information.
\end{definition}

Logical reasoning encompasses many subtypes, including deductive and inductive. In this work, we primarily focus on the process of applying logical operators in Boolean algebra. 

Next, recall the definition of a counterfactual: \textit{the counterfactual $Y_x$ under model $\mathcal{M}$ is given by $Y_x(\mathbf{u}) \coloneqq Y_{\mathcal{M}_x}(\mathbf{u})$} (Definition \ref{def:cf}). We define counterfactual reasoning as follows.

\begin{definition}[Counterfactual reasoning]
\label{def:cf_reasoning}
    The process of constructing valid answers to ``what if'' questions about imagined alternate outcomes in novel settings, as in: ``What value would $Y$ have taken had $X$ been $x$ in context $\mathbf{U} = \mathbf{u}$?''
\end{definition}

Then, the \textit{ability to reason} abstractly, logically, or counterfactually is the ability to correctly execute these reasoning processes with robust and flexible generalization. Benchmarks like ARC and CausalARC are designed to measure signs of this ability by presenting the test-taker with diverse and novel problems while controlling for experience and priors. 
%An ideal reasoning evaluation framework should  disentangle cases of true general reasoning from high skill with limited generalization. ARC and CausalARC both address this issue by requiring that each task is ``novel (measuring the ability to understand a new task, rather than skill)'' and assuming a shared set of priors common to all test-takers \citep{chollet2019measure}.
Nevertheless, these benchmarks measure the end-point of reasoning: i.e., the extent to which the final output is correct. Under Definitions \ref{def:abstract_reasoning}, \ref{def:logical_reasoning}, and \ref{def:cf_reasoning}, reasoning is a \textit{process} and not an end-product. Thus, there is increasing interest in benchmarks that also assess the intermediate steps of AI reasoning \citep{shojaee2025illusion}. Additionally, fairly controlling for experience and priors is challenging in the context of large LMs that do not release details of their pretraining and fine-tuning, and it is known that state-of-the-art LMs generally have prior knowledge of ARC. These are some reasons (among many) why we argue that strong performance on CausalARC and analogous benchmarks is \textit{necessary but not sufficient} for demonstrating that LMs can reason.

\subsection{World Models}
\label{appendix:world_models}

Emergent behaviors in large LMs have raised new questions on whether learning accurate world models can improve AI reasoning \citep{hao2023llmworld}. Notably, world models have been variously defined across literatures \citep{ha2018world}. In model-based reinforcement learning (RL), a world model is often a predictor of the evolution of an environment under arbitrary policies (e.g., an approximation of the transition function of a Markov decision process) \citep{richens2024robust}. While the  usefulness of world models in RL has been a matter of significant debate \citep{huang2020model}, the necessity of world models has been proven for multiple settings. Given a sufficiently diverse set of \textit{goal-directed tasks}, \citet{richens2025general} prove the conditions under which any agent satisfying a regret bound for these tasks will necessarily have learned a \textit{predictive world model} that converges to the true model as the agent approaches optimality. Given regret-bounded policies for a sufficiently large set of \textit{distributional shifts}, \citet{richens2024robust} prove the conditions under which the agent necessarily learns an approximate \textit{causal world model} that converges to the true causal model under optimal policies. \citet{richens2024robust} express causal world models as causal Bayesian networks (DAGs), though their results extend to fully specified SCMs. While the present work does not explore CausalARC in the context of RL, we are motivated by the intuition provided by \citep{richens2024robust,richens2025general} that world models can facilitate (and can even be necessary for) optimal decision-making, planning, and reasoning in AI.  

\clearpage

\section{ARC Through a Causal Lens: Extended Discussion}

\begin{figure}[!h]
    \centering
    \begin{minipage}{0.6\textwidth}
    \begin{center}
        \begin{tabular}{c}
            \toprule[1pt]
             \textbf{A.} \textit{Original demonstrations}  \\ 
             \midrule
             \includegraphics[height=0.06\textheight]{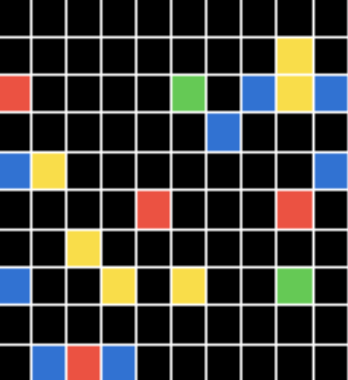} $\to$
            \includegraphics[height=0.06\textheight]{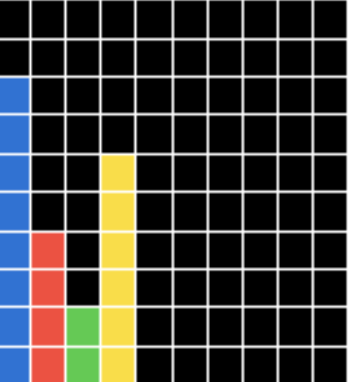} \hspace{5mm}
            \includegraphics[height=0.06\textheight]{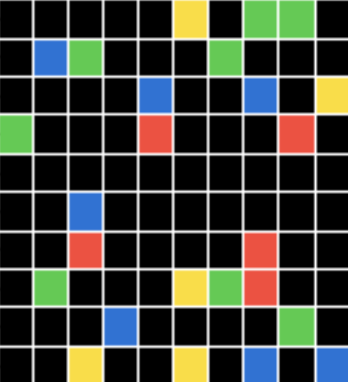} $\to$
            \includegraphics[height=0.06\textheight]{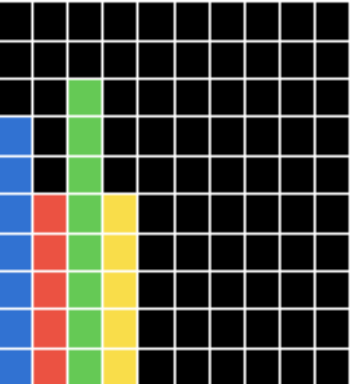} \\
            \midrule[1pt]
            %%%%%%%%%%%%%%%%%%%%%%%%%%
            \textbf{B.} $\mathbf{x},\mathbf{y} \sim \mathcal{M}_{\texttt{f3cdc58f}}$  \\
            \midrule
            \includegraphics[height=0.06\textheight]{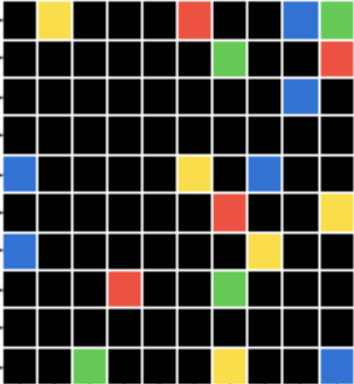} $\to$
            \includegraphics[height=0.06\textheight]{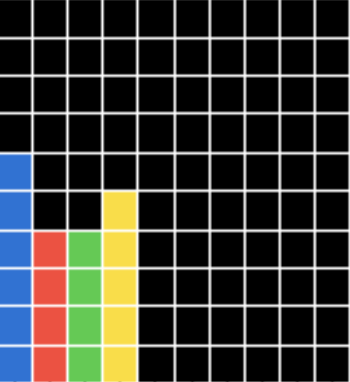} \hspace{5mm}
            \includegraphics[height=0.06\textheight]{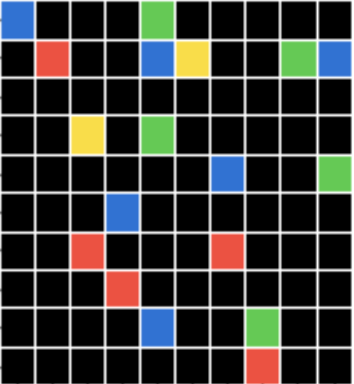} $\to$
            \includegraphics[height=0.06\textheight]{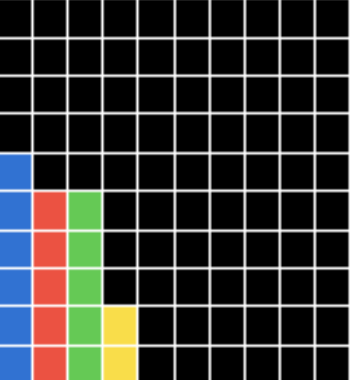} \\
            \bottomrule[1pt]
        \end{tabular}
    \end{center}
    \end{minipage}\hspace{3mm}
    %%%
    \begin{minipage}{0.3\textwidth}
    \begin{center}
        \begin{tikzpicture}
        
            \tikzstyle{mystyle}=[
                    circle,
                    minimum size=8mm,
                    draw=NavyBlue,
                    thick,
                    inner sep=0pt,
                    scale=0.8
                    ]

            % Define a custom horizontal shading with 4 colors.
              \pgfdeclarehorizontalshading{fourcolor}{2cm}{
                color(0cm)=(Red!20);
                color(0.75cm)=(Goldenrod!20);
                color(1.25cm)=(Green!20);
                color(2cm)=(NavyBlue!20)
              }
            
            % Nodes.  
            \node[mystyle,shading=fourcolor,shading angle=45] (x) at (2,1.5) {$\mathbf{X}$} ;
            \node[mystyle,shading=fourcolor,shading angle=45] (c) at (5,1.5) {$\mathbf{C}$} ;
            \node[mystyle,shading=fourcolor,shading angle=45] (y) at (3.5,0) {$\mathbf{Y}$} ;
        
            % Edges.
            \draw[->,thick] (x) -- (c) ;
            %\draw[->,thick] (x) -- (y) ;
            \draw[->,thick] (c) -- (y) ;

            % Label.
            \node[] (label) at (3.5,-1) {\textbf{C.} Causal graph $\mathcal{G}_{\texttt{f3cdc58f}}$} ;
        
        \end{tikzpicture}
    \end{center}
    \end{minipage}
    \caption{Input-output arrays for ARC-AGI-1 task \texttt{f3cdc58f}. (\textbf{A}) A subset of official demonstration pairs $(\mathbf{x}_{train},  \mathbf{y}_{train})$. (\textbf{B}) Randomly sampled pairs from the SCM defined in Example \ref{example:scm_colors}. (\textbf{C}) A DAG representation for task \texttt{f3cdc58f}, where $\mathbf{c}$ is an array-level feature over $\mathbf{x}$.} %\jm{Could intervene on $\mathbf{c}[n]$ to illustrate intervention that alters graph.}}
    \label{fig:scm_colors}
\end{figure}

We present another example of an ARC-AGI-1 task where the SCM is easily recoverable by observing $(\mathbf{x}_{train}, \mathbf{y}_{train})$. In this case, the SCM includes an endogenous variable that is an array-level feature (count of total pixels per color). Note that when indexing in arrays, we use notation similar to Python (e.g., $\mathbf{y}[:,-1]$ denotes all rows of $\mathbf{y}$ for the final column). 

\begin{example}[A fully recovered SCM with array-level features] \label{example:scm_colors}
Consider ARC-AGI-1 task \texttt{f3cdc58f} (easy level \citep{legris2024,legris2025}; Figure \ref{fig:scm_colors}). We can define an SCM $\mathcal{M}_{\texttt{f3cdc58f}} = \langle \mathbf{U}, p(\mathbf{u}), \mathbf{V}, \mathcal{F} \rangle$ where $\mathbf{V}$ are the elements of our arrays, $f \in \mathcal{F}$ are expressed in Equations \ref{eq:x_f3cdc58f}–\ref{eq:y_f3cdc58f}, $p(\mathbf{u})$ is categorical with support $\mathcal{X}$, and endogenous variables in output array $\mathbf{y}$ are causal children of an array-level feature over $\mathbf{x}$ (the total number of elements in $\mathbf{x}$ that are a certain color).
\begin{align}
     \mathbf{u} \sim \mathrm{Cat}(\mathrm{len}(\mathcal{X}), p) \;\;\; \text{for} \; \mathcal{X} &= [0,{\color{NavyBlue}1},{\color{Red}2},{\color{Green}3},{\color{Goldenrod}4}], \; p = [0.8, 0.05,  0.05, 0.05, 0.05] \\
    \mathbf{x} &= \mathrm{id}_{\mathbf{u}} \label{eq:x_f3cdc58f}\\
    \mathbf{c} &= [\mathrm{count}(\mathbf{x} == j)]_{j \in \mathcal{X}} \\
    \mathbf{y}[-\mathbf{c}[j]:,j-1] &= \begin{cases} j \quad \text{if} \; j \in \mathcal{X} \setminus \{0\} \\ 0 \quad \text{else.} \end{cases} \label{eq:y_f3cdc58f}
\end{align}
\end{example}

\section{Extended Empirics}
\label{appendix:experimental_design}

\subsection{Experimental Details}

\textbf{Compute Resources} \; All LM experiments were run on an AWS EC2 g6e.xlarge instance featuring one GPU with 48GB GPU memory and four vCPUs.\footnote{\href{https://aws.amazon.com/ec2/instance-types/g6e/}{https://aws.amazon.com/ec2/instance-types/g6e/}} All other experiments were run locally on a MacBook Pro (Apple M2 Pro chip, 16 GB memory).

\textbf{Models} \;  Benchmarking experiments used the models described in Table \ref{tab:models}. All models used default hyperparameters and temperature 0.0 (except o4-mini, which required temperature 1.0). Models were queried using the \texttt{langchain} Python package.\footnote{\href{https://python.langchain.com/}{https://python.langchain.com/}}

\begin{table}[!h]
    \centering
    \begin{adjustbox}{max width=\textwidth}
    \begin{tabular}{l l l l }
    \toprule[1pt]
       \textsc{provider} & \textsc{model} & \textsc{api id} & \textsc{url} \\
    \midrule[1pt]
       \textit{OpenAI} & GPT-4o mini & gpt-4o-mini-2024-07-18 & \href{https://platform.openai.com/docs/models}{https://platform.openai.com/docs/models} \\
       & o4-mini  & o4-mini-2025-04-16 \\ 
       \midrule
       \textit{Anthropic} & Claude Haiku 3.5 & claude-3-5-haiku-20241022 & \href{https://docs.anthropic.com/en/docs/about-claude/models/}{https://docs.anthropic.com/en/docs/about-claude/models/} \\
       & Claude Sonnet 4 &  claude-sonnet-4-20250514 \\
       \midrule
       \textit{Google} & Genini 2.5 Flash & gemini-2.5-flash & \href{https://ai.google.dev/gemini-api/docs/models}{https://ai.google.dev/gemini-api/docs/models}\\
       \midrule
       \textit{XAI} & Grok 4 & grok-4 & \href{https://docs.x.ai/docs/models}{https://docs.x.ai/docs/models} \\
       \midrule
       \textit{Meta (Bedrock)} & Llama 4 Scout 17B & us.meta.llama4-scout-17b-instruct-v1:0 & \href{https://llama.developer.meta.com/docs/models/}{https://llama.developer.meta.com/docs/models/} \\
    \bottomrule[1pt]
    \end{tabular}
    \end{adjustbox}
    \vspace{2mm}
    \caption{Models used in benchmarking experiments.}
    \label{tab:models}
\end{table}

\textbf{Prompt Formulation} \; Several experiments  compare the impacts of L1 versus L3 in-context demonstrations on LM performance. To prevent the conflation of benefits from PCH level versus query length, the total number of in-context demonstrations was held constant across L1 and L3 prompts. Headers for L3 demonstrations cause a small increase in prompt length, as these are slightly longer than L1 example headers. For example, the L1 prompt in Figure \ref{fig:l1_prompt_cf_reasoning} is 4130 characters long, while the L3 prompt in Figure \ref{fig:l3_prompt_cf_reasoning} is 4267 characters. In this case, including L3 demonstrations resulted in a 3.2\% length increase.

\subsubsection{Counterfactual Reasoning with In-Context Learning}

\textbf{Tasks} \; Counterfactual reasoning experiments used task instances sampled from all task themes. Counting tasks were \texttt{SCMfuy3} and \texttt{SCMm5ob}. Extending tasks were \texttt{SCMz750} and \texttt{SCMwoev}. Ordering tasks were \texttt{SCMffb8} and \texttt{SCMtzlq}. All tasks with tunable array sizes (all except \texttt{SCMfuy3} and \texttt{SCMffb8}) used size $25 \times 25$.

\begin{figure}[!h]
    \centering
    \begin{tabular}{c/c/c}
    \toprule[1pt]
    \textsc{logical and} & \textsc{logical or} & \textsc{logical xor} \\
        \includegraphics[height=0.04\textheight]{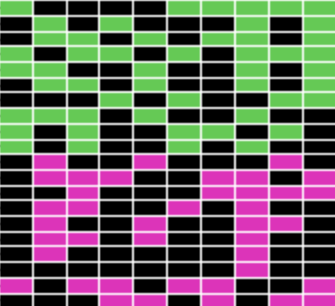} $\to$ \includegraphics[height=0.04\textheight]{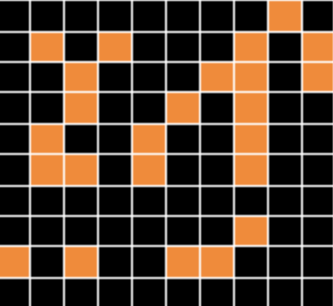} & 
        \includegraphics[height=0.04\textheight]{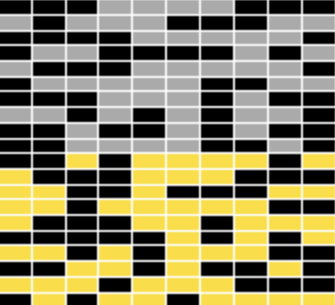} $\to$ \includegraphics[height=0.04\textheight]{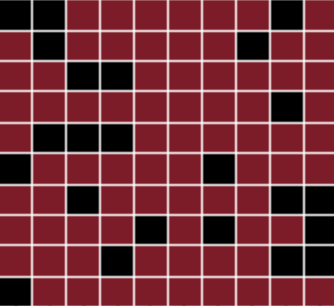} & 
        \includegraphics[height=0.04\textheight]{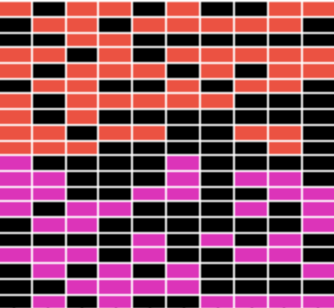} $\to$ \includegraphics[height=0.04\textheight]{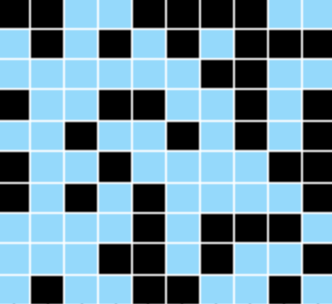} \\
    \bottomrule[1pt]
    \end{tabular}
    \caption{Samples from CausalARC task \texttt{SCMdky5}, where causal functions were \textit{and}, \textit{or}, and \textit{xor}.}
    \label{fig:logical_tasks}
\end{figure}

Logical counterfactual reasoning tasks used variants of \texttt{SCMdky5}, \texttt{SCMu3am}, and \texttt{SCMtcbq}, which are based to varying degrees on ARC-AGI-1 task \texttt{31d5ba1a} (Example \ref{example:scm_xor}). Causal functions for \texttt{SCMdky5} were a single logical operator (\textit{and}, \textit{or}, \textit{xor}) while \texttt{SCMu3am} and \texttt{SCMtcbq} used two logical operators. Task \texttt{SCMdky5} was generated for each logical operator at three dimensionalities ($10 \times 10$, $15 \times 15$, and $20 \times 20$), each with their own color palette (9 tasks total; Figure \ref{fig:logical_tasks}). Tasks \texttt{SCMu3am} and \texttt{SCMtcbq} were generated analogously, using one logical operator combination each (6 tasks total; 3 per SCM).

\subsubsection{Program Synthesis}

\textbf{Tasks} \; A single task was sampled from each of four SCMs: \texttt{SCMm5ob} (counting), \texttt{SCMev5t} (counting), \texttt{SCMfwpq} (extension), and \texttt{SCMz750} (extension). Example prompts are provided in Figures \ref{fig:l1_prompt_prog_synth} and \ref{fig:l3_prompt_prog_synth}. Prompts were sampled with varying numbers of in-context demonstrations (4, 6, 8) to assess impacts on program correctness. 

\subsubsection{Causal Discovery with Logical Reasoning}

\textbf{Tasks} \; For each of three tasks sampled from \texttt{SCMtcbq} (where logical operators composed \textit{xor} after \textit{and}), five prompts were sampled per level of the PCH. Tasks featured scaling problem complexity with respect to output array size ($10 \times 10$, $15 \times 15$, and $20 \times 20$). 

\begin{figure}[!h]
    \centering
    \begin{tabular}{c/c/c/c/c/c}
    \toprule[1pt]
        \multicolumn{2}{c}{\textsc{counting}} & \multicolumn{2}{c}{\textsc{extension}} & \multicolumn{2}{c}{\textsc{ordering}}  \\
        \cmidrule(lr){1-2} \cmidrule(lr){3-4} \cmidrule(lr){5-6}
        \texttt{SCMfuy3} & \texttt{SCMm5ob} & \texttt{SCMwoev} & \texttt{SCMz750} & \texttt{SCMffb8} & \texttt{SCMtzlq}  \\
        \includegraphics[height=0.03\textheight]{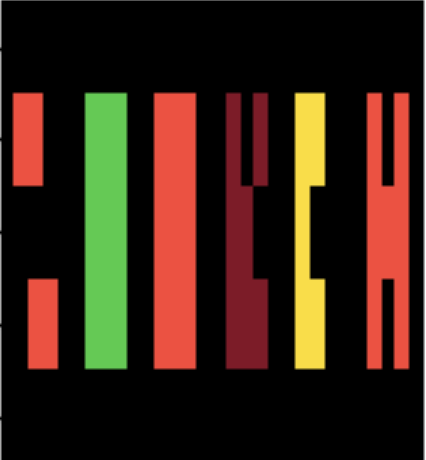} $\to$ \includegraphics[height=0.03\textheight]{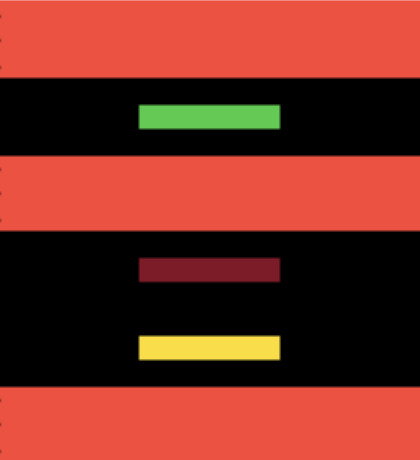} & 
        \includegraphics[height=0.03\textheight]{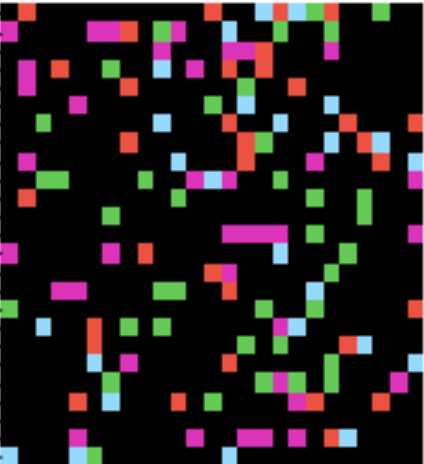} $\to$ \includegraphics[height=0.03\textheight]{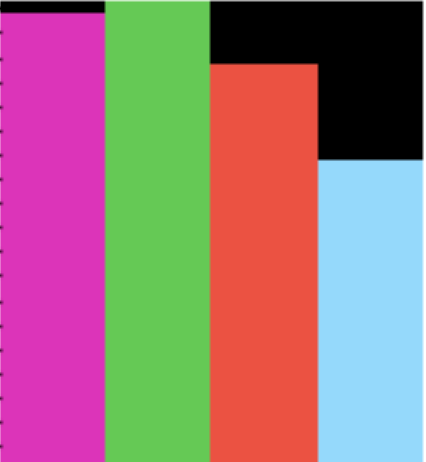} &
        \includegraphics[height=0.03\textheight]{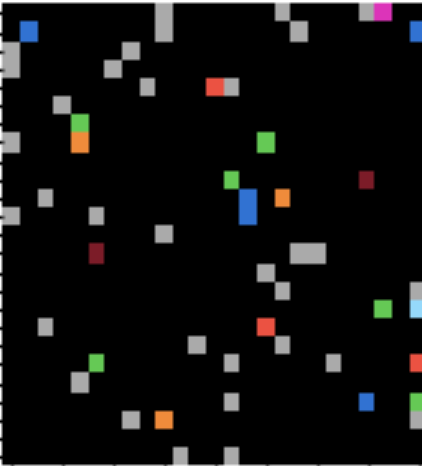} $\to$ \includegraphics[height=0.03\textheight]{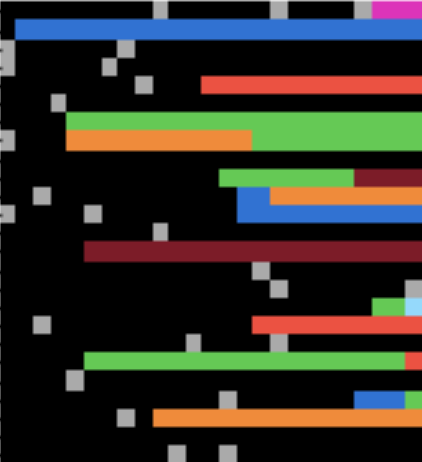} &
        \includegraphics[height=0.03\textheight]{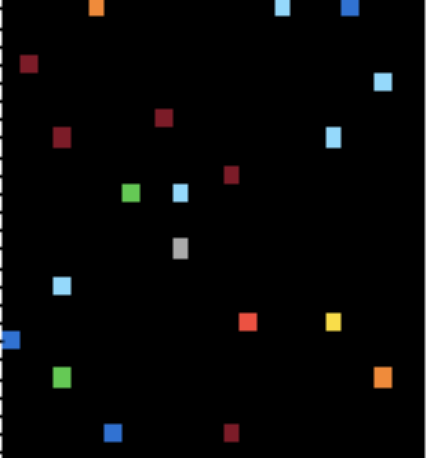} $\to$ \includegraphics[height=0.03\textheight]{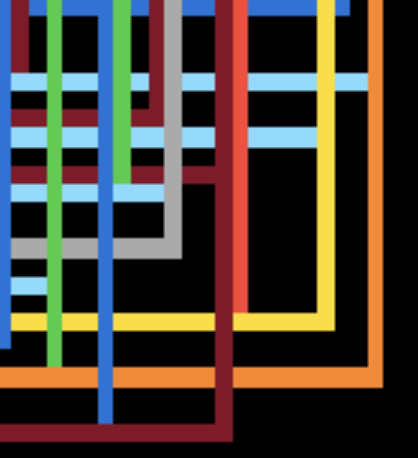} &
        \includegraphics[height=0.03\textheight]{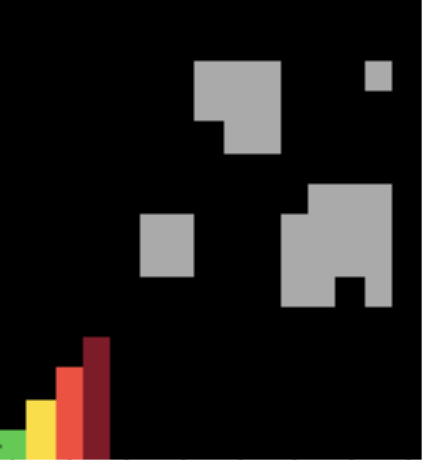} $\to$ \includegraphics[height=0.03\textheight]{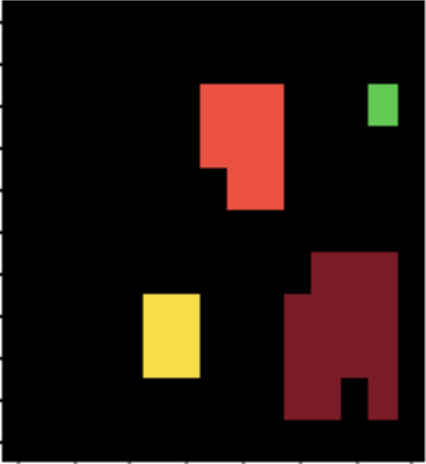}  & 
        \includegraphics[height=0.03\textheight]{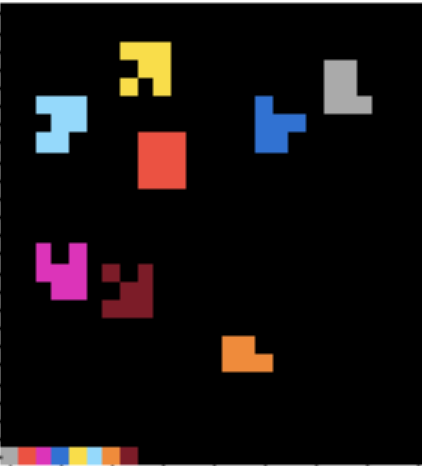} $\to$ \includegraphics[height=0.03\textheight]{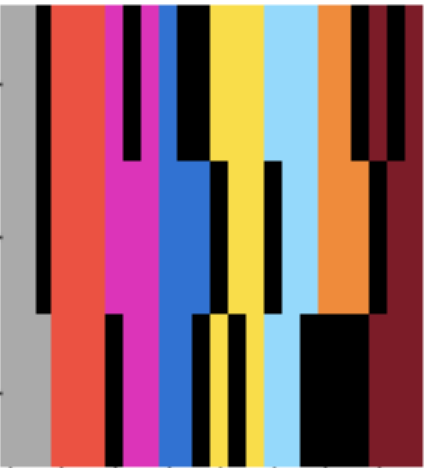}
         \\
    \bottomrule[1pt]
    \end{tabular}
    \caption{Counting, extension, and ordering test cases for the counterfactual reasoning experiment.}
    \label{fig:cf_not_logical_grids}
\end{figure}

\clearpage

\begin{figure}[!h]
    \centering

\begin{prompt}
You must solve the following puzzle by discovering the deterministic rule that maps inputs to outputs. You will then be asked to predict the output for a counterfactual example. Both the inputs and outputs are 2D grids of colored pixels. We provide example input-output pairs as demonstration. Grids are provided as Python arrays. You must output only a single Python array, and do not explain your reasoning.

Example input-output arrays:

[[0, 4, 4, 0, 4, 0, 0, 4, 0, 4], [4, 4, 0, 0, 4, 0, 4, 4, 0, 4], [0, 4, 4, 0, 0, 4, 4, 4, 0, 4], [4, 4, 4, 4, 0, 0, 4, 0, 4, 0], [4, 0, 0, 0, 0, 0, 0, 4, 4, 0], [4, 4, 4, 0, 0, 4, 0, 0, 0, 4], [4, 4, 4, 0, 4, 4, 0, 4, 4, 4], [0, 0, 4, 0, 0, 4, 4, 4, 0, 4], [4, 4, 4, 0, 4, 0, 4, 4, 4, 0], [0, 0, 4, 4, 0, 4, 0, 0, 4, 4], [0, 0, 5, 0, 5, 0, 5, 0, 0, 5], [0, 5, 5, 5, 5, 5, 0, 5, 0, 5], [5, 5, 0, 0, 0, 5, 5, 0, 0, 5], [0, 0, 0, 0, 5, 5, 0, 0, 0, 0], [0, 0, 0, 5, 5, 0, 0, 5, 5, 0], [5, 0, 0, 5, 5, 5, 0, 0, 5, 5], [5, 0, 0, 0, 5, 5, 0, 0, 5, 0], [0, 5, 5, 5, 5, 0, 5, 0, 5, 0], [5, 0, 0, 0, 5, 0, 5, 5, 5, 0], [5, 0, 0, 5, 0, 5, 5, 0, 5, 0]] -> [[0, 0, 2, 0, 2, 0, 0, 0, 0, 2], [0, 2, 0, 0, 2, 0, 0, 2, 0, 2], [0, 2, 0, 0, 0, 2, 2, 0, 0, 2], [0, 0, 0, 0, 0, 0, 0, 0, 0, 0], [0, 0, 0, 0, 0, 0, 0, 2, 2, 0], [2, 0, 0, 0, 0, 2, 0, 0, 0, 2], [2, 0, 0, 0, 2, 2, 0, 0, 2, 0], [0, 0, 2, 0, 0, 0, 2, 0, 0, 0], [2, 0, 0, 0, 2, 0, 2, 2, 2, 0], [0, 0, 0, 2, 0, 2, 0, 0, 2, 0]]

Example input-output arrays:

[[4, 0, 0, 4, 0, 0, 0, 4, 4, 4], [0, 4, 0, 4, 0, 0, 4, 0, 0, 0], [0, 4, 4, 0, 4, 0, 4, 4, 0, 0], [0, 4, 0, 0, 4, 0, 0, 0, 0, 0], [0, 0, 4, 0, 0, 0, 0, 0, 0, 4], [4, 0, 0, 0, 4, 4, 0, 0, 0, 0], [0, 0, 0, 0, 4, 4, 4, 0, 4, 4], [4, 0, 0, 0, 4, 0, 4, 0, 0, 4], [4, 4, 4, 0, 4, 4, 0, 4, 0, 0], [4, 0, 0, 4, 0, 4, 0, 0, 0, 0], [5, 5, 5, 0, 5, 0, 5, 0, 5, 5], [0, 5, 5, 0, 5, 5, 0, 5, 0, 5], [5, 0, 5, 5, 5, 0, 0, 0, 0, 0], [5, 0, 0, 5, 5, 0, 5, 0, 5, 5], [5, 5, 5, 5, 5, 5, 0, 5, 5, 5], [5, 5, 0, 0, 5, 0, 0, 0, 5, 5], [0, 0, 5, 5, 0, 5, 5, 5, 5, 5], [5, 0, 0, 0, 0, 5, 0, 0, 5, 5], [0, 5, 5, 5, 5, 0, 5, 5, 0, 0], [0, 0, 0, 0, 0, 0, 5, 5, 0, 5]] -> [[2, 0, 0, 0, 0, 0, 0, 0, 2, 2], [0, 2, 0, 0, 0, 0, 0, 0, 0, 0], [0, 0, 2, 0, 2, 0, 0, 0, 0, 0], [0, 0, 0, 0, 2, 0, 0, 0, 0, 0], [0, 0, 2, 0, 0, 0, 0, 0, 0, 2], [2, 0, 0, 0, 2, 0, 0, 0, 0, 0], [0, 0, 0, 0, 0, 2, 2, 0, 2, 2], [2, 0, 0, 0, 0, 0, 0, 0, 0, 2], [0, 2, 2, 0, 2, 0, 0, 2, 0, 0], [0, 0, 0, 0, 0, 0, 0, 0, 0, 0]]

Example input-output arrays:

[[0, 0, 0, 4, 0, 0, 0, 0, 0, 4], [0, 4, 0, 0, 4, 0, 0, 0, 4, 0], [4, 4, 0, 4, 0, 4, 4, 4, 4, 0], [4, 0, 0, 0, 4, 4, 0, 4, 0, 4], [4, 0, 0, 0, 0, 0, 4, 0, 0, 4], [0, 4, 4, 4, 0, 0, 0, 0, 4, 0], [0, 4, 4, 0, 0, 0, 0, 0, 0, 0], [0, 0, 0, 0, 0, 0, 4, 0, 4, 4], [0, 4, 4, 0, 4, 0, 4, 4, 0, 4], [4, 4, 0, 0, 4, 0, 4, 4, 4, 4], [5, 0, 0, 5, 5, 0, 0, 0, 0, 5], [0, 0, 5, 0, 5, 0, 0, 0, 0, 0], [5, 0, 0, 0, 0, 5, 0, 5, 0, 0], [0, 0, 0, 5, 5, 5, 0, 5, 0, 5], [0, 0, 0, 0, 0, 5, 5, 0, 0, 5], [5, 5, 5, 0, 0, 5, 5, 0, 0, 5], [0, 5, 5, 5, 0, 0, 5, 5, 0, 5], [0, 5, 5, 0, 0, 0, 0, 5, 5, 5], [5, 5, 0, 5, 5, 5, 0, 5, 0, 0], [0, 5, 5, 0, 5, 5, 0, 5, 5, 0]] -> [[0, 0, 0, 2, 0, 0, 0, 0, 0, 2], [0, 0, 0, 0, 2, 0, 0, 0, 0, 0], [2, 0, 0, 0, 0, 2, 0, 2, 0, 0], [0, 0, 0, 0, 2, 2, 0, 2, 0, 2], [0, 0, 0, 0, 0, 0, 2, 0, 0, 2], [0, 2, 2, 0, 0, 0, 0, 0, 0, 0], [0, 2, 2, 0, 0, 0, 0, 0, 0, 0], [0, 0, 0, 0, 0, 0, 0, 0, 2, 2], [0, 2, 0, 0, 2, 0, 0, 2, 0, 0], [0, 2, 0, 0, 2, 0, 0, 2, 2, 0]]

Counterfactual: Now imagine that we intervened on the previous input by changing some colors.

[[0, 0, 0, 8, 0, 0, 0, 0, 0, 8], [0, 8, 0, 0, 8, 0, 0, 0, 8, 0], [8, 8, 0, 8, 0, 8, 8, 8, 8, 0], [8, 0, 0, 0, 8, 8, 0, 8, 0, 8], [8, 0, 0, 0, 0, 0, 8, 0, 0, 8], [0, 8, 8, 8, 0, 0, 0, 0, 8, 0], [0, 8, 8, 0, 0, 0, 0, 0, 0, 0], [0, 0, 0, 0, 0, 0, 8, 0, 8, 8], [0, 8, 8, 0, 8, 0, 8, 8, 0, 8], [8, 8, 0, 0, 8, 0, 8, 8, 8, 8], [5, 0, 0, 5, 5, 0, 0, 0, 0, 5], [0, 0, 5, 0, 5, 0, 0, 0, 0, 0], [5, 0, 0, 0, 0, 5, 0, 5, 0, 0], [0, 0, 0, 5, 5, 5, 0, 5, 0, 5], [0, 0, 0, 0, 0, 5, 5, 0, 0, 5], [5, 5, 5, 0, 0, 5, 5, 0, 0, 5], [0, 5, 5, 5, 0, 0, 5, 5, 0, 5], [0, 5, 5, 0, 0, 0, 0, 5, 5, 5], [5, 5, 0, 5, 5, 5, 0, 5, 0, 0], [0, 5, 5, 0, 5, 5, 0, 5, 5, 0]] -> 
\end{prompt}

    \caption{L1 prompt for counterfactual reasoning.}
    \label{fig:l1_prompt_cf_reasoning}
\end{figure}
\begin{figure}[!h]
    \centering

\begin{prompt}
You must solve the following puzzle by discovering the deterministic rule that maps inputs to outputs. You will then be asked to predict the output for a counterfactual example. Both the inputs and outputs are 2D grids of colored pixels. We provide example input-output pairs along with counterfactual examples, which represent interventions on the original examples. Grids are provided as Python arrays. You must output only a single Python array, and do not explain your reasoning.

Example input-output arrays:

[[4, 0, 0, 4, 0, 0, 0, 4, 4, 4], [0, 4, 0, 4, 0, 0, 4, 0, 0, 0], [0, 4, 4, 0, 4, 0, 4, 4, 0, 0], [0, 4, 0, 0, 4, 0, 0, 0, 0, 0], [0, 0, 4, 0, 0, 0, 0, 0, 0, 4], [4, 0, 0, 0, 4, 4, 0, 0, 0, 0], [0, 0, 0, 0, 4, 4, 4, 0, 4, 4], [4, 0, 0, 0, 4, 0, 4, 0, 0, 4], [4, 4, 4, 0, 4, 4, 0, 4, 0, 0], [4, 0, 0, 4, 0, 4, 0, 0, 0, 0], [5, 5, 5, 0, 5, 0, 5, 0, 5, 5], [0, 5, 5, 0, 5, 5, 0, 5, 0, 5], [5, 0, 5, 5, 5, 0, 0, 0, 0, 0], [5, 0, 0, 5, 5, 0, 5, 0, 5, 5], [5, 5, 5, 5, 5, 5, 0, 5, 5, 5], [5, 5, 0, 0, 5, 0, 0, 0, 5, 5], [0, 0, 5, 5, 0, 5, 5, 5, 5, 5], [5, 0, 0, 0, 0, 5, 0, 0, 5, 5], [0, 5, 5, 5, 5, 0, 5, 5, 0, 0], [0, 0, 0, 0, 0, 0, 5, 5, 0, 5]] -> [[2, 0, 0, 0, 0, 0, 0, 0, 2, 2], [0, 2, 0, 0, 0, 0, 0, 0, 0, 0], [0, 0, 2, 0, 2, 0, 0, 0, 0, 0], [0, 0, 0, 0, 2, 0, 0, 0, 0, 0], [0, 0, 2, 0, 0, 0, 0, 0, 0, 2], [2, 0, 0, 0, 2, 0, 0, 0, 0, 0], [0, 0, 0, 0, 0, 2, 2, 0, 2, 2], [2, 0, 0, 0, 0, 0, 0, 0, 0, 2], [0, 2, 2, 0, 2, 0, 0, 2, 0, 0], [0, 0, 0, 0, 0, 0, 0, 0, 0, 0]]

Counterfactual: Now imagine that we intervened on the previous input by fixing some values.

[[4, 0, 0, 4, 0, 0, 0, 4, 4, 4], [0, 4, 0, 4, 0, 0, 4, 0, 0, 0], [0, 4, 4, 0, 4, 0, 4, 4, 0, 0], [0, 4, 0, 0, 4, 0, 0, 0, 0, 0], [0, 0, 4, 0, 0, 0, 0, 0, 0, 4], [4, 0, 0, 0, 4, 4, 0, 0, 0, 0], [0, 0, 0, 0, 4, 4, 4, 0, 4, 4], [4, 0, 0, 0, 4, 0, 4, 0, 0, 4], [4, 4, 4, 0, 4, 4, 0, 4, 0, 0], [4, 0, 0, 4, 0, 4, 0, 0, 0, 0], [0, 0, 0, 0, 0, 0, 0, 0, 0, 0], [0, 0, 0, 0, 0, 0, 0, 0, 0, 0], [0, 0, 0, 0, 0, 0, 0, 0, 0, 0], [0, 0, 0, 0, 0, 0, 0, 0, 0, 0], [0, 0, 0, 0, 0, 0, 0, 0, 0, 0], [0, 0, 0, 0, 0, 0, 0, 0, 0, 0], [0, 0, 0, 0, 0, 0, 0, 0, 0, 0], [0, 0, 0, 0, 0, 0, 0, 0, 0, 0], [0, 0, 0, 0, 0, 0, 0, 0, 0, 0], [0, 0, 0, 0, 0, 0, 0, 0, 0, 0]] -> [[0, 0, 0, 0, 0, 0, 0, 0, 0, 0], [0, 0, 0, 0, 0, 0, 0, 0, 0, 0], [0, 0, 0, 0, 0, 0, 0, 0, 0, 0], [0, 0, 0, 0, 0, 0, 0, 0, 0, 0], [0, 0, 0, 0, 0, 0, 0, 0, 0, 0], [0, 0, 0, 0, 0, 0, 0, 0, 0, 0], [0, 0, 0, 0, 0, 0, 0, 0, 0, 0], [0, 0, 0, 0, 0, 0, 0, 0, 0, 0], [0, 0, 0, 0, 0, 0, 0, 0, 0, 0], [0, 0, 0, 0, 0, 0, 0, 0, 0, 0]]

Example input-output arrays:

[[0, 0, 0, 4, 0, 0, 0, 0, 0, 4], [0, 4, 0, 0, 4, 0, 0, 0, 4, 0], [4, 4, 0, 4, 0, 4, 4, 4, 4, 0], [4, 0, 0, 0, 4, 4, 0, 4, 0, 4], [4, 0, 0, 0, 0, 0, 4, 0, 0, 4], [0, 4, 4, 4, 0, 0, 0, 0, 4, 0], [0, 4, 4, 0, 0, 0, 0, 0, 0, 0], [0, 0, 0, 0, 0, 0, 4, 0, 4, 4], [0, 4, 4, 0, 4, 0, 4, 4, 0, 4], [4, 4, 0, 0, 4, 0, 4, 4, 4, 4], [5, 0, 0, 5, 5, 0, 0, 0, 0, 5], [0, 0, 5, 0, 5, 0, 0, 0, 0, 0], [5, 0, 0, 0, 0, 5, 0, 5, 0, 0], [0, 0, 0, 5, 5, 5, 0, 5, 0, 5], [0, 0, 0, 0, 0, 5, 5, 0, 0, 5], [5, 5, 5, 0, 0, 5, 5, 0, 0, 5], [0, 5, 5, 5, 0, 0, 5, 5, 0, 5], [0, 5, 5, 0, 0, 0, 0, 5, 5, 5], [5, 5, 0, 5, 5, 5, 0, 5, 0, 0], [0, 5, 5, 0, 5, 5, 0, 5, 5, 0]] -> [[0, 0, 0, 2, 0, 0, 0, 0, 0, 2], [0, 0, 0, 0, 2, 0, 0, 0, 0, 0], [2, 0, 0, 0, 0, 2, 0, 2, 0, 0], [0, 0, 0, 0, 2, 2, 0, 2, 0, 2], [0, 0, 0, 0, 0, 0, 2, 0, 0, 2], [0, 2, 2, 0, 0, 0, 0, 0, 0, 0], [0, 2, 2, 0, 0, 0, 0, 0, 0, 0], [0, 0, 0, 0, 0, 0, 0, 0, 2, 2], [0, 2, 0, 0, 2, 0, 0, 2, 0, 0], [0, 2, 0, 0, 2, 0, 0, 2, 2, 0]]

Counterfactual: Now imagine that we intervened on the previous input by changing some colors.

[[0, 0, 0, 8, 0, 0, 0, 0, 0, 8], [0, 8, 0, 0, 8, 0, 0, 0, 8, 0], [8, 8, 0, 8, 0, 8, 8, 8, 8, 0], [8, 0, 0, 0, 8, 8, 0, 8, 0, 8], [8, 0, 0, 0, 0, 0, 8, 0, 0, 8], [0, 8, 8, 8, 0, 0, 0, 0, 8, 0], [0, 8, 8, 0, 0, 0, 0, 0, 0, 0], [0, 0, 0, 0, 0, 0, 8, 0, 8, 8], [0, 8, 8, 0, 8, 0, 8, 8, 0, 8], [8, 8, 0, 0, 8, 0, 8, 8, 8, 8], [5, 0, 0, 5, 5, 0, 0, 0, 0, 5], [0, 0, 5, 0, 5, 0, 0, 0, 0, 0], [5, 0, 0, 0, 0, 5, 0, 5, 0, 0], [0, 0, 0, 5, 5, 5, 0, 5, 0, 5], [0, 0, 0, 0, 0, 5, 5, 0, 0, 5], [5, 5, 5, 0, 0, 5, 5, 0, 0, 5], [0, 5, 5, 5, 0, 0, 5, 5, 0, 5], [0, 5, 5, 0, 0, 0, 0, 5, 5, 5], [5, 5, 0, 5, 5, 5, 0, 5, 0, 0], [0, 5, 5, 0, 5, 5, 0, 5, 5, 0]] -> 
\end{prompt}

    \caption{L3 prompt for counterfactual reasoning.}
    \label{fig:l3_prompt_cf_reasoning}
\end{figure}

\begin{figure}[!h]
    \centering

\begin{prompt}
You must solve the following puzzle by discovering the deterministic rule that maps inputs to outputs. Both the inputs and outputs are 2D Python arrays of colored pixels. We provide example input-output pairs as demonstration. To solve the problem, express the deterministic rule as a Python program. Do not explain your reasoning, and only output a single Python program.

Example input-output arrays:

[[0, 0, 0, 0, 0, 9, 0, 4, 0, 3], [0, 4, 0, 0, 0, 0, 0, 0, 0, 0], [0, 0, 0, 0, 4, 0, 2, 0, 0, 0], [0, 0, 0, 0, 0, 9, 0, 0, 2, 0], [0, 0, 0, 0, 0, 0, 0, 0, 0, 3], [0, 9, 0, 0, 0, 4, 0, 0, 0, 2], [0, 0, 0, 0, 0, 9, 0, 2, 4, 0], [3, 0, 3, 0, 3, 0, 0, 0, 3, 0], [0, 0, 4, 3, 0, 0, 3, 0, 4, 0], [9, 3, 0, 0, 0, 0, 2, 0, 0, 0]] -> [[0, 0, 0, 3], [0, 0, 0, 3], [4, 0, 0, 3], [4, 0, 0, 3], [4, 2, 9, 3], [4, 2, 9, 3], [4, 2, 9, 3], [4, 2, 9, 3], [4, 2, 9, 3]]

Example input-output arrays:

[[0, 0, 0, 0, 0, 0, 0, 0, 0, 0], [9, 0, 0, 2, 0, 0, 9, 0, 0, 0], [0, 0, 0, 0, 0, 0, 2, 0, 9, 0], [9, 4, 0, 0, 9, 3, 0, 0, 0, 0], [0, 0, 0, 2, 0, 0, 2, 3, 0, 0], [0, 0, 0, 0, 0, 0, 0, 0, 0, 0], [0, 9, 4, 0, 9, 0, 0, 0, 2, 9], [0, 4, 0, 0, 0, 2, 0, 0, 0, 0], [0, 0, 3, 0, 9, 0, 9, 0, 9, 4], [0, 0, 9, 0, 4, 0, 0, 0, 3, 0]] -> [[0, 0, 9, 0], [0, 0, 9, 0], [0, 0, 9, 0], [0, 0, 9, 0], [0, 0, 9, 0], [0, 0, 9, 0], [0, 2, 9, 0], [4, 2, 9, 0], [4, 2, 9, 3], [4, 2, 9, 3], [4, 2, 9, 3], [4, 2, 9, 3]]

Example input-output arrays:

[[0, 0, 0, 0, 0, 3, 0, 0, 0, 3], [0, 0, 0, 0, 0, 0, 0, 0, 0, 0], [0, 9, 0, 0, 3, 0, 0, 0, 0, 0], [0, 9, 3, 0, 0, 0, 0, 0, 0, 0], [0, 0, 0, 0, 0, 0, 0, 4, 0, 0], [3, 0, 0, 0, 0, 3, 0, 9, 0, 0], [0, 0, 4, 0, 0, 9, 0, 0, 9, 4], [0, 3, 0, 0, 0, 0, 0, 0, 0, 0], [9, 0, 4, 3, 2, 9, 0, 0, 0, 0], [4, 0, 0, 0, 0, 0, 0, 0, 0, 0]] -> [[0, 0, 0, 3], [0, 0, 9, 3], [0, 0, 9, 3], [4, 0, 9, 3], [4, 0, 9, 3], [4, 0, 9, 3], [4, 0, 9, 3], [4, 2, 9, 3]]

Example input-output arrays:

[[0, 0, 2, 0, 0, 0, 0, 0, 0, 0], [0, 0, 9, 0, 0, 0, 0, 0, 0, 0], [0, 0, 0, 0, 2, 0, 0, 0, 0, 0], [0, 9, 2, 2, 2, 0, 0, 0, 0, 9], [0, 0, 2, 0, 0, 0, 0, 0, 0, 0], [0, 4, 3, 0, 0, 0, 0, 0, 3, 0], [0, 0, 0, 2, 3, 0, 0, 0, 0, 0], [0, 0, 4, 9, 2, 0, 0, 0, 4, 4], [0, 0, 0, 0, 4, 0, 0, 0, 0, 0], [0, 0, 4, 0, 0, 0, 0, 9, 3, 9]] -> [[0, 2, 0, 0], [0, 2, 0, 0], [4, 2, 9, 0], [4, 2, 9, 0], [4, 2, 9, 3], [4, 2, 9, 3], [4, 2, 9, 3], [4, 2, 9, 3]]

Example input-output arrays:

[[0, 9, 0, 0, 0, 0, 0, 0, 0, 0], [0, 0, 0, 0, 0, 3, 0, 9, 2, 0], [0, 0, 0, 0, 0, 2, 0, 0, 0, 0], [0, 4, 2, 0, 0, 0, 0, 9, 0, 0], [0, 0, 0, 4, 0, 0, 0, 0, 0, 0], [0, 0, 0, 0, 0, 0, 2, 0, 0, 0], [0, 0, 0, 2, 4, 0, 0, 3, 0, 0], [3, 2, 0, 4, 0, 3, 0, 2, 0, 0], [0, 3, 0, 0, 4, 0, 9, 9, 4, 0], [0, 0, 0, 0, 0, 0, 0, 0, 0, 0]] -> [[0, 2, 0, 0], [4, 2, 0, 0], [4, 2, 9, 3], [4, 2, 9, 3], [4, 2, 9, 3], [4, 2, 9, 3], [4, 2, 9, 3]]

Example input-output arrays:

[[0, 0, 0, 0, 0, 3, 0, 0, 0, 3], [0, 0, 0, 0, 0, 0, 0, 0, 0, 0], [0, 9, 0, 0, 3, 0, 0, 0, 0, 0], [0, 9, 3, 0, 0, 0, 0, 0, 0, 0], [0, 0, 0, 0, 0, 0, 0, 4, 0, 0], [3, 0, 0, 0, 0, 3, 0, 9, 0, 0], [0, 0, 4, 0, 0, 9, 0, 0, 9, 4], [0, 3, 0, 0, 0, 0, 0, 0, 0, 0], [9, 0, 4, 3, 2, 9, 0, 0, 0, 0], [4, 0, 0, 0, 0, 0, 0, 0, 0, 0]] -> [[0, 0, 0, 3], [0, 0, 9, 3], [0, 0, 9, 3], [4, 0, 9, 3], [4, 0, 9, 3], [4, 0, 9, 3], [4, 0, 9, 3], [4, 2, 9, 3]]
\end{prompt}

    \caption{L1 prompt for program synthesis with six in-context examples.}
    \label{fig:l1_prompt_prog_synth}
\end{figure}
\begin{figure}[!h]
    \centering

\begin{prompt}
You must solve the following puzzle by discovering the deterministic rule that maps inputs to outputs. Both the inputs and outputs are 2D Python arrays of colored pixels. We provide example input-output pairs along with counterfactual examples, which represent interventions on the original examples. To solve the problem, express the deterministic rule as a Python program. Do not explain your reasoning, and only output a single Python program.

Example input-output arrays:

[[4, 3, 0, 0, 0, 9, 9, 9, 0, 0], [0, 0, 0, 3, 3, 0, 0, 0, 0, 0], [0, 0, 0, 0, 0, 0, 2, 0, 9, 0], [0, 3, 0, 0, 0, 0, 0, 0, 2, 0], [0, 0, 0, 2, 0, 0, 0, 0, 9, 0], [0, 0, 0, 0, 0, 0, 9, 2, 0, 0], [0, 2, 0, 0, 0, 0, 0, 3, 2, 4], [4, 0, 0, 0, 0, 0, 0, 9, 0, 0], [0, 0, 3, 0, 0, 0, 0, 2, 0, 0], [0, 0, 0, 9, 0, 0, 3, 0, 0, 2]] -> [[0, 2, 9, 0], [0, 2, 9, 3], [0, 2, 9, 3], [0, 2, 9, 3], [0, 2, 9, 3], [4, 2, 9, 3], [4, 2, 9, 3], [4, 2, 9, 3]]

Counterfactual: Now imagine that we intervened on the previous input by rotating or flipping it.

[[0, 0, 9, 0, 4, 0, 0, 0, 3, 0], [0, 0, 3, 0, 9, 0, 9, 0, 9, 4], [0, 4, 0, 0, 0, 2, 0, 0, 0, 0], [0, 9, 4, 0, 9, 0, 0, 0, 2, 9], [0, 0, 0, 0, 0, 0, 0, 0, 0, 0], [0, 0, 0, 2, 0, 0, 2, 3, 0, 0], [9, 4, 0, 0, 9, 3, 0, 0, 0, 0], [0, 0, 0, 0, 0, 0, 2, 0, 9, 0], [9, 0, 0, 2, 0, 0, 9, 0, 0, 0], [0, 0, 0, 0, 0, 0, 0, 0, 0, 0]] -> [[0, 0, 9, 0], [0, 0, 9, 0], [0, 0, 9, 0], [0, 0, 9, 0], [0, 0, 9, 0], [0, 0, 9, 0], [0, 2, 9, 0], [4, 2, 9, 0], [4, 2, 9, 3], [4, 2, 9, 3], [4, 2, 9, 3], [4, 2, 9, 3]]

Example input-output arrays:

[[0, 0, 0, 0, 0, 0, 4, 0, 0, 0], [0, 0, 0, 0, 0, 0, 0, 0, 4, 0], [0, 0, 0, 2, 0, 0, 0, 0, 0, 0], [0, 4, 0, 0, 0, 9, 0, 0, 4, 0], [0, 0, 0, 0, 0, 3, 0, 0, 3, 0], [0, 0, 0, 0, 0, 0, 4, 2, 0, 0], [0, 2, 0, 0, 0, 0, 3, 0, 0, 0], [0, 0, 2, 0, 0, 0, 0, 0, 0, 0], [0, 3, 0, 0, 0, 0, 0, 4, 0, 0], [0, 0, 0, 0, 0, 9, 4, 0, 0, 0]] -> [[4, 0, 0, 0], [4, 0, 0, 0], [4, 0, 0, 0], [4, 2, 0, 3], [4, 2, 0, 3], [4, 2, 9, 3], [4, 2, 9, 3]]

Counterfactual: Now imagine that we intervened on the previous input by changing some colors.

[[7, 7, 7, 7, 7, 7, 4, 7, 7, 7], [7, 7, 7, 7, 7, 7, 7, 7, 4, 7], [7, 7, 7, 2, 7, 7, 7, 7, 7, 7], [7, 4, 7, 7, 7, 9, 7, 7, 4, 7], [7, 7, 7, 7, 7, 3, 7, 7, 3, 7], [7, 7, 7, 7, 7, 7, 4, 2, 7, 7], [7, 2, 7, 7, 7, 7, 3, 7, 7, 7], [7, 7, 2, 7, 7, 7, 7, 7, 7, 7], [7, 3, 7, 7, 7, 7, 7, 4, 7, 7], [7, 7, 7, 7, 7, 9, 4, 7, 7, 7]] -> [[4, 7, 7, 7], [4, 7, 7, 7], [4, 7, 7, 7], [4, 2, 7, 3], [4, 2, 7, 3], [4, 2, 9, 3], [4, 2, 9, 3]]

Example input-output arrays:

[[0, 0, 0, 0, 0, 0, 0, 3, 0, 0], [4, 0, 0, 0, 0, 9, 3, 0, 0, 0], [0, 0, 0, 0, 0, 0, 0, 4, 0, 0], [0, 0, 0, 0, 2, 0, 4, 0, 0, 0], [0, 3, 0, 0, 4, 0, 0, 0, 2, 0], [0, 0, 0, 0, 0, 0, 0, 0, 0, 0], [0, 0, 0, 0, 2, 4, 2, 0, 0, 0], [0, 2, 0, 9, 3, 0, 9, 0, 3, 0], [0, 0, 0, 0, 0, 0, 0, 4, 0, 0], [0, 3, 4, 0, 4, 9, 4, 0, 0, 0]] -> [[4, 0, 0, 0], [4, 0, 0, 0], [4, 0, 0, 0], [4, 0, 0, 3], [4, 2, 0, 3], [4, 2, 9, 3], [4, 2, 9, 3], [4, 2, 9, 3], [4, 2, 9, 3]]

Counterfactual: Now imagine that we intervened on the previous input by rotating or flipping it.

[[0, 4, 9, 0, 9, 0, 0, 0, 2, 0], [0, 0, 0, 0, 4, 0, 2, 0, 0, 0], [0, 0, 0, 2, 0, 0, 0, 0, 0, 0], [0, 0, 0, 4, 0, 0, 0, 0, 0, 0], [0, 0, 9, 0, 9, 0, 0, 0, 0, 4], [9, 0, 0, 0, 2, 0, 0, 4, 9, 4], [0, 0, 0, 0, 0, 0, 0, 2, 0, 0], [0, 0, 9, 0, 0, 3, 0, 3, 0, 0], [0, 0, 0, 0, 0, 0, 0, 0, 0, 0], [0, 2, 0, 0, 0, 0, 3, 4, 0, 0]] -> [[4, 0, 9, 0], [4, 2, 9, 0], [4, 2, 9, 0], [4, 2, 9, 0], [4, 2, 9, 3], [4, 2, 9, 3], [4, 2, 9, 3]]

\end{prompt}

    \caption{L3 prompt for program synthesis with six in-context examples.}
    \label{fig:l3_prompt_prog_synth}
\end{figure}

\begin{figure}[!h]
    \centering

\begin{prompt}
    You must solve the following causal discovery problem, where the cells in an input array are causal parents of cells in an output array. Both the inputs and outputs are 2D Python arrays of colored pixels. We provide example input-output pairs as demonstration. You must predict the causal function(s) that relate parent cells in the input to their children in the output. Be concise: do not explain your reasoning, and start your answer with `The logical operators are'.
    
Example input-output arrays:

[[4, 0, 4, 0, 0, 4, 4, 4, 4, 0], [4, 4, 0, 4, 4, 4, 0, 4, 0, 0], [0, 0, 4, 0, 0, 0, 4, 4, 4, 0], [4, 4, 4, 0, 4, 4, 0, 0, 0, 0], [4, 4, 0, 0, 4, 0, 0, 4, 4, 0], [0, 0, 4, 4, 4, 0, 0, 4, 4, 4], [0, 4, 4, 0, 4, 4, 4, 4, 4, 0], [4, 0, 4, 4, 0, 0, 0, 4, 0, 0], [4, 4, 4, 4, 0, 4, 0, 4, 0, 4], [4, 4, 0, 4, 4, 0, 0, 0, 0, 4], [0, 5, 0, 5, 0, 0, 5, 5, 5, 0], [0, 0, 5, 0, 5, 0, 0, 0, 0, 0], [0, 0, 0, 0, 0, 5, 5, 5, 5, 5], [5, 0, 0, 5, 5, 5, 5, 0, 5, 5], [5, 0, 0, 5, 0, 5, 0, 0, 0, 0], [0, 5, 0, 5, 5, 0, 5, 0, 5, 5], [0, 0, 5, 0, 0, 5, 0, 5, 5, 5], [5, 0, 5, 0, 0, 0, 5, 5, 0, 0], [0, 0, 5, 5, 0, 0, 0, 0, 0, 0], [0, 5, 5, 5, 0, 5, 0, 0, 0, 5], [2, 0, 0, 0, 2, 0, 2, 0, 0, 2], [2, 0, 2, 2, 2, 2, 0, 0, 0, 0], [2, 0, 2, 2, 0, 2, 2, 0, 2, 2], [2, 0, 0, 2, 2, 0, 0, 2, 2, 0], [0, 0, 2, 2, 2, 2, 2, 2, 0, 0], [0, 2, 2, 2, 2, 0, 2, 0, 0, 0], [0, 0, 2, 0, 2, 0, 0, 2, 2, 0], [0, 0, 0, 2, 2, 0, 0, 2, 2, 2], [2, 2, 0, 0, 2, 2, 0, 0, 2, 2], [0, 2, 0, 2, 0, 0, 0, 2, 2, 2]] -> [[1, 0, 0, 0, 1, 0, 0, 1, 1, 1], [1, 0, 1, 1, 0, 1, 0, 0, 0, 0], [1, 0, 1, 1, 0, 1, 0, 1, 0, 1], [0, 0, 0, 1, 0, 1, 0, 1, 1, 0], [1, 0, 1, 1, 1, 1, 1, 1, 0, 0], [0, 1, 1, 0, 0, 0, 1, 0, 1, 1], [0, 0, 0, 0, 1, 1, 0, 0, 0, 0], [1, 0, 1, 1, 1, 0, 0, 0, 1, 1], [1, 1, 1, 1, 1, 1, 0, 0, 1, 1], [0, 0, 0, 0, 0, 0, 0, 1, 1, 0]]

Example input-output arrays:

[[0, 4, 4, 4, 0, 4, 4, 0, 4, 4], [0, 4, 4, 4, 0, 0, 4, 4, 4, 4], [0, 4, 0, 4, 4, 0, 0, 4, 4, 0], [0, 0, 0, 0, 0, 4, 0, 0, 0, 4], [0, 4, 4, 0, 0, 4, 0, 4, 4, 4], [0, 4, 0, 4, 4, 4, 4, 0, 0, 0], [0, 0, 0, 0, 4, 4, 4, 0, 0, 0], [0, 0, 0, 4, 0, 0, 0, 4, 0, 0], [0, 4, 0, 0, 4, 0, 4, 4, 0, 0], [4, 0, 4, 4, 4, 0, 4, 4, 0, 4], [0, 0, 0, 5, 5, 5, 0, 0, 5, 5], [0, 5, 0, 5, 0, 5, 0, 5, 0, 5], [0, 5, 5, 5, 5, 5, 0, 5, 0, 0], [0, 5, 5, 0, 0, 0, 5, 0, 5, 5], [0, 0, 5, 0, 5, 5, 5, 5, 0, 0], [0, 0, 5, 0, 5, 0, 5, 5, 5, 5], [5, 5, 5, 0, 0, 0, 0, 0, 5, 5], [5, 5, 0, 0, 5, 0, 5, 0, 5, 5], [0, 0, 5, 0, 0, 0, 0, 5, 5, 0], [5, 5, 5, 0, 5, 0, 0, 5, 5, 5], [0, 0, 2, 0, 0, 2, 0, 0, 0, 0], [2, 2, 2, 2, 0, 2, 2, 0, 0, 2], [0, 0, 2, 0, 0, 2, 0, 2, 2, 2], [2, 0, 0, 2, 2, 0, 2, 0, 2, 2], [0, 2, 2, 2, 0, 2, 0, 2, 2, 2], [0, 2, 0, 0, 2, 2, 2, 2, 0, 0], [2, 0, 0, 0, 0, 0, 0, 2, 0, 2], [2, 2, 0, 2, 0, 2, 2, 2, 0, 2], [2, 0, 0, 2, 0, 0, 0, 2, 2, 2], [2, 0, 0, 2, 2, 0, 0, 2, 2, 0]] -> [[0, 0, 1, 1, 0, 0, 0, 0, 1, 1], [1, 0, 1, 0, 0, 1, 1, 1, 0, 0], [0, 1, 1, 1, 1, 1, 0, 0, 1, 1], [1, 0, 0, 1, 1, 0, 1, 0, 1, 0], [0, 1, 0, 1, 0, 0, 0, 0, 1, 1], [0, 1, 0, 0, 0, 1, 0, 1, 0, 0], [1, 0, 0, 0, 0, 0, 0, 1, 0, 1], [1, 1, 0, 1, 0, 1, 1, 1, 0, 1], [1, 0, 0, 1, 0, 0, 0, 0, 1, 1], [0, 0, 1, 1, 0, 0, 0, 0, 1, 1]]

...
\end{prompt}

    \caption{Excerpt of L1 prompt for logical reasoning. Four in-context examples were provided.}
    \label{fig:l1_prompt_discovery}
\end{figure}
\begin{figure}[!h]
    \centering

\begin{prompt}

You must solve the following causal discovery problem, where the cells in an input array are causal parents of cells in an output array. Both the inputs and outputs are 2D Python arrays of colored pixels. We provide example input-output pairs along with counterfactual examples, which represent interventions on the original examples. You must predict the causal function(s) that relate parent cells in the input to their children in the output. Be concise: do not explain your reasoning, and start your answer with `The logical operators are'.
    
Example input-output arrays:

[[0, 4, 0, 0, 0, 4, 4, 0, 0, 0], [0, 4, 4, 0, 4, 0, 0, 4, 4, 4], [0, 0, 0, 0, 0, 4, 4, 4, 4, 4], [4, 4, 0, 4, 0, 0, 0, 4, 0, 0], [4, 0, 0, 4, 0, 0, 0, 4, 0, 0], [0, 4, 0, 4, 0, 4, 0, 0, 4, 0], [0, 4, 0, 0, 0, 4, 4, 0, 4, 4], [4, 4, 4, 0, 4, 4, 4, 0, 0, 4], [0, 4, 4, 0, 0, 4, 0, 4, 4, 4], [4, 4, 4, 0, 0, 0, 0, 4, 0, 4], [5, 0, 5, 5, 5, 5, 5, 0, 5, 0], [5, 5, 5, 5, 0, 5, 0, 0, 0, 5], [0, 0, 0, 0, 0, 0, 0, 5, 5, 5], [0, 0, 0, 0, 5, 5, 0, 0, 5, 0], [0, 0, 0, 0, 0, 5, 5, 5, 5, 5], [5, 5, 5, 0, 0, 0, 5, 0, 0, 5], [0, 0, 0, 5, 0, 5, 0, 0, 0, 5], [5, 5, 0, 0, 0, 0, 0, 5, 5, 0], [5, 5, 5, 0, 5, 0, 5, 0, 0, 5], [0, 5, 5, 0, 5, 0, 5, 0, 5, 5], [2, 0, 0, 2, 0, 0, 0, 2, 0, 2], [2, 2, 0, 2, 2, 2, 2, 2, 2, 2], [2, 0, 0, 0, 0, 2, 0, 0, 2, 0], [2, 2, 2, 0, 2, 0, 0, 2, 0, 2], [2, 2, 0, 0, 0, 0, 2, 0, 2, 0], [2, 2, 2, 0, 2, 0, 2, 2, 0, 0], [0, 2, 0, 0, 0, 2, 2, 0, 2, 0], [2, 0, 0, 2, 2, 0, 0, 2, 2, 0], [0, 2, 0, 0, 2, 2, 2, 2, 2, 0], [0, 2, 0, 2, 0, 2, 0, 2, 2, 2]] -> [[1, 0, 0, 1, 0, 1, 1, 1, 0, 1], [1, 0, 1, 1, 1, 1, 1, 1, 1, 0], [1, 0, 0, 0, 0, 1, 0, 1, 0, 1], [1, 1, 1, 0, 1, 0, 0, 1, 0, 1], [1, 1, 0, 0, 0, 0, 1, 1, 1, 0], [1, 0, 1, 0, 1, 0, 1, 1, 0, 0], [0, 1, 0, 0, 0, 0, 1, 0, 1, 1], [0, 1, 0, 1, 1, 0, 0, 1, 1, 0], [0, 0, 1, 0, 1, 1, 1, 1, 1, 1], [0, 0, 1, 1, 0, 1, 0, 1, 1, 0]]

Counterfactual: Now imagine that we intervened on the original input by changing some colors.

[[4, 4, 4, 4, 0, 0, 0, 4, 4, 4], [0, 0, 0, 4, 0, 0, 4, 4, 4, 0], [4, 0, 0, 4, 4, 4, 0, 4, 4, 0], [0, 0, 0, 0, 4, 0, 0, 0, 0, 4], [4, 0, 0, 0, 0, 4, 0, 4, 4, 0], [0, 0, 4, 4, 0, 0, 0, 4, 4, 0], [4, 0, 4, 0, 0, 4, 4, 4, 0, 0], [0, 4, 0, 0, 0, 0, 0, 4, 4, 0], [0, 0, 0, 0, 4, 0, 4, 0, 0, 4], [4, 0, 4, 0, 4, 4, 0, 4, 4, 4], [0, 0, 6, 0, 6, 6, 6, 6, 6, 6], [6, 0, 0, 0, 0, 6, 6, 6, 0, 0], [6, 0, 6, 6, 0, 0, 6, 0, 0, 6], [6, 0, 0, 6, 6, 0, 0, 6, 0, 6], [6, 0, 0, 6, 6, 0, 6, 0, 6, 0], [6, 0, 6, 6, 0, 0, 6, 6, 6, 6], [0, 6, 0, 6, 0, 6, 0, 6, 0, 0], [0, 6, 6, 0, 6, 6, 0, 0, 6, 0], [6, 0, 6, 6, 0, 6, 0, 0, 0, 6], [0, 0, 0, 6, 0, 6, 0, 6, 6, 6], [0, 0, 2, 2, 2, 2, 0, 0, 2, 2], [0, 2, 2, 0, 2, 2, 2, 0, 2, 2], [2, 2, 2, 2, 2, 0, 0, 0, 0, 0], [0, 2, 0, 0, 0, 0, 2, 0, 0, 2], [2, 0, 0, 0, 0, 0, 0, 0, 0, 2], [0, 2, 0, 0, 0, 0, 2, 2, 2, 2], [2, 0, 2, 0, 2, 0, 2, 0, 2, 0], [0, 0, 2, 2, 0, 2, 0, 0, 0, 0], [0, 2, 2, 0, 0, 0, 2, 0, 2, 0], [0, 0, 2, 2, 2, 2, 0, 2, 0, 0]] -> [[0, 0, 0, 1, 1, 1, 0, 1, 0, 0], [0, 1, 1, 0, 1, 1, 0, 1, 1, 1], [0, 1, 1, 0, 1, 0, 0, 0, 0, 0], [0, 1, 0, 0, 1, 0, 1, 0, 0, 0], [0, 0, 0, 0, 0, 0, 0, 0, 1, 1], [0, 1, 1, 1, 0, 0, 1, 0, 0, 1], [1, 0, 1, 0, 1, 1, 1, 1, 1, 0], [0, 1, 1, 1, 0, 1, 0, 0, 1, 0], [0, 1, 1, 0, 0, 0, 1, 0, 1, 1], [0, 0, 1, 1, 1, 0, 0, 0, 1, 1]]

...
\end{prompt}

    \caption{Excerpt of L3 prompt for logical reasoning. Four in-context examples were provided.}
    \label{fig:l3_prompt_discovery}
\end{figure}

\clearpage

\begin{figure}[!h]
    \centering
\begin{response}
    \# For ARC-style input/output:
    
if \_\_name\_\_ == "\_\_main\_\_":

    \quad import sys, json
    
    \quad data = json.load(sys.stdin)
    
    \quad inp = data["input"]
    
    \quad out = solve(inp)
    
    \quad print(json.dumps(out))
    
\end{response}
    \caption{Evidence that o4-mini may have been fine-tuned for ARC-like program synthesis (``\texttt{For ARC-style input/output:... out = solve(inp)}''), excerpted from a response to a CausalARC query that did not mention ARC in-context.}
    \label{fig:o4_fine_tuning}
\end{figure}

\begin{figure}[!h]
    \centering
\begin{response}
    Replicate 1: The logical operators are: For each output cell at `(row, col)', let `A = (input[row][col] != 0)', `B = (input[row + 10][col] != 0)', and `C = (input[row + 20][col] != 0)'. The output cell `output[row][col]' is `1' if `(A AND B) XOR C' is true, and `0' otherwise. \\ \\

    Replicate 2: The logical operators are `((x AND y) XOR z)', where `x', `y', and `z' are boolean values indicating if the corresponding input cell (from the first, second, and third 15x15 input arrays, respectively) is non-zero. The output cell is `2' if the expression is true, and `0' if false. \\ \\

    Replicate 3: The logical operators are XOR and AND. For each output cell `O[r][c]', its value is determined by three input cells: `P1 = I[r][c]', `P2 = I[r+10][c]', and `P3 = I[r+20][c]'. The output value is `1' if `(P3 != 0) XOR ((P1 != 0) AND (P2 != 0))' is true, and `0' otherwise.
\end{response}
    \caption{Three correct replicate responses from Gemini 2.5 Flash for the causal discovery with logical reasoning task. The causal parents and logical operators were correctly identified.}
    \label{fig:gemini_response_discovery}
\end{figure}

\begin{figure}[!h]
    \centering
\begin{response}
    Replicate 1: The logical operators are AND and XOR. \\ \\
    
   Replicate 3: The logical operators are AND between the first and second input grids, then XOR with the third input grid. \\ \\
   
   Replicate 4: The logical operators are AND and XOR, applied pointwise as (color4 AND color5) XOR color2.
\end{response}
    \caption{Three correct replicate responses from Grok 4 for the causal discovery with logical reasoning task. Logical operators were correctly identified and causal parents were explicitly declared in some cases.}
    \label{fig:grok_response_discovery}
\end{figure}

\clearpage

\subsection{Experimental Results}
\label{appendix:experimental_results}

\begin{figure}[!h]
    \centering
    \includegraphics[width=\linewidth]{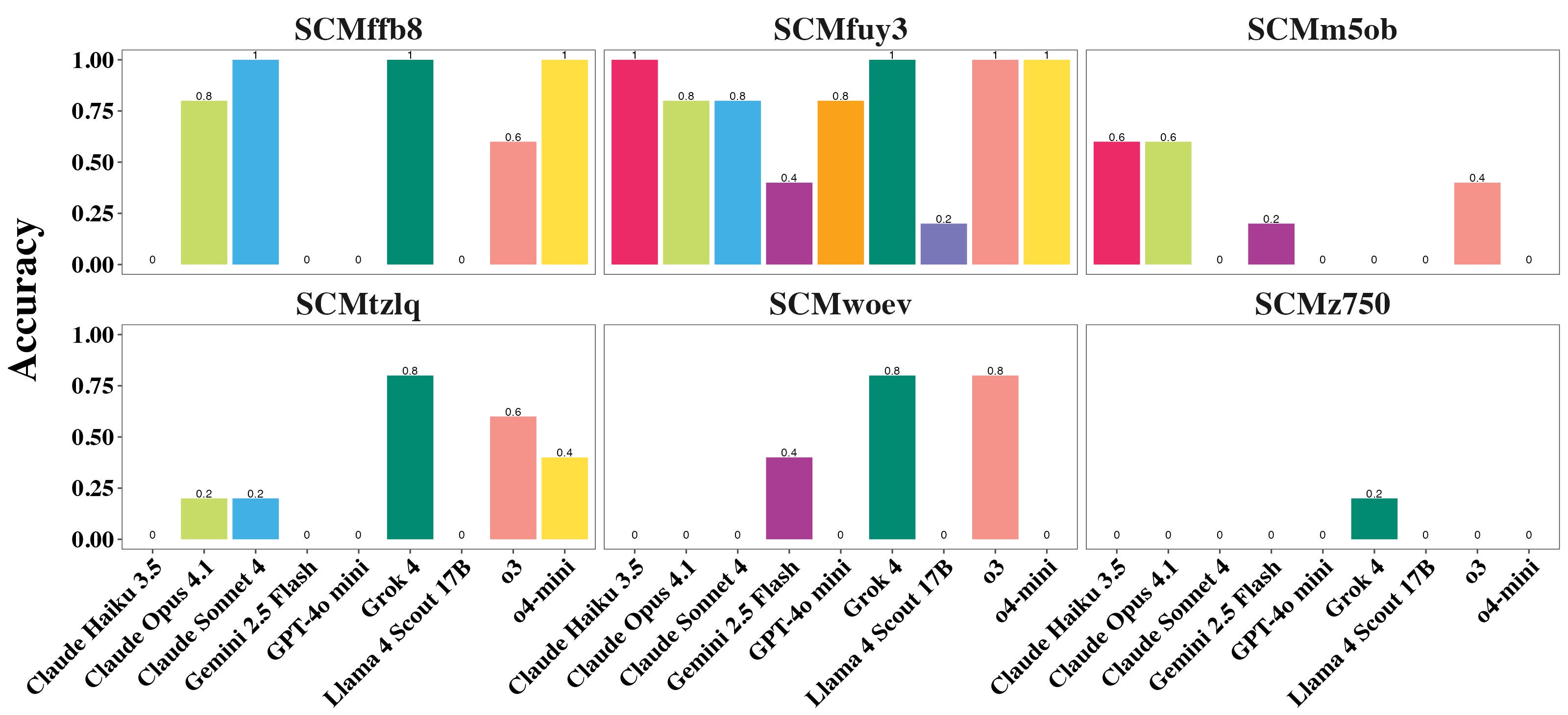}
    \caption{\textbf{Counterfactual reasoning} on counting, extension, and ordering tasks (Figure \ref{fig:cf_not_logical_grids}). Scores are over five random prompt samples.}
    \label{fig:cf_not_logical_tasks}
\end{figure}

\begin{figure}[!h]
    \centering
    \includegraphics[width=\linewidth]{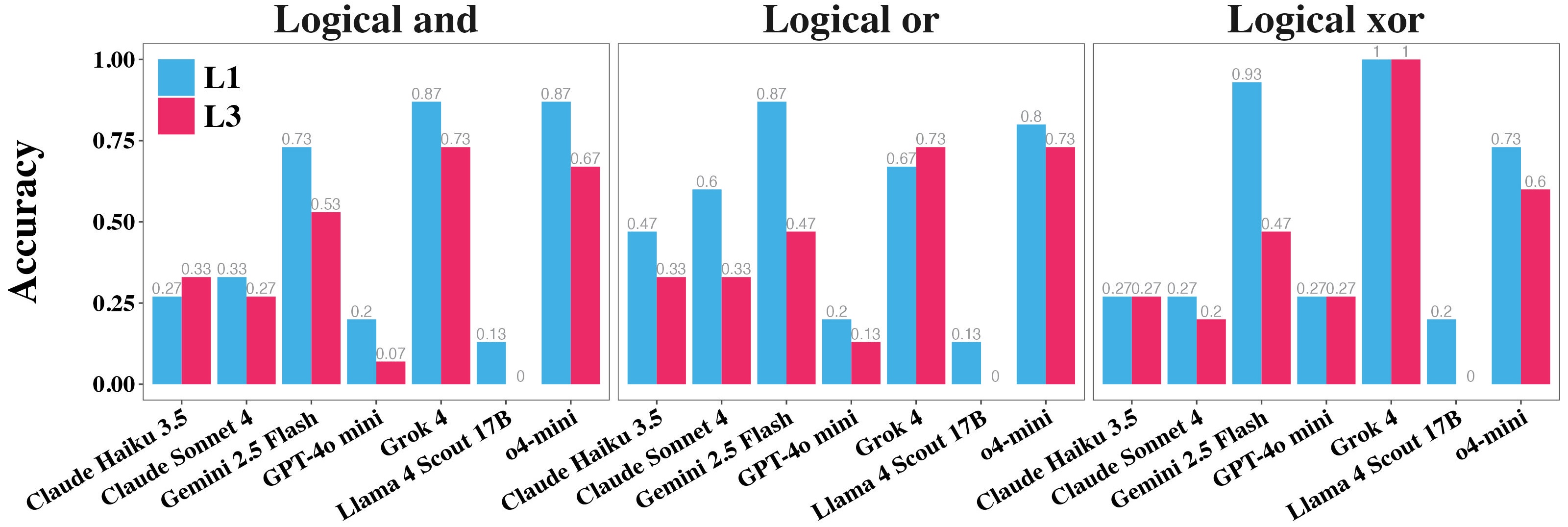}
    \includegraphics[width=\linewidth]{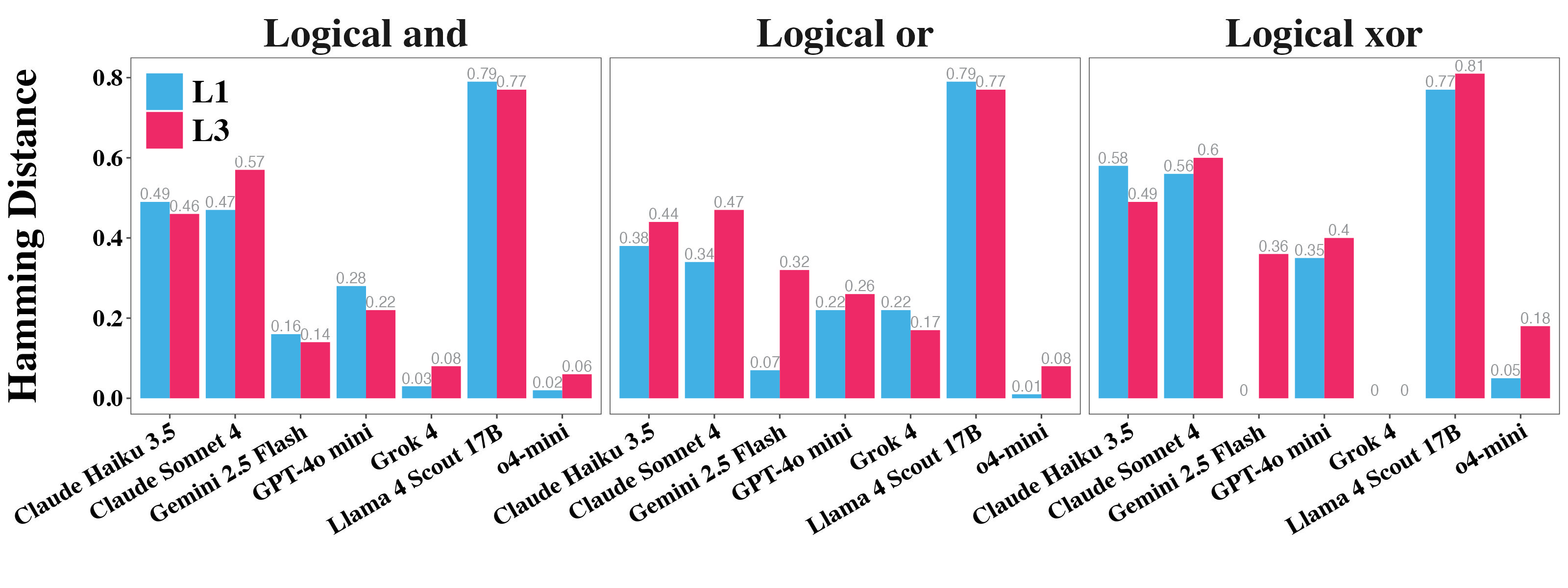}
    \caption{\textbf{Counterfactual reasoning} results on variants of CausalARC task \texttt{SCMdky5}. Scores are over five random prompt samples. Raw data are in Table \ref{tab:logical_results}. Metrics are accuracy (fraction tasks answered correctly; values between 0 and 1; higher is better) and Hamming distance (values between 0 and 1; lower is better). Accurate prediction corresponds to a Hamming distance of 0.}
    \label{fig:logical_results}
\end{figure}

\begin{figure}[!h]
    \centering
    \includegraphics[width=\linewidth]{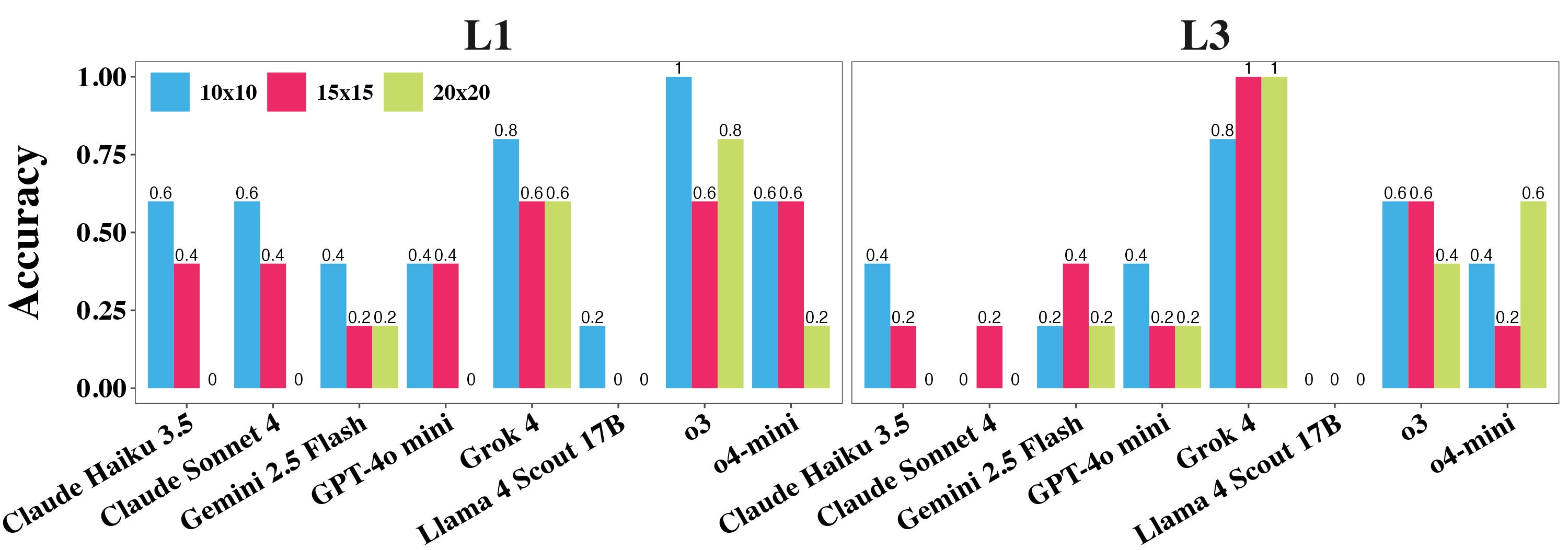}
    \caption{\textbf{Counterfactual reasoning} results on variants of CausalARC task \texttt{SCMtcbq}, where logical operators compose (\textit{xor} after \textit{and}). Accuracy was measured with respect to scaling the dimensionality of the output array ($10 \times 10$, $15 \times 15$, $20 \times 20$). Scores are over five random prompt samples. Raw data are provided in Table \ref{tab:logical_results_mixed}.}
    \label{fig:logical_results_compose}
\end{figure}

\begin{table}[!h]
    \centering
    \begin{adjustbox}{max width=\textwidth}
    \begin{tabular}{l cccc cccc cccc}
    \toprule[1pt]
       & \multicolumn{4}{c}{\textbf{\textsc{LOGICAL AND}}} & \multicolumn{4}{c}{\textbf{\textsc{LOGICAL OR}}} & \multicolumn{4}{c}{\textbf{\textsc{LOGICAL XOR}}}   \\
        \cmidrule(lr){2-5} \cmidrule(lr){6-9} \cmidrule(lr){10-13}  
        & \multicolumn{2}{c}{\textsc{L1}} & \multicolumn{2}{c}{\textsc{L3}} & \multicolumn{2}{c}{\textsc{L1}} & \multicolumn{2}{c}{\textsc{L3}} & \multicolumn{2}{c}{\textsc{L1}} & \multicolumn{2}{c}{\textsc{L3}}   \\
        \cmidrule(lr){2-3} \cmidrule(lr){4-5} \cmidrule(lr){6-7}  \cmidrule(lr){8-9} \cmidrule(lr){10-11} \cmidrule(lr){12-13} 
        & \cmark $\uparrow$ & \textit{HD} $\downarrow$ & \cmark $\uparrow$ & \textit{HD} $\downarrow$ & \cmark $\uparrow$ & \textit{HD} $\downarrow$ & \cmark $\uparrow$ & \textit{HD} $\downarrow$ & \cmark $\uparrow$ & \textit{HD} $\downarrow$ & \cmark $\uparrow$ & \textit{HD} $\downarrow$   \\
        \midrule
       \textit{GPT-4o mini} & 0.20 & 0.28 (0.24) & 0.07 & 0.22 (0.15) & 0.20 & 0.22 (0.16) & 0.13 & 0.26 (0.17) & 0.27 & 0.35 (0.23) & 0.27 & 0.4 (0.26) \\
       \textit{o4-mini} & \textbf{0.87} & \textbf{0.02} (0.07) & 0.67 & \textbf{0.06} (0.12) & 0.80 & \textbf{0.01} (0.03) & \textbf{0.73} & \textbf{0.08} (0.17) & 0.73 & 0.05 (0.13) & 0.60 & 0.18 (0.23) \\ 
       \textit{Claude Haiku 3.5} & 0.27 & 0.49 (0.43) & 0.33 & 0.46 (0.41) & 0.47 & 0.38 (0.45) & 0.33 & 0.44 (0.42) & 0.27 & 0.58 (0.42) & 0.27 & 0.49 (0.41) \\
       \textit{Claude Sonnet 4} & 0.33 & 0.47 (0.45) & 0.27 & 0.57 (0.43) & 0.60 & 0.34 (0.47) & 0.33 & 0.47 (0.46) & 0.27 & 0.56 (0.41) & 0.20 & 0.60 (0.41)\\
       \textit{Grok 4} &  \textbf{0.87} & 0.03 (0.09) & \textbf{0.73} & 0.08 (0.13) & 0.67 & 0.22 (0.33) & \textbf{0.73} & 0.17 (0.30) & \textbf{1.0} & \textbf{0.0} (0.0) & \textbf{1.0} & \textbf{0.0} (0.0) \\
       \textit{Gemini 2.5 Flash} & 0.73 & 0.16 (0.34) & 0.53 & 0.14 (0.18) & \textbf{0.87} & 0.07 (0.25) & 0.47 & 0.32 (0.38) & 0.93 & \textbf{0.0} (0.01) & 0.47 & 0.36 (0.42) \\
       \textit{Llama 4 Scout 17B} & 0.13 & 0.79 (0.37) & 0.0 & 0.77 (0.33) & 0.13 & 0.79 (0.37) & 0.0 & 0.77 (0.33) & 0.20 &  0.77 (0.4) & 0.0 & 0.81 (0.28) \\
    \bottomrule[1pt]
    \end{tabular}
    \end{adjustbox}
    \vspace{2mm} 
    \caption{\textbf{Counterfactual reasoning} results on CausalARC logical reasoning task \texttt{SCMdky5} (Figure \ref{fig:logical_tasks}), where causal functions are a single logical operator. Scores are over five random prompt samples. Results are disaggregated by prompt formulation: L1 prompts include only L1 demonstrations (Figure \ref{fig:l1_prompt_cf_reasoning}), while L3 prompts include alternating L1 and L3 demonstrations (Figure \ref{fig:l3_prompt_cf_reasoning}). Both formulations feature a counterfactual test case. Metrics are accuracy ({\cmark}; values between 0 and 1; higher is better) and Hamming distance (HD, mean and standard deviation; values between 0 and 1; lower is better).  Best performance per column is in bold.}
    \label{tab:logical_results}
\end{table}

\begin{table}[!h]
    \centering
    \begin{adjustbox}{max width=\textwidth}
    \begin{tabular}{l cccc cccc}
    \toprule[1pt]
       & \multicolumn{4}{c}{\textbf{\textsc{COMPOSITION}}} & \multicolumn{4}{c}{\textbf{\textsc{ALTERNATION}}} \\
        \cmidrule(lr){2-5} \cmidrule(lr){6-9} 
        & \multicolumn{2}{c}{\textsc{L1}} & \multicolumn{2}{c}{\textsc{L3}} & \multicolumn{2}{c}{\textsc{L1}} & \multicolumn{2}{c}{\textsc{L3}} \\
        \cmidrule(lr){2-3} \cmidrule(lr){4-5} \cmidrule(lr){6-7}  \cmidrule(lr){8-9}  
        & \cmark $\uparrow$ & \textit{HD} $\downarrow$ & \cmark $\uparrow$ & \textit{HD} $\downarrow$ & \cmark $\uparrow$ & \textit{HD} $\downarrow$ & \cmark $\uparrow$ & \textit{HD} $\downarrow$  \\
        \midrule
       \textit{GPT-4o mini} & 0.27 & 0.34 (0.25) & 0.27 & 0.34 (0.24) & 0.13 & 0.24 (0.11) & 0.13 & 0.24 (0.19) \\
       \textit{o4-mini} &  0.47 & 0.18 (0.23) & 0.40 & 0.19 (0.2) & 0.40 & \textbf{0.13} (0.11) & 0.53 & 0.14 (0.21) \\        
       \textit{o3} & \textbf{0.80} & \textbf{0.06} (0.14) & 0.53 & 0.16 (0.23) & - & - & - & - \\
       \textit{Claude Haiku 3.5} & 0.33 & 0.51 (0.42) & 0.20 & 0.57 (0.36) & 0.13 & 0.50 (0.39) & 0.20 & 0.57 (0.40) \\
       \textit{Claude Sonnet 4} & 0.33 & 0.45 (0.46) & 0.07 & 0.65 (0.38) & 0.13 & 0.50 (0.38) & 0.13 & 0.51 (0.41) \\
       \textit{Grok 4} & 0.67 & \textbf{0.06} (0.13) & \textbf{0.93} & \textbf{0.02} (0.06) & \textbf{0.53} & 0.19 (0.22) & \textbf{0.73} & \textbf{0.10} (0.19) \\
       \textit{Gemini 2.5 Flash} & 0.27 & 0.48 (0.45) & 0.27 & 0.58 (0.42) & 0.40 & 0.38 (0.45) & 0.07 & 0.69 (0.35) \\
       \textit{Llama 4 Scout 17B} & 0.07 & 0.74 (0.39) & 0.0 & 0.86 (0.28) & 0.0 & 0.86 (0.29) & 0.07 & 0.80 (0.34) \\
    \bottomrule[1pt]
    \end{tabular}
    \end{adjustbox}
    \vspace{2mm} 
    \caption{\textbf{Counterfactual reasoning} results on CausalARC logical reasoning tasks, where causal functions are multiple logical operators. Mixed operators include two cases: (1) task \texttt{SCMu3am}, where causal functions alternate by row of the input array (\textit{or}, \textit{and}); and (2) task \texttt{SCMtcbq}, where logical operators compose (\textit{xor} after \textit{and}). Scores are over five random prompt samples. Results are disaggregated by prompt formulation: L1 prompts include only L1 demonstrations (Figure \ref{fig:l1_prompt_cf_reasoning}), while L3 prompts include alternating L1 and L3 demonstrations (Figure \ref{fig:l3_prompt_cf_reasoning}). Both formulations feature a counterfactual test case. Metrics are accuracy ({\cmark}; higher is better) and Hamming distance (HD, mean and standard deviation; lower is better).  Best performance per column is in bold.}
    \label{tab:logical_results_mixed}
\end{table}

\clearpage

\begin{figure}[!h]
    \centering
    \includegraphics[width=\linewidth]{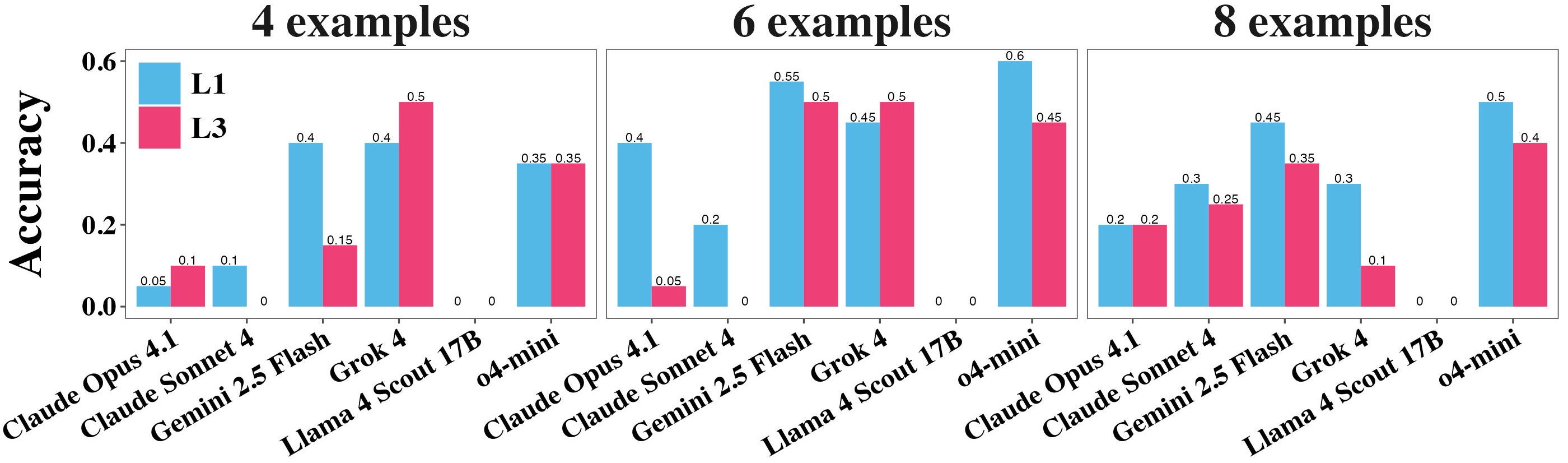}
    \includegraphics[width=\linewidth]{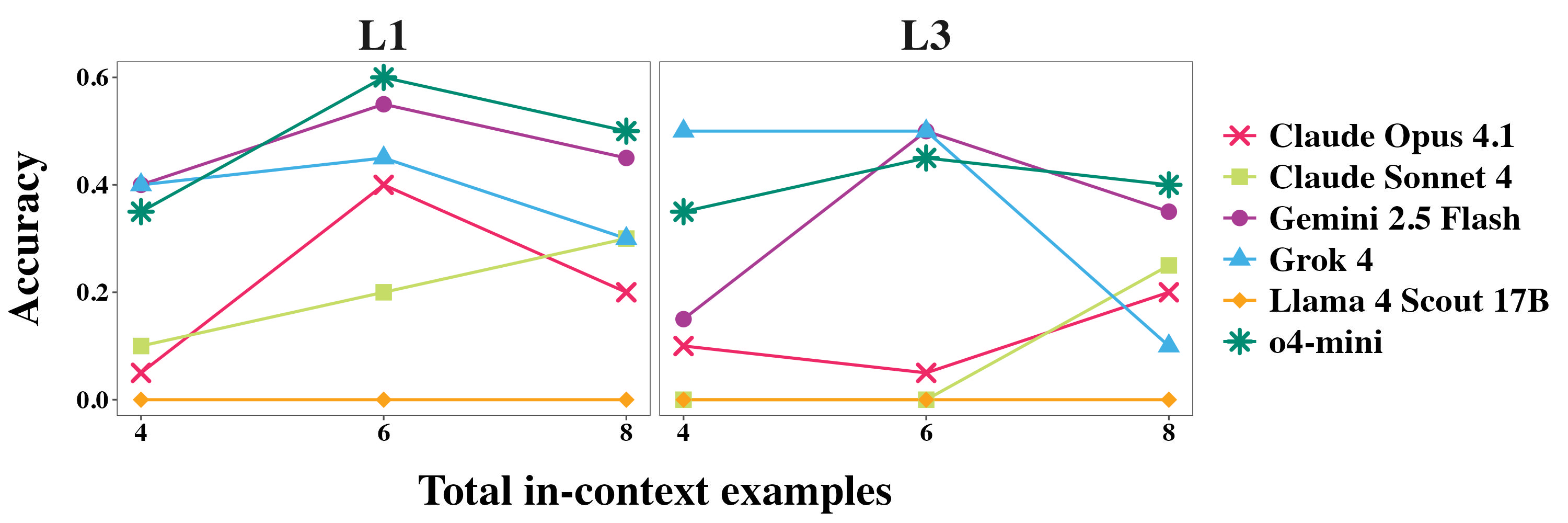}
    \caption{\textbf{Program synthesis} accuracy by total in-context examples. For each of four tasks, five prompts were sampled per level of the PCH. Accuracy was measured with respect to the output array generated by passing a test input array to the LM-generated Python method. Tasks were sampled from \texttt{SCMm5ob} (counting), \texttt{SCMev5t} (counting), \texttt{SCMfwpq} (extension), and \texttt{SCMz750} (extension).}
    \label{fig:prog_synth_bar}
\end{figure}

\begin{figure}[!h]
    \centering
    \includegraphics[width=\linewidth]{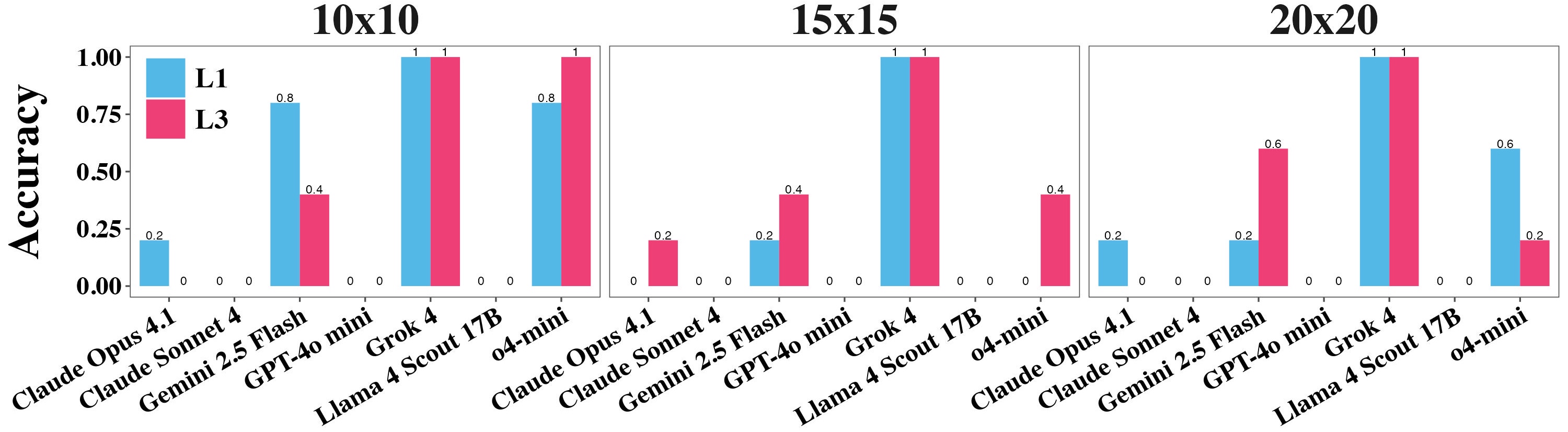}
    \includegraphics[width=\linewidth]{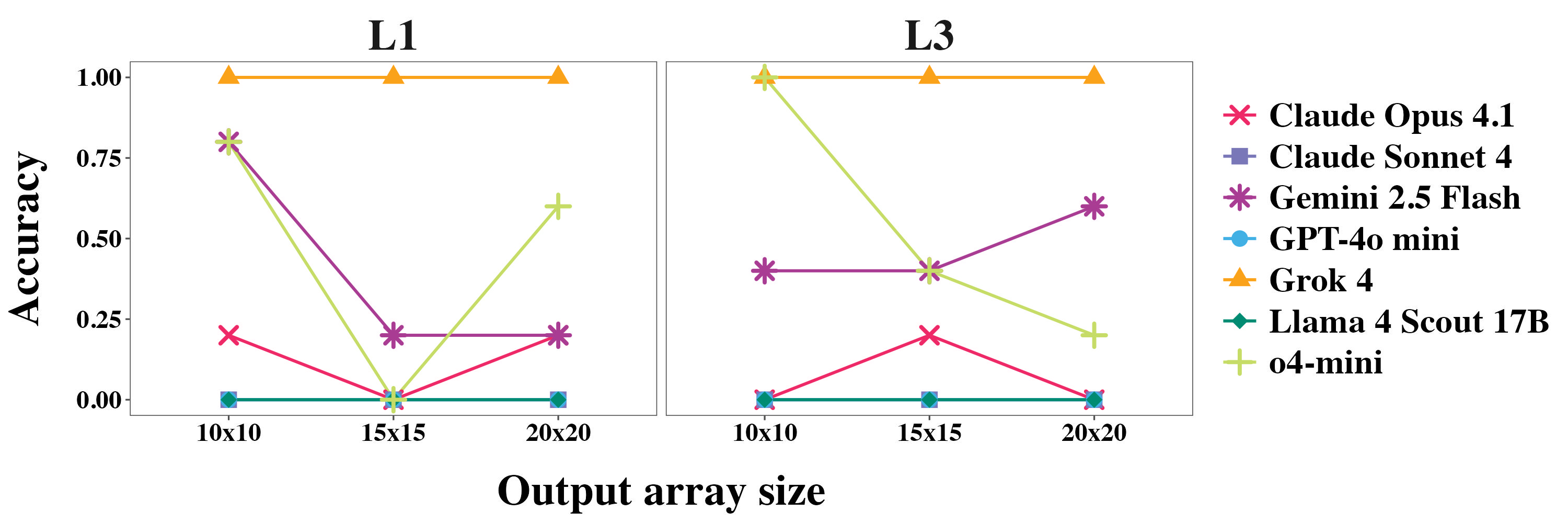}
    \caption{\textbf{Causal discovery with logical reasoning} accuracy as complexity scales (with respect to output array size). For each of three CausalARC tasks sampled from task \texttt{SCMtcbq} (where logical operators compose \textit{xor} after \textit{and}), five prompts were sampled per level of the PCH. The LM was asked to predict the logical operators acting as causal functions, which requires knowledge of the causal parents of each output cell as a prerequisite.}
    \label{fig:discovery_logical}
\end{figure}

\clearpage

\begin{table}[!h]
    \centering
    \begin{tabular}{l c c c}
    \toprule[1pt]
       $n$ & \textsc{logical and} & \textsc{logical or} & \textsc{logical xor} \\
    \midrule
        \textit{1k} & 1.0 & 0.0 & 40.0 \\
        \textit{5k} & 1.0 & 0.0 & 42.0\\
        \textit{10k} & 0.0 & 0.0 & 40.0\\
    \bottomrule[1pt]
    \end{tabular}
    \vspace{2mm}
    \caption{Causal discovery with PC algorithm \citep{spirtes2000causation} on CausalARC task \texttt{SCMdky5}. Distributions of L1 arrays were sampled  at $n = 1000$, $n = 5000$, and $n = 10000$ observations per distribution. Arrays were flattened before being passed to PC. Reported values are the structural Hamming distance for the predicted causal graph with respect to the true graph as data sample size increases. A structural Hamming distance of 0 indicates that the output graph was identical to the true graph. PC algorithm was performed with chi-square conditional independence tests ($\alpha = 0.01$) using the \texttt{causal-learn} Python package (\href{https://causal-learn.readthedocs.io/en/latest/}{https://causal-learn.readthedocs.io/en/latest/}).}
    \label{tab:pc_results}
\end{table}

\begin{figure}[!h]
    \centering
    \fbox{\begin{tabular}{c}
    \textsc{true graph} \\
    \midrule \\
       \includegraphics[width=0.8\linewidth]{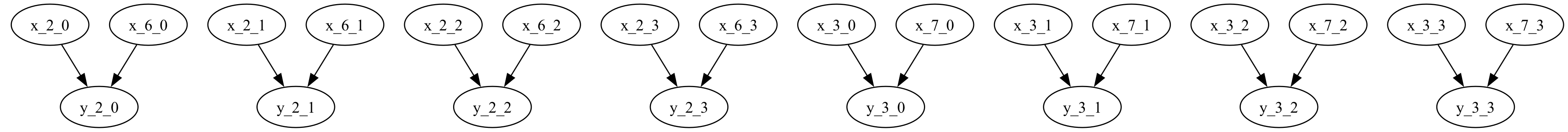} \\
    \includegraphics[width=0.8\linewidth]{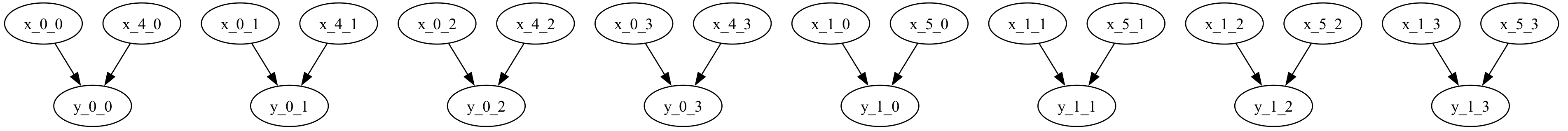} \\
    \end{tabular}} \\
    \vspace{5mm}
    \fbox{\begin{tabular}{c}
    \textsc{logical and $|$ $n = 5000$} \\
    \midrule \\
       \includegraphics[width=0.4\linewidth]{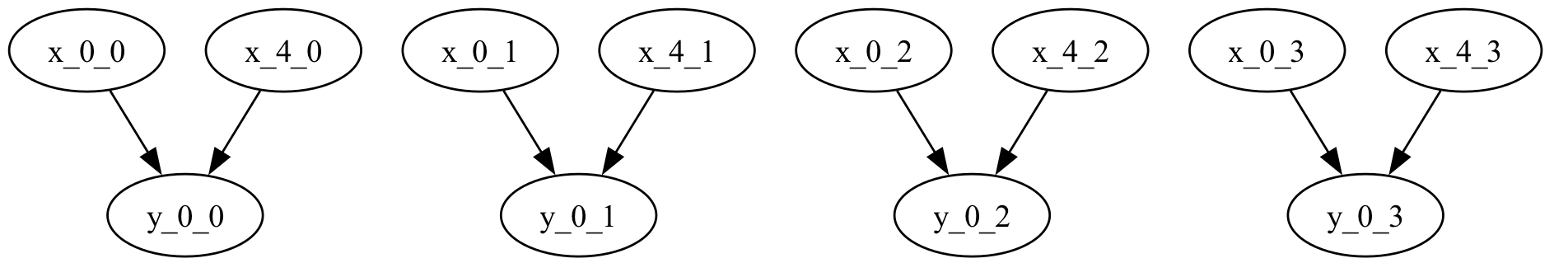} \\
    \includegraphics[width=0.4\linewidth]{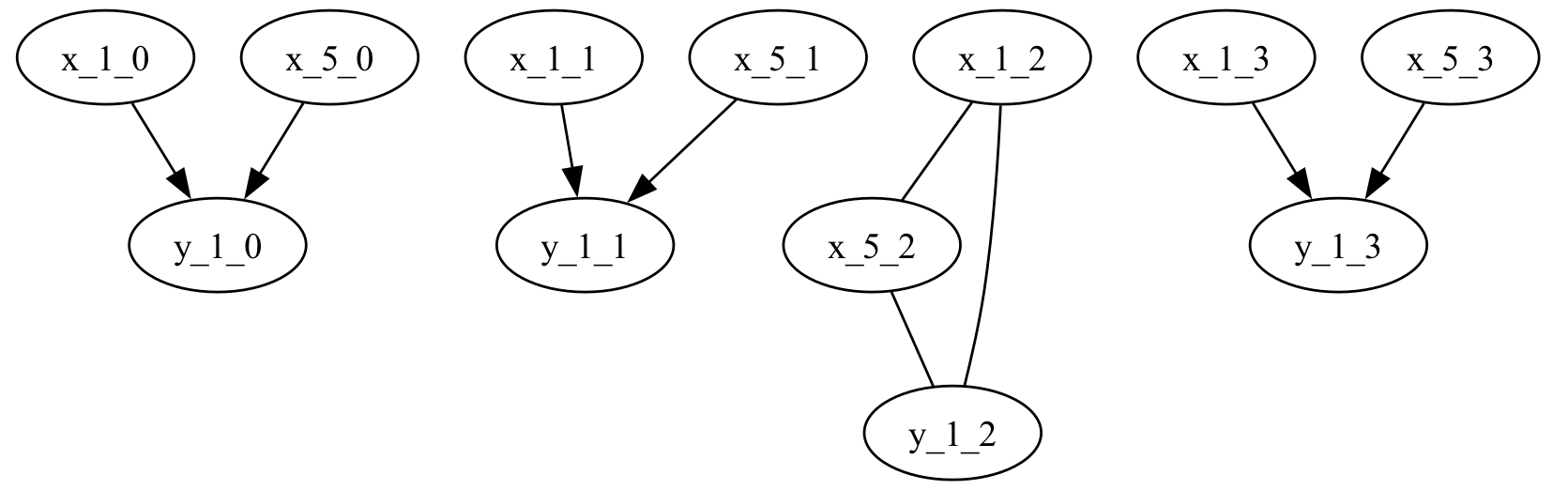} \\
    \includegraphics[width=0.4\linewidth]{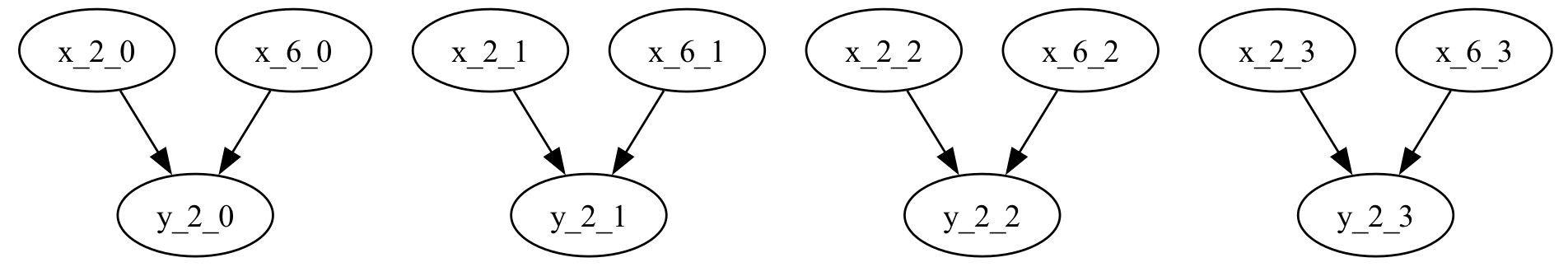} \\
    \includegraphics[width=0.4\linewidth]{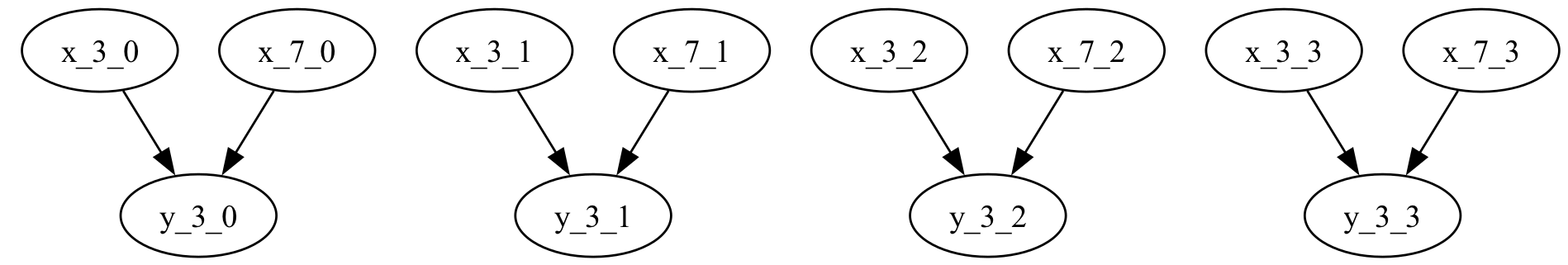} \\
    \end{tabular}}
    \hspace{5mm}
    \fbox{\begin{tabular}{c}
    \textsc{logical xor $|$ $n = 10000$} \\
    \midrule \\
       \includegraphics[width=0.4\linewidth]{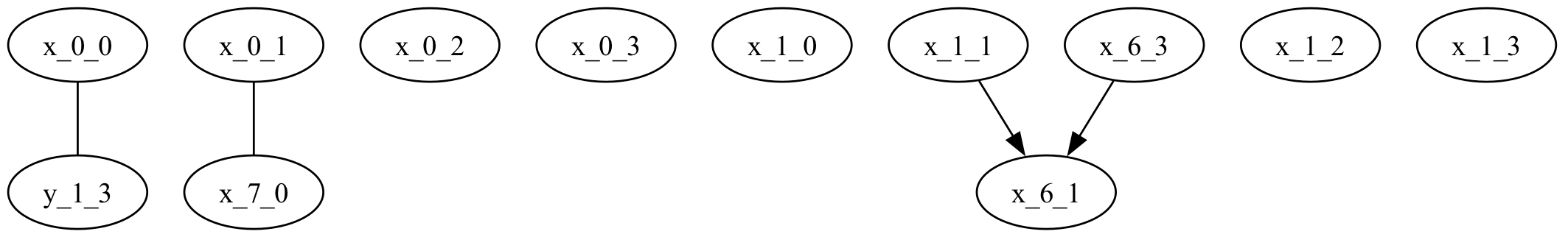} \\
    \includegraphics[width=0.4\linewidth]{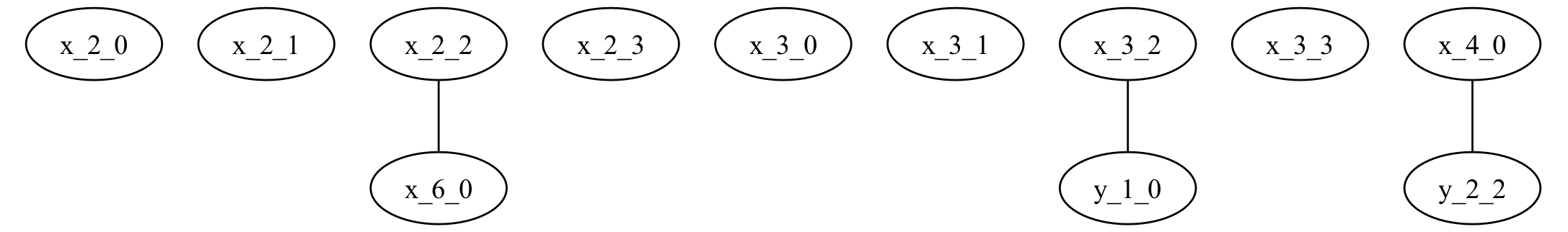} \\
    \includegraphics[width=0.4\linewidth]{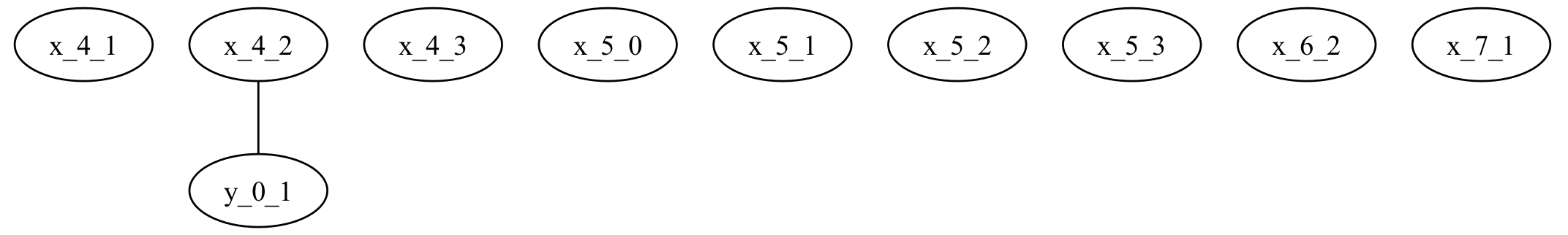} \\
    \includegraphics[width=0.4\linewidth]{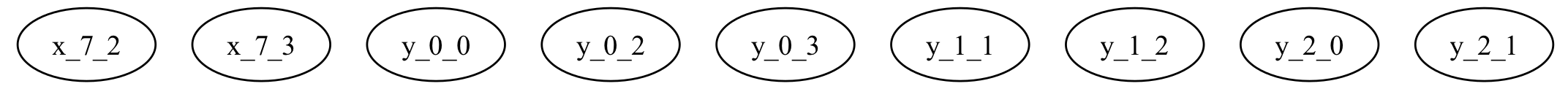} \\
    \includegraphics[width=0.25\linewidth]{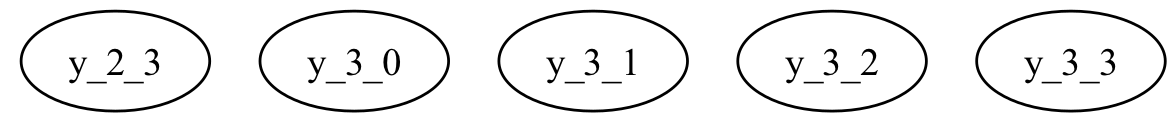} \\
    \end{tabular}}
    \caption{Causal graphs predicted by PC algorithm \citep{spirtes2000causation} for CausalARC task \texttt{SCMdky5}. PC algorithm was performed with chi-square conditional independence tests ($\alpha = 0.01$) using the \texttt{causal-learn} Python package (\href{https://causal-learn.readthedocs.io/en/latest/}{https://causal-learn.readthedocs.io/en/latest/}). Data sample size is denoted by $n$.}
    \label{fig:pc_results}
\end{figure}

%%%%%%%%%%%%%%%%%%%%%%%%%%%%%%%%%%%%%%%%%%%%%%%%%%%%%%%%%%%%

\end{document}